\documentclass[10pt,journal,compsoc,dvipsnames]{IEEEtran}

\ifCLASSOPTIONcompsoc
  \usepackage[nocompress]{cite}
\else
  \usepackage{cite}
\fi

\usepackage{bbm}
\usepackage{amsmath}
\usepackage{graphicx}
\usepackage[colorlinks=true, allcolors=blue]{hyperref}

\usepackage{amsfonts}
\usepackage{stmaryrd}
\usepackage{bm}
\usepackage{amssymb}
\usepackage{amsthm}
\usepackage{booktabs}
\usepackage{array}
\usepackage{multirow}
\usepackage{fancyhdr}  %
\usepackage{enumerate}
\usepackage[linesnumbered,ruled]{algorithm2e}  %
\SetKwComment{Comment}{$\triangleright$\ }{}

\usepackage{dsfont}

\ifCLASSOPTIONcompsoc
 \usepackage[caption=false,font=footnotesize,labelfont=sf,textfont=sf]{subfig}
\else
 \usepackage[caption=false,font=footnotesize]{subfig}
\fi

\usepackage{footnote}
\makesavenoteenv{tabular}

\usepackage{verbatim}

\newcommand{\diag}{\mathop{\mathrm{diag}}\nolimits}
\newcommand{\pois}{\mathop{\mathrm{Poisson}}\nolimits}
\newcommand{\bern}{\mathop{\mathrm{Ber}}\nolimits}

\newcolumntype{M}[1]{>{\centering\arraybackslash}m{#1}}

\DeclareMathOperator*{\argmin}{arg\,min}

\begin{document}
\definecolor{Plum}{rgb}{0.56, 0.27, 0.52}

\title{Learning Inter-Modal Correspondence and Phenotypes from Multi-Modal Electronic Health Records}

\author{Kejing Yin, %
        William K. Cheung, ~\IEEEmembership{Member,~IEEE,}
        Benjamin C. M. Fung, ~\IEEEmembership{Senior Member,~IEEE,}
        Jonathan Poon%
\IEEEcompsocitemizethanks{\IEEEcompsocthanksitem Kejing Yin and William K. Cheung are with the Department of Computer Science, Hong Kong Baptist University, Kowloon Tong, Hong Kong.\protect\\
E-mail: \{cskjyin, william\}@comp.hkbu.edu.hk
\IEEEcompsocthanksitem Benjamin C. M. Fung is with the School of Information Studies, McGill University, Montreal, Canada.\protect\\
E-mail: ben.fung@mcgill.ca
\IEEEcompsocthanksitem Jonathan Poon is with Hong Kong Hospital Authority, Hong Kong.\protect\\
E-mail: jonathan@ha.org.hk
}%
}

\markboth{IEEE Transactions on Knowledge and Data Engineering,~Vol.~XX, No.~X, August~20XX}%
{Yin \MakeLowercase{\textit{et al.}}: Learning Inter-Modal Correspondence and Phenotypes from Multi-Modal Electronic Health Records}

\IEEEtitleabstractindextext{%
\begin{abstract}
Non-negative tensor factorization has been shown a practical solution to automatically discover phenotypes from the electronic health records (EHR) with minimal human supervision. Such methods generally require an input tensor describing the inter-modal interactions to be pre-established; however, the correspondence between different modalities (e.g., correspondence between medications and diagnoses) can often be missing in practice. Although heuristic methods can be applied to estimate them, they inevitably introduce errors, and leads to sub-optimal phenotype quality. This is particularly important for patients with complex health conditions (e.g., in critical care) as multiple diagnoses and medications are simultaneously present in the records. To alleviate this problem and discover phenotypes from EHR with unobserved inter-modal correspondence, we propose the collective hidden interaction tensor factorization (cHITF) to infer the correspondence between multiple modalities jointly with the phenotype discovery. We assume that the observed matrix for each modality is marginalization of the unobserved inter-modal correspondence, which are reconstructed by maximizing the likelihood of the observed matrices. Extensive experiments conducted on the real-world MIMIC-III dataset demonstrate that cHITF effectively infers clinically meaningful inter-modal correspondence, discovers phenotypes that are more clinically relevant and diverse, and achieves better predictive performance compared with a number of state-of-the-art computational phenotyping models.
\end{abstract}

}

\maketitle

\section{Introduction}\label{sec:introduction}

The adoption of electronic health records (EHR) has been growing rapidly over the last decade. Consequently, a large amount of clinical data about patients were accumulated, including diagnosis codes, medication prescriptions, and laboratory tests, triggering numerous studies on secondary analysis of the EHR data to accelerate clinical research~\cite{jensen2012mining,wang2016diagnosis,luo2016tensor,xu2017patient,liu2019complication}. However, due to the complex nature of healthcare and the data collection process, the raw EHR data normally contains heavy missingness, frequent inaccuracy and potential bias~\cite{hripcsak2013next}, hindering the application of data-driven approaches to analyzing them. Thus, it is often required to map the raw EHR data to clinically meaningful concepts, \textit{i.e.}, \textit{phenotypes}~\cite{ho2014limestone}, and the process of discovering phenotypes from the raw EHR data is called \textit{phenotyping}. A phenotype is formally defined as a group of clinical features that are highly relevant and better characterizes the health status of patients. Conventionally, phenotyping is done in a supervised manner, involving an iterative process of manually labelling the case and control patients, and summarizing and refining the discriminative features for a pre-specified diseases~\cite{lasko2013computational}, which is obviously time-consuming and labor-intensive~\cite{hripcsak2013next}. To expedite the phenotyping process, applying machine learning methods, especially the unsupervised ones, to automatically extract phenotypes from large-scale EHR data without intensive human supervision has been gaining increasing attention recently~\cite{dl:inHealthInfomatics:review,wang2015rubik,kim2017discriminative}.

Among the efforts, the non-negative tensor CP factorization has been found particularly promising due to its high degree of interpretability and capability of preserving high-order interactions. For example, a third-order tensor $\mathcal{X}$ can be constructed to represent the interactions among patients, diagnoses and medications, where the tensor entry $x_{pdm}$ can be interpreted as ``patient $p$ was prescribed medication $m$ in response to the diagnosis $d$''. Like its two-dimensional counterpart, the non-negative matrix factorization~\cite{lee1999learning}, the non-negative CP factorization takes a high-order interaction tensor as input and learns a set of non-negative rank-one tensors to approximate the input tensor. As a result of its linearity and additive property, the non-negative CP factorization is able to reveal the ``parts-of-whole'' relationship underlying data~\cite{hu2015scalable,ho2014limestone}.

Despite of the great success in computational phenotyping by non-negative CP factorization, there are several fundamental challenges that hinders its application to some of the real-world scenarios, including:

\textbf{Challenge 1: Hidden interactions.} As the input to any ordinary tensor factorization models, including the non-negative CP factorization, the tensor need to be well defined to represent the interaction between different modalities. However, this information is often not available in practice. Take diagnoses and medications as an example, the EHR data typically only contain the list of patients' diagnoses and the list of the medication prescriptions, yet the correspondence between the medication and the diagnoses are totally unrecorded. Existing methods turn to an alternative strategy, namely to consider the ``co-occurrence'' relationship, which implicitly assume that all medications and diagnoses co-occurring in the same clinical visit would correspond to each other equally. %

\begin{figure}
  \centering
  \includegraphics[width=0.9\columnwidth]{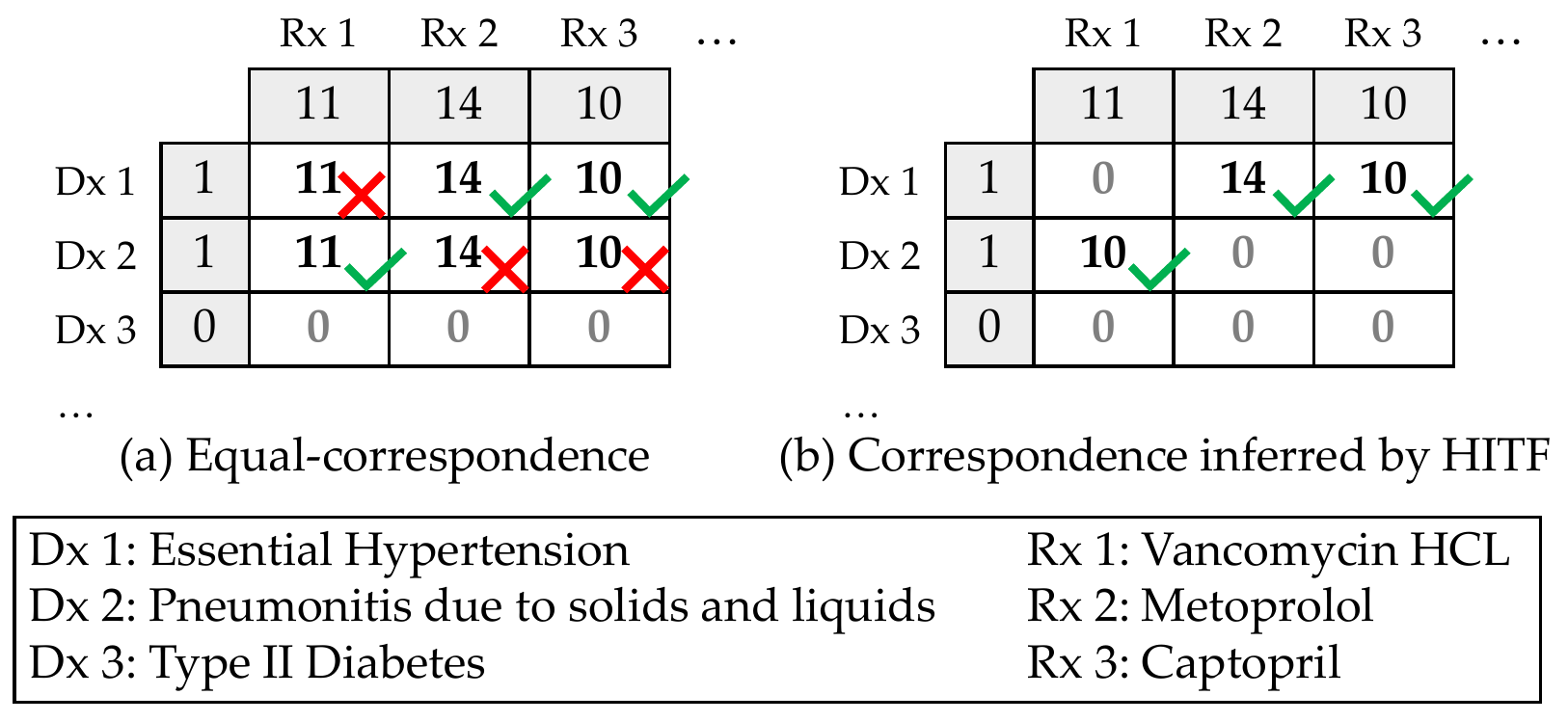}
  \caption{Real examples of the diagnosis-medication correspondence from MIMIC-III dataset: Each row denotes a diagnosis (Dx1/Dx2/Dx3) and the ``1/0'' value next to it indicates if the diagnosis is present or not. Each column denotes a medication (Rx1/Rx2/Rx3) and the number underneath each medication denotes the amount of prescribed medications. (a) Adopting equal-correspondence strategy. (b) Correspondence inferred by the proposed HITF model, which is more reasonable.}\label{fig:equal-corrs}
\end{figure}

This ``equal-correspondence'' assumption can be reasonable for some specific types of datasets,  \textit{e.g.}, primary care or outpatient data, where patients typically have very distinct diagnosis in each clinical visit and the medications prescribed are associated with the diagnosis. However, the real-world EHR data can often be highly complex, for instance the inpatient care or intensive care data, where patients are generally with very complex medical conditions: patients could have more than a dozen of diagnosis codes assigned and tens of medications prescribed. The ``equal-correspondence'' assumption no longer hold in this case. As a real example, Fig.~\ref{fig:equal-corrs} shows a part of the diagnosis codes and medications of a patient extracted from the MIMIC-III dataset. This patient was diagnosed essential hypertension, and prescribed with three medications: vancomycin HCL, metoprolol and captopril. Fig.~\ref{fig:equal-corrs}(a) is the correspondence based on the ``equal-correspondence'' assumption, where metoprolol is assumed to be corresponding to hypertension and pneumonitis equally as they co-occurred in the same hospital visit, and all three medications are assumed to correspond to hypertension equally as well. However, in clinical practice, metoprolol is often used to treat hypertension, but not pneumonitis. Recall that the non-negative CP factorization aims at approximating the input tensor, constructing the input tensor under such assumption cause inevitable error. On the other hand, our proposed model, as illustrated in Fig.~\ref{fig:equal-corrs}{(b)}, can infer that the medication metoprolol only corresponds to hypertension. This is achieved because HITF does not rely on the ``equal-correspondence'' assumption to construct the input data, but rather it explicitly infers the hidden correspondence between the two modalities.

\textbf{Challenge 2: Multi-modality.} The unprecedented richness of EHR is unfolded through its multi-modal nature. Apart from the diagnoses and medications, a typical EHR dataset (\textit{e.g.} MIMIC-III) can involve other modalities including procedures, laboratory test results, and input fluids. Each of them contains unique information about the patients in different aspects, yet they usually can be very correlated and often interact with each other in a complex way~\cite{johnson2016machine}, posing additional difficulties in modeling the multi-modal EHR data in one single tensor and discovering phenotypes from it. First of all, with more modalities involved, it is more difficult to define unambiguously the interaction among modalities, based on which the input tensor should be constructed. As the number of modalities grows, the physical meaning of the tensor entries become subtle and the interpretability could suffer. For instance, the interactions among diagnoses, medications and lab tests can be vaguer than that between only diagnoses and medications, in that lab tests could be requested for confirming a diagnosis, or otherwise monitoring the usage of a medication. Moreover, the running time of tensor factorization models typically grow exponentially with the number of modalities, making it unscalable to multi-modal data. Furthermore, different modalities often have different data types, for example binary data for the diagnoses, indicating the presence or absence of a diagnosis code in patients' records, while the amount of fluids input to the patients are recorded in real values. One of the keys to improving the quality of the phenotypes derived is appropriate combination of different modalities to maximize the interpretability.

To tackle the above challenges, we propose a novel framework \textbf{c}ollective \textbf{H}idden \textbf{I}nteraction \textbf{T}ensor \textbf{F}actorization (cHITF) that allows different modalities to be combined so that the unobserved inter-modality correspondence together with the phenotypes can be simultaneously inferred with only marginalized observations of the hidden interaction tensor. We conducted extensive experiments based on the real-world MIMIC-III and eICU datasets. The inferred inter-modal correspondences and phenotypes are found to be highly interpretable and clinically relevant as confirmed by clinicians, and the phenotypes are found to outperform the state-of-the-art computational phenotyping models on predictive task as well.

\section{Related Work}\label{sec:related-work}
Among the earliest efforts of computational phenotyping, \cite{ho2014limestone} proposed to apply the non-negative tensor factorization method to discover multiple phenotypes from the EHR data. Various constraints have later been incorporated in different ways to further improve the interpretability of the results, namely to derive phenotypes that are more clinically meaningful and relevant. For example, \cite{ho2014marble} incorporates a bias component to capture the overall baseline characteristics among the whole population; in \cite{wang2015rubik}, existing medical knowledge are integrated as additional constraints by defining a guidance matrix; in \cite{yang2017tagited} and \cite{kim2017discriminative}, label information including in-hospital mortality and medical cost are leveraged to discover more discriminative phenotypes by incorporating a supervised term, and \cite{kim2017discriminative} also exploits the clustering structure among the diagnoses and medications to further ensure the discovered phenotypes being more distinct from each other. More recently, researchers also start to model the temporal information in EHR to enhance the interpretability~\cite{yin2019learning,perros2017spartan}.

However, these models construct the input tensor based on the ``equal-correspondence'' assumption illustrated earlier, which limits its applicability to phenotyping from clinical data with more complex inter-modal interactions, \textit{e.g.}, data collected in ICU. The work most similar to ours is \cite{arxiv2016cmf,gunasekar2016mining}, which performs non-negative matrix factorization over a collection of matrices with one shared dimension, each corresponding to one modality. Yet this model essentially ignores the  interactions among different modalities which has been proven important by numerous computational phenotyping models based on the tensor factorization framework as mentioned above. In summary, it remains a fundamentally challenging issue to discover phenotypes from clinical data where the ``equal-correspondence'' assumption between modalities are no longer reliable, while preserving the inter-modal interactions.

Other matrix/tensor factorization models aiming at improving robustness while modeling the skewed data, such as expectile matrix factorization~\cite{zhu2017expectile} and M-estimation~\cite{tu2019m}, also have potentials to be applied to discover phenotypes from EHR data, which are generally long-tail distributed. However, they do not consider interactions and correspondence between different modalities that are explicitly modeled by our proposed framework.

Besides, some recent research focus on applying deep learning and graph mining techniques to analyze EHR data, \textit{e.g.} \cite{choi2017gram,song2019medical}. Most of them are supervised predictive models, requiring a large amount of expensive label efforts. Although there also exist unsupervised ones, they focus on learning representations for down-stream tasks.
On the contrary, tensor factorization models, including our proposed one, primarily focuses on discovering interpretable factors underlying data, and do not rely on labels to learn.

\section{Background and Preliminaries}\label{sec:preliminaries}
Before presenting our framework, we provide necessary background and preliminaries about tensor and its non-negative CP factorization. We also give an overview of applying tensor factorization framework to computational phenotyping problem. Unless otherwise specified, we use the notations defined in Table~\ref{tab:notations}.

\begin{table}
\centering
\caption{Symbols and notations used in this paper}\label{tab:notations}
\begin{tabular}{cl}
\toprule
\textbf{Symbol} & \multicolumn{1}{c}{\textbf{Definition}} \\ \midrule
   $\mathcal{V}$      &  Observations: a collection of matrices  \\
   $\mathbf{V}^{(n)}$ &  The $n^{th}$ matrix with non-binary values in $\mathcal{V}$ \\
   $\mathbf{V^\prime}^{(n)}$ &  The $n^{th}$ matrix with binary values in $\mathcal{V}$ \\
   $\hat{\mathbf{V}}^{(n)}$ & The reconstruction of the $\mathbf{V}^{(n)}$ \\
   $\mathcal{X}$      &  The hidden interaction tensor \\
   $\mathcal{U}$      &  The collection of the factor matrices \\
   $\mathbf{U}^{(s)}$ &  The factor matrix corresponding to patients \\
   $\mathbf{U}^{(n)}$ &  The factor matrix corresponding to the $n^{th}$ modality \\
   $\mathbf{e}$       &  Vector of all ones \\
   $\sigma^2$         & Variance of Gaussian distribution \\\hline
   $I_s$              & Size of the shared dimension (patients)\\
   $I_n$              & Size of the $n^{th}$ modality\\
   $N$                & Number of modalities\\
   $R$                & Number of phenotypes\\\hline
   $\circ$           & Outer product \\
   $\mathds{1}(\cdot)$ & The indicator function \\
   $\Phi(\cdot)$      & The CDF of standard Gaussian distribution \\
   $\operatorname{erf}(\cdot)$  & the error function \\
   $\operatorname{diag}(\mathbf{x})$  & A diagonal matrix with vector $\mathbf{x}$ on its diagonal\\
   $\llbracket\cdot,\dots,\cdot\rrbracket$   & Shorthand for CP factorization \\\bottomrule
\end{tabular}
\end{table}

\subsection{Tensor and Its Non-negative CP Factorization}

\noindent\textbf{Tensor and Rank-One Tensor.}~ Tensors are multidimensional arrays~\cite{papalexakis2017tensors}. The \textit{order} of a tensor (\textit{a.k.a.} modes or ways) refers to the number of dimensions of the tensor. For example, a two-dimensional matrix is a second-order tensor. A $D^{th}$-order tensor $\mathcal{X}$ is \textit{rank-one} if it can be written as the outer product of $D$ vectors, \textit{i.e.}, $\mathcal{X}=\mathbf{u}^{(1)}\circ\mathbf{u}^{(2)}\circ\dots\circ\mathbf{u}^{(D)}$.

\medskip\noindent\textbf{CP Factorization.}~ The CP factorization~\cite{hitchcock1927expression,chi2012tensors} approximates the input tensor with the sum of component rank-one tensors. For example the CP decomposition of a $N^{th}$-order tensor $\mathcal{X}$ is defined as follows:
\begin{equation}\label{Eq:CP}
 \mathcal{X} \approx \llbracket \mathbf{U}^{(1)},\mathbf{U}^{(2)},\dots,\mathbf{U}^{(D)} \rrbracket = \sum_{r=1}^R \mathbf{u}_r^{(1)}\circ \mathbf{u}_r^{(2)} \circ \dots \circ \mathbf{u}_r^{(D)},
 \end{equation}
where $\llbracket\cdot\rrbracket$ is a shorthand for the CP factorization, and $R$ is the number of rank-one tensors. $\mathbf{U}^{(d)}$ is called the \textit{CP factor matrix} corresponding to the $d^{th}$ mode of the tensor, and is obtained by combining the vectors from the rank-one tensors, \textit{i.e.}, $\mathbf{U}^{(d)}=[\mathbf{u}^{(d)}_1, \mathbf{u}^{(d)}_2,\dots,\mathbf{u}^{(d)}_R]$.

\medskip\noindent\textbf{Tensor Slice.}~ We define a \textit{slice} of a tensor as a matrix obtained by fixing all but two indices of the tensor. For instance, for a third-order tensor $\mathcal{X}$, the slice $\mathbf{X}_{:j:}$ is obtained by varying two indices of the tensor (the first and the third mode in this example) while fixing the remaining one. With the notion of the CP factor matrices defined as aforementioned, the slice $\mathbf{X}_{:j:}$ can be written as~\cite{kolda2009tensor}:
\begin{equation}\label{eq:slice} \mathbf{X}_{:j:} \approx \mathbf{U}^{(1)} \diag( \mathbf{u}_{j:}^{(2)}) {\mathbf{U}^{(3)}}^\top. \end{equation}

\medskip\noindent\textbf{Tensor Marginalization.} We define the maginalization of a tensor as a matrix obtained by summing all slices of the tensor along two fixed modes. For a third-order tensor $\mathcal{X}$, the accumulation along the second dimension is:
\begin{equation}\label{eq:marginalization} \mathbf{V} = \sum_{j=1}^{J} \mathbf{X}_{:j:} \approx \mathbf{U}^{(1)} \diag( \mathbf{e}^\top \mathbf{U}^{(2)}) {\mathbf{U}^{(3)}}^\top,
\end{equation}
where $\mathbf{e}$ is the vector of all ones.

\subsection{NTF for Computational Phenotyping}
Non-negative constraints can be imposed on the factor matrices when applying the CP factorization, leading to an additive model which is often referred to as the non-negative tensor factorization (NTF)~\cite{shashua2005non,chi2012tensors}. NTF is known to yield interpretable factors: the input tensor can be regarded as the sum of $R$ latent concepts, each one corresponds to a rank-one tensor. In each rank-one tensor, the non-negative vector $\mathbf{u}^{(n)}_r$ represents the soft-clustering membership of the corresponding items in the $n^{th}$ mode.

The process of applying NTF to computational phenotyping roughly contains three steps: constructing the input tensor, solving for the factor matrices, and extracting interpretations of the phenotype definitions~\cite{ho2014limestone}. As already briefly described in Section~\ref{sec:introduction}, existing models adopt the ``equal-correspondence'' strategy to construct the input tensor. Specifically, they set the tensor values to either the number of co-occurrence of items in different modes, or zero or one indicating the presence or absence of the co-occurrence. After the input tensor is constructed, the NTF algorithms are applied to solve for the factor matrices. One of the most popular NTF algorithms is CP-APR~\cite{chi2012tensors}, which assumes Poisson distributions for the entries of the input tensor to model the counting data, and formulate the problem as maximizing the likelihood of the input tensor. Formally, CP-APR solves the following optimization problem~\cite{chi2012tensors}:
\begin{equation}\label{eq:cntfloss} %
\begin{aligned}
\underset{\mathbf{U}^{(n)}}{\argmin}
\quad & f(\mathcal{M}) \equiv \sum_{\mathbf{i}} m_\mathbf{i} - x_\mathbf{i} \log m_\mathbf{i} \\
\text{s.t.} \quad & \mathcal{M} = \llbracket \lambda; \mathbf{U}^{(1)},  \dots, \mathbf{U}^{(N)} \rrbracket,\\
& \mathbf{U}^{(n)} \geq \mathbf{0}\quad \text{for } n=1,\dots,N,\\
& \Vert \mathbf{u}^{(n)}_r \Vert_1 = 1 \quad \forall r, n,
\end{aligned}
\end{equation}
where $\boldsymbol{\lambda}$ absorbs the $\ell_1$ norms of the columns of the matrix factors.

\begin{figure}
    \centering
    \includegraphics[width=0.9\columnwidth]{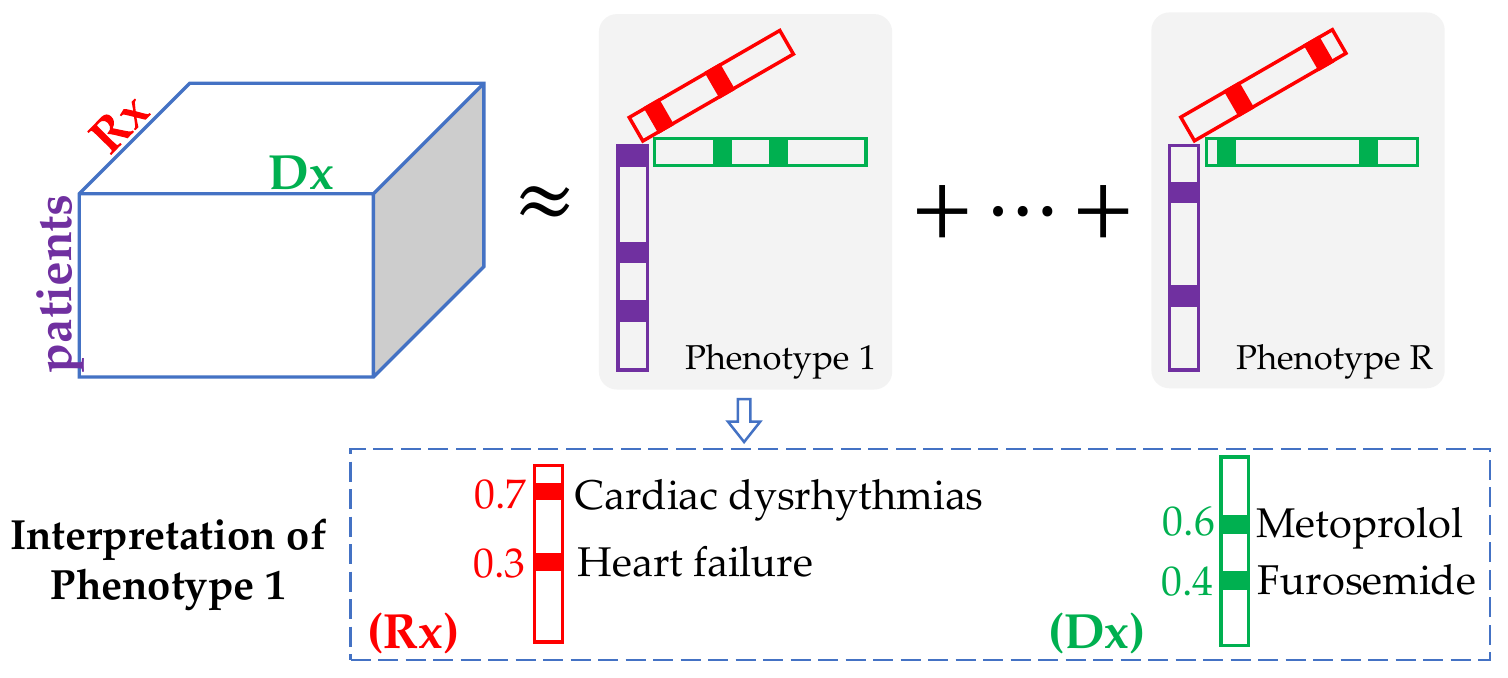}
    \caption{Non-negative Tensor Factorization for Computational Phenotyping: Each resulting rank-one tensor is interpreted as a phenotype, where its entries with non-zero values are extracted as the definition of the phenotype. For example, phenotype 1 consists of two diagnoses: \textit{Cardiac dysrhythmias} and \textit{heart failure}, and two medications: \textit{Metoprolol} and \textit{Furosemide}. ``Dx'' denotes diagnosis and ``Rx'' denotes medications.}\label{fig:ntf_phenotyping}
\end{figure}

The final step is to interpret the latent factors. Fig.~\ref{fig:ntf_phenotyping} illustrates how the phenotypes are extracted from the non-negative factor matrices. Essentially, the items with positive values in each mode in the $r^{th}$ rank-one tensor are extracted as the definition of the $r^{th}$ phenotype.

\section{Proposed Framework}\label{sec:model}
We first present the building block of our framework, the \textbf{H}idden \textbf{I}nteraction \textbf{T}ensor \textbf{F}actorization (HITF) model, which aims at discovering the hidden correspondence between two modalities when the latent interaction tensor is not observed. Then we introduce our cHITF framework, which ensembles more modalities in a flexible and easy-to-interpret way.

\subsection{HITF: Hidden Interaction Tensor Factorization}
\begin{figure}
    \centering
    \includegraphics[width=0.93\columnwidth]{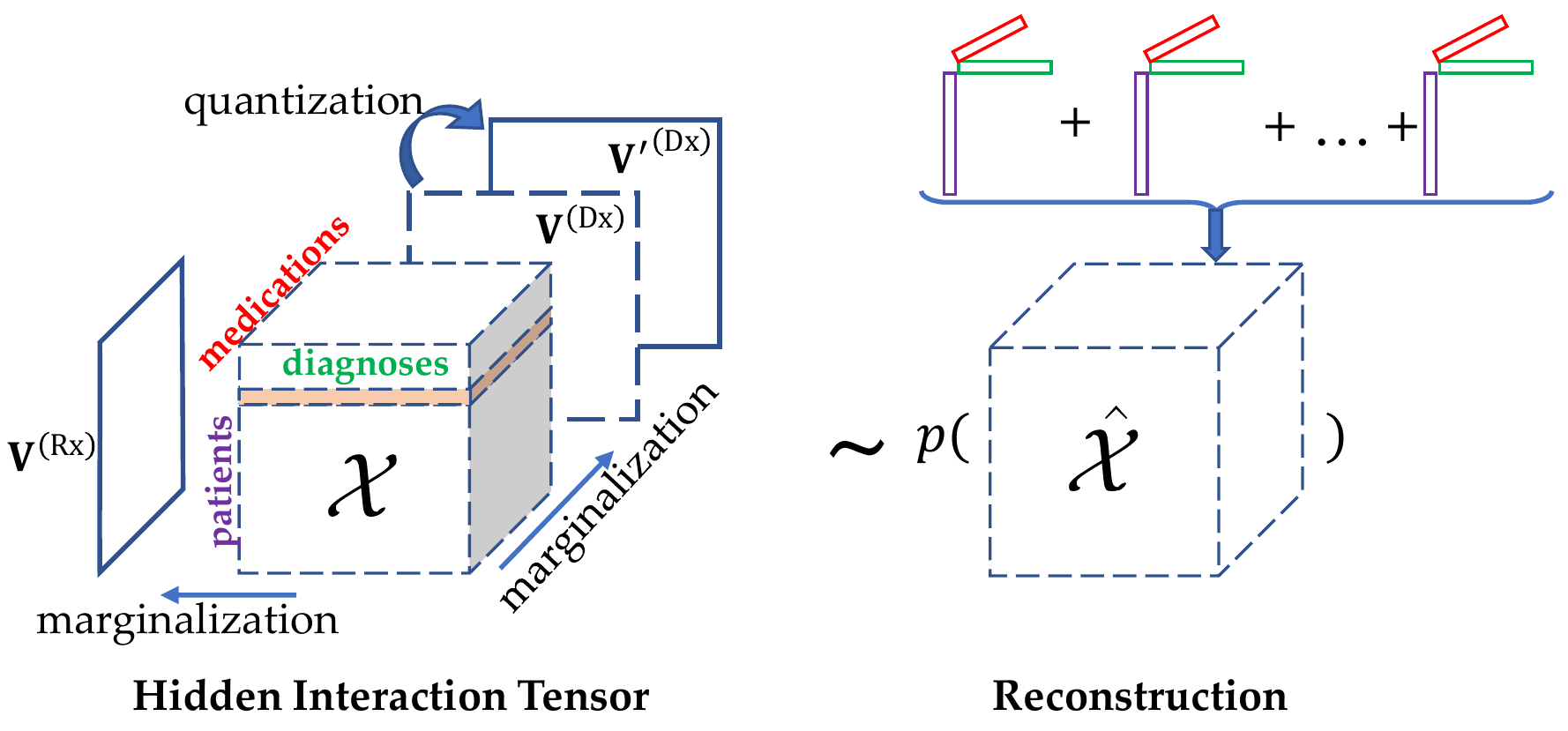}
    \caption{The building block: Hidden Interaction Tensor Factorization (HITF), illustrated in a third-order example with medications and diagnoses. Only the marginalization along the medication mode and the diagnoses mode are known, leaving the interactions totally unobserved. The hidden interaction tensor are assumed to be drawn from some distribution $p$ parameterized by the CP factorization of the hidden interaction tensor.}
    \label{fig:hitf}
\end{figure}

HITF aims at discovering the unobserved correspondence between items from different modalities given the marginalized observations~\cite{yin2018joint}. Let $\mathcal{V}=\{\mathbf{V}^{(n)}\in \mathbb{R}^{I_s \times I_n}\}_{n=1}^N$ denotes a set of $N$ observation matrices, each of which corresponds to a specific modality. The observation matrices share the first dimension with size $I_s$. We assume that the observation matrices are generated by marginalizing an unobserved high-order tensor that describes the inter-modality interactions. As an illustrative example, Fig.~\ref{fig:hitf} depicts the HITF model with two modalities: the medication prescriptions and the diagnosis codes in EHR. We observe two matrices: the \textit{patient-by-medication} matrix $\mathbf{V}^{(1)}$, and the \textit{patient-by-diagnosis} matrix $\mathbf{V}^{(2)}$, recording the medications being prescribed and the diagnosis codes being assigned to each patient, respectively. It is reasonable to assume that there exists some correspondence between this two modalities --- the medications are prescribed to the patients in response to some of the diagnoses. Therefore, we assume that there is a high-order hidden interaction tensor $\mathcal{X}$ with $N+1$ modes, describing the inter-modality correspondence, and the observation matrices $\mathcal{V}$ are obtained by marginalizing the hidden interaction tensor as shown in Eq.~(\ref{eq:marginalization}).

Similar to the ordinary non-negative CP factorization, we factorize the hidden interaction tensor into a set of latent factor matrices $\mathcal{U}=\{\mathbf{U}^{(s)}\} \cup \{\mathbf{U}^{(n)} \}_{n=1}^N$, where $\mathbf{U}^{(s)} \in \mathbb{R}^{I_s \times R}$ is associated with the shared dimension of the observation matrices (\textit{e.g.} the patient mode), $\mathbf{U}^{(n)} \in \mathbb{R}^{I_n \times R}$ is associated with the $n^{th}$ mode, and $R$ is the number of latent factors, \textit{i.e.}, phenotypes. We denote the reconstruction from the latent factors as $\mathcal{\hat{X}}$. Following the CP factorization, the entries of the hidden interaction tensor $\mathcal{X}$ are assumed to be drawn from some distribution (\textit{e.g.} Poisson or Gaussian) with the mean being the reconstructed tensor $\hat{\mathcal{X}}$, \textit{i.e.},
\begin{equation}\label{eq:some_distribution}
  x_{\mathbf{i}} \sim p(\hat{x}_{\mathbf{i}}, \boldsymbol{\theta}),
\end{equation}
where $\mathbf{i}$ denotes the index of the tensor entry and $\boldsymbol{\theta}$ is the set of parameters %
of the underlying distribution of the hidden interaction tensor.

In ordinary CP factorization, the factors can be estimated via minimizing the reconstruction error or maximizing the likelihood of the input tensor. However, in our setting, we only observe the marginalization of the high-order interaction tensor. Therefore, we solve for the factors by maximizing the likelihood of the marginalizations, instead of that of the tensor itself. To this end, we first apply the same marginalization to the reconstruction and obtain the marginalization along the $n^{th}$ dimension $\hat{\mathbf{V}}^{(n)}$ as follows:
\begin{equation}\label{eq:marginalize_reconstuction}
  \hat{\mathbf{V}}^{(n)} = \mathbf{U}^{(s)} \prod_{k\neq n} \diag\left(\mathbf{e}^\intercal\mathbf{U}^{(k)}\right) {\mathbf{U}^{(n)}}^\intercal.
\end{equation}
Then we derive the likelihood of the observation matrices. Finally, we can apply the projected gradient descent to solve for the factors. Here, we present the derivation of the likelihood of the marginalized observations under two commonly used distributions: Poisson distribution for counting data and Gaussian distributions for real-valued data.

\subsubsection{Poisson Distribution}
Poisson distributions are naturally used to model counting data, and are parameterized solely by their mean, and the sum of independent Poisson-distributed random variables follows another Poisson distribution where the mean is the sum of the parameters of the composing Poisson distributions. Therefore, we have:
\begin{equation}\label{eq:poisson_sum}
  \mathbf{V}^{(n)} \sim \pois(\mathbf{\hat{V}}^{(n)})
\end{equation}
The log likelihood of the observation $\mathbf{V}^{(n)}$ is then given by:
\begin{align}\label{eq:loglikelihood_poisson}
  \mathcal{L}_{\text{Poisson, integer}}\left(\mathbf{V}^{(n)}\right)
  = &{} \sum_{i,j}\log\left( p\left(v_{ij}^{(n)} \Big\rvert \{\mathbf{U}^{(n)}\}_{n=1}^N\right)\right) \nonumber\\
   = &{} \sum_{i,j} \left\{-\hat{v}_{ij} + v_{ij}\log( \hat{v}_{ij} )\right\} + c,
\end{align}
where $c$ is a constant.

In practice, for some modalities, the observation is binary instead of concrete countings, for example the entries of the \textit{patient-by-diagnosis} matrix take value of one if the corresponding diagnosis code is present in the patient's records, zero otherwise. Directly fitting the binary observation using the above Poisson distribution is obviously undesirable. Inspired by the related work in binary matrix factorization~\cite{hsieh2015pu}, we assume that the binary observations are generated via a deterministic quantization process, \textit{i.e.},
\begin{equation}\label{eq:quantization}
  {v^\prime}^{(n)}_{ij} = \mathds{1}\left(v_{ij}^{(n)} > 0\right),
\end{equation}
where $\mathds{1}(\cdot)$ is the indicator function. Thus, it is obvious that the entry of the binary observation matrix ${v^\prime}^{(n)}_{ij}$ follows a Bernoulli distribution with its mean being the probability of ${v}^{(n)}_{ij}$ larger than zero. Formally, we have:
\begin{equation}\label{eq:poisson_quantization_values}\small
  \begin{split}
  \Pr\left({v^\prime}_{ij}^{(n)}=1\right) & = \Pr\left(v_{ij}^{(n)} > 0\right) = 1 - \Pr\left(v_{ij}^{(n)}=0\right) \\
       & = 1 - \exp\left(-{\hat{v}_{ij}^{(n)}}\right) \frac{\left(\hat{v}_{ij}^{(n)}\right)^0}{0!}.
  \end{split}
\end{equation}

We can reorganize the above equation into matrix form and obtain:
\begin{equation}\label{eq:poisson_quantization}
  {\mathbf{V}^\prime}^{(n)} \sim \bern\left(1-\exp\left( -\hat{\mathbf{V}}^{(n)} \right)\right).
\end{equation}

The log likelihood of binary observations can then be derived as below:
\begin{align}\label{eq:loglikelihood_poisson_bin}\small
  &\mathcal{L}_{\text{Poisson, binary}}\left({\mathbf{V}^\prime}^{(n)}\right) \nonumber\\
  = &{} \sum_{i,j}\left({v^\prime}^{(n)}_{ij}\log\left(\exp\left({\hat{v^\prime}}^{(n)}_{ij}\right)-1\right) - {\hat{v^\prime}}^{(n)}_{ij} \right).
\end{align}

\subsubsection{Gaussian Distribution}
Gaussian distributions are often preferred when the observations are real-valued data. Similar to \cite{xiong2010temporal}, we consider the following Gaussian distribution for the hidden interaction tensor:
\begin{equation}\label{eq:gaussian}
  x_{\mathbf{i}} \sim \mathcal{N}(\hat{x}_{\mathbf{i}}, \sigma^2),
\end{equation}
where $\sigma^2$ is the variance of the Gaussian distributions, and is a hyper-parameter shared for all entries of the hidden interaction tensor. Summing up multiple Gaussian distributions yields another Gaussian distribution with its mean and variance being the sum of that of its composing Gaussian distributions, which gives:
\begin{align}\label{eq:gaussian_sum}\small
  \mathbf{V}^{(n)} &\sim \mathcal{N}\left(\mathbf{\hat{V}}^{(n)}, \left(\sum_{k\not=n}{I_k}\right)\sigma^2\right).
\end{align}

The log likelihood of the marginalized observation with the hidden interaction tensor following Gaussian distribution then can be computed by:
\begin{align}\label{eq:loglikelihood_gaussian}\small
    & \mathcal{L}_{\text{Gaussian, real-value}} \left( \mathbf{V}^{(n)} \right)  \nonumber\\
  = & -\frac{1}{2} \sum_{i, j}\left\{ \log(2\pi t_n \sigma^2) +\frac{1}{t_n\sigma^2} (v_{ij} - \hat{v}_{ij})^2\right\},
\end{align}
where $t_n=\sum_{k\not=n} I_k$.

For binary marginalized observations, we apply the same quantization process as in Eq.~(\ref{eq:quantization}) and obtain the probability of ${v^\prime}_{ij}$ taking value of one as below:
\begin{equation}\label{eq:gaussian_quantization_prob}\small
  p_{ij} %
        = 1 - \Pr\left(\hat{v}_{ij}^{(n)} \leq 0\right) = \frac{1}{2} - \frac{1}{2}\operatorname{erf}\left( -\frac{\hat{v}^{(n)}_{ij}}{\sqrt{2t_n}\sigma} \right),
\end{equation}
where $\Phi(\cdot)$ is the cumulative distribution function (CDF) of a standard Gaussian distribution, and $\operatorname{erf}(\cdot)$ denotes its error function given by:
\begin{equation}\label{eq:erf}
  \operatorname{erf}(x) = \frac{2}{\sqrt{\pi}}\int_{0}^{x} e^{-t^2} dt,
\end{equation}
which can be numerically approximated by its Taylor expansion with sufficient degree.

Then, it is straightforward to see that the log likelihood of the binarized value of the observation which follows Gaussian distribution can be written as:
\begin{align}\label{eq:loglikelihood_gaussian_binary}\small
  &\mathcal{L}_{\text{Gaussian, binary}} \left( {\mathbf{V}^\prime}^{(n)} \right) \nonumber\\
  = & \sum_{i,j} v^\prime_{ij}\log p_{ij} + (1-v^\prime_{ij}) \log (1-p_{ij}),
\end{align}
where $p_{ij}$ is defined in Eq.~(\ref{eq:gaussian_quantization_prob}).

\subsection{Towards Multiple Modalities: collective Hidden Interaction Tensor Factorization Framework}
\begin{figure*}
    \centering
    \includegraphics[width=0.9\linewidth]{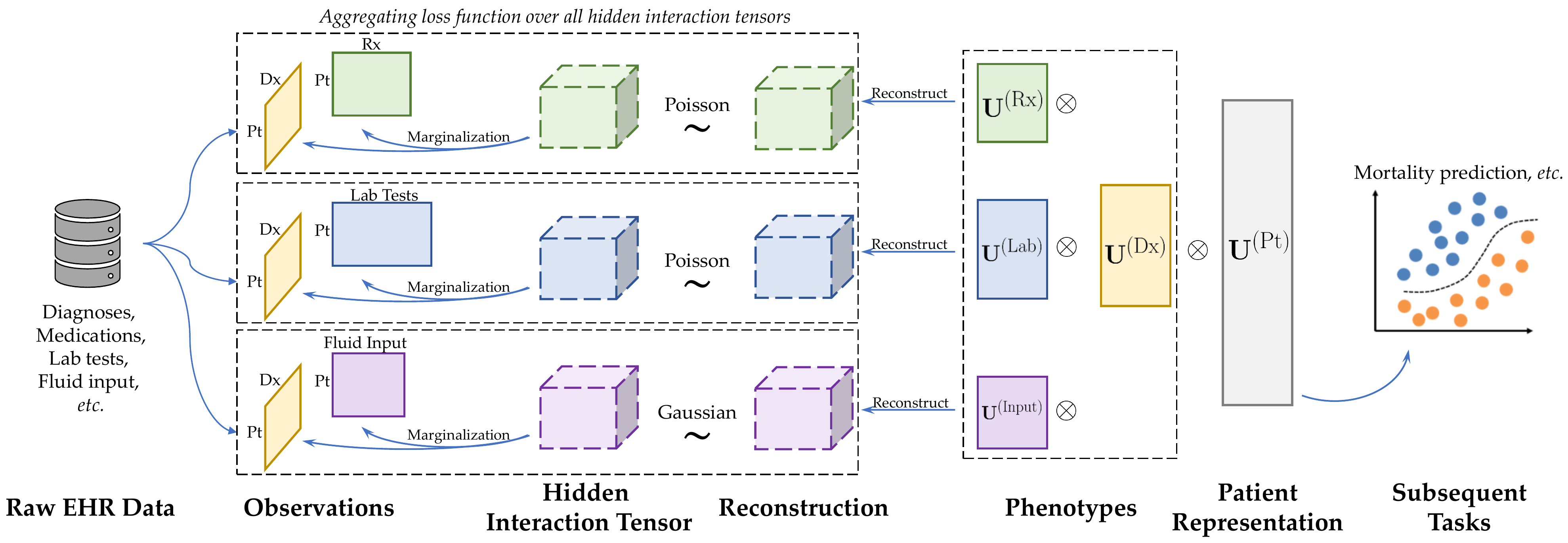}
    \caption{Overview of the cHITF framework, illustrated with an concrete example with three hidden interaction tensors and four modalities. Generally the number of interaction tensors and the modalities involved in each tensor can be determined by the nature of the data and the problem flexibly. Each hidden interaction tensor may follow different distributions, and marginalize to a subset of the observed matrices. The factor matrices (\textit{i.e.}, the phenotypes and the patient representations) are used to reconstruct the hidden interaction tensor, with the factor matrix corresponding to the same modality being forced to be the same. Finally, the learned patient representations can be used as features for subsequent tasks.}
    \label{fig:chitf_framework}
\end{figure*}

Although HITF is capable of modeling more than two modalities, directly applying it to the multi-modal EHR data for computational phenotyping can often be hindered by the difficulties in interpreting the correspondence among all the modalities as aforementioned. Thus, a more suitable way is to discover the correspondence of interest between each two modalities for better interpretation. For example, given the diagnoses, medications and lab tests, we can infer the correspondence between diagnoses and medications, and that between diagnoses and lab tests, respectively. %
To achieve this, we introduce the \textbf{c}ollective \textbf{H}idden \textbf{I}nteraction \textbf{T}ensor \textbf{F}actorization (cHITF) framework to simultaneously learn the phenotypes and infer the correspondence between modalities. We illustrate the framework in Fig.~\ref{fig:chitf_framework} with the following modalities involved:
\begin{itemize}[\setlength{\IEEElabelindent}{5pt} \setlength{\labelsep}{3pt}]
  \item \textbf{Diagnosis codes} are represented in a binary matrix denoted by $\mathbf{V}^{(\text{Dx})} \in \{0,1\}^{I_s, I_{\text{Dx}}}$ containing the diagnoses that are assigned to patients.
  \item \textbf{Medication prescriptions} are organized in a counting matrix $\mathbf{V}^{(\text{Rx})} \in \mathbb{Z}_{\geq0}^{I_s, I_{\text{Rx}}}$ containing the number of times each medications being prescribed to patients.
  \item \textbf{Abnormal lab tests} are organized similarly with the medications, \textit{i.e.}, $\mathbf{V}^{(\text{Lab})} \in \mathbb{Z}_{\geq0}^{I_s, I_{\text{Lab}}}$, containing the number of abnormal lab tests observed for each patient.
  \item \textbf{Input fluids} are represented in a non-negative real-valued matrix, $\mathbf{V}^{(\text{Input})} \in \mathbb{R}_{\geq0}^{I_s, I_{\text{Input}}}$. This matrix describes the total amount of fluids administrated to patients.
\end{itemize}
Among them, the modality of diagnosis plays a central role of the clinical data, in that the medications are prescribed and fluids are administrated mostly due to the diagnoses; the abnormal lab test results are metrics that reflect abnormal biochemical state caused by the pathology or diagnosis. Obviously, it is easier to interpret the correspondence between diagnoses and each one of the other three modalities. Therefore, we assume that there exist three hidden interaction tensors %
capturing the correspondence between the diagnoses and the medications, between the diagnoses and the lab tests, and between the diagnoses and the input fluids respectively.
We apply HITF to jointly solve each of them. The patient mode is the shared dimension (corresponding to the $\mathbf{U}^{(s)}$ factor matrix) in all the sub-problems, and the diagnosis mode is also shared due to our diagnosis centered design; hence, we force their corresponding factor matrices to be the same in the framework.

We now formalize the cHITF framework with an arbitrary number of modalities and latent interaction tensors. Given the set of marginalized observation matrices $\mathcal{V}$, we assume that there exists a set of $M$ hidden interaction tensors $\{ \mathcal{X}^{(m)}\}_{m=1}^M$, and each of them is marginalized to a subset of the observation matrices $\mathcal{V}_m = \{ \mathbf{V}^{(m,k)} \}_{k=1}^{K_m} \subset \mathcal{V}$, where $K_m$ is the number of composing modalities of the $m^{th}$ hidden interaction tensor (\textit{e.g.} $K_m=2, m=1,2,3$ for the example in Fig.~\ref{fig:chitf_framework}). We indicate the distribution of the $m^{th}$ hidden interaction tensor by $D_m \in \{ \text{Poisson}, \text{Gaussian} \}$, and the data type of the observation matrix $\mathbf{V}^{(m,k)}$ by $T_{m,k}\in \{ \text{integer}, \text{binary}, \text{real-value} \}$. The parameters to be learned are the latent factor matrices, \textit{i.e.}, $\mathcal{U}=\{\mathbf{U}^{(s)}\} \cup \{\mathbf{U}^{(n)} \}_{n=1}^N$. We reorganize the latent factors according to the hidden interaction tensors. Specifically, we associate a subset of the latent factors $\mathcal{U}_m=\{\mathbf{U}^{(s)}\} \cup\{\mathbf{U}^{(m,k)} \}_{k=1}^{K_m}\subset\mathcal{U}$ with the $m^{th}$ hidden interaction tensor. For example, a \textit{patient-diagnosis-medication} hidden interaction tensor is associated with the latent factors for patient, diagnosis and medication. Some modalities are involved in more than one hidden interaction tensors (\textit{e.g.} diagnosis). We force the latent factor matrices of each hidden interaction tensor that actually correspond to the same modality being the same. For instance, the latent factor matrix for diagnosis in the \textit{patient-diagnosis-medication} and \textit{patient-diagnosis-lab-test} hidden tensors are forced to be the same. Meanwhile, we allow the latent factors of different modalities for different hidden interaction tensors to be independent of each other. This yields the following optimization problem:
\begin{equation}\small
\begin{aligned}
\min_{\mathcal{U}} ~~~ & \sum_{m=1}^{M} \sum_{k=1}^{K_m} -\mathcal{L}_{D_m, T_{m,k}} \left(\mathbf{V}^{(m,k)} \mid \hat{\mathbf{V}}^{(m,k)} \right) + \Omega(\mathcal{U})\\
\textrm{s.t.} ~~~ & \mathcal{U}_m=\{\mathbf{U}^{(s)}\} \cup\left\{\mathbf{U}^{(m,k)} \right\}_{k=1}^{K_m} \subset \mathcal{U} ~~~~ m=1,\dots,M,\\
                    & \hat{\mathbf{V}}^{(m,k)} = {\mathbf{U}}^{(s)}\prod_{k^\prime\neq k}\diag\left( \mathbf{e}^\intercal\mathbf{U}^{(m,k^\prime)} \right){\mathbf{U}^{(m,k)}}^\intercal ~~\forall (m,k),\\
                    & \mathbf{U}^{(m, k)} = \mathbf{U}^{(m^\prime, k^\prime)} \\
                    &~~~\forall (m,k,m^\prime,k^\prime)\in \left\{ (m,k,m^\prime,k^\prime) \mid \mathbf{V}^{(m, k)} = \mathbf{V}^{(m^\prime, k^\prime)} \right\},   \\
                    & \mathbf{U}^{(s)} \geq \mathbf{0},
                    ~~~ \mathbf{U}^{(m, k)} \geq \mathbf{0} ~~~~~\forall (m,k),
\end{aligned}\label{eq:chitf_optimization}
\end{equation}
where $\Omega(\mathcal{U})$ is the regularization imposed on the latent factors.

\subsection{Interpretability-Enhancing Regularizations}
Additional regularizations can be incorporated to further improve the interpretability of the learned phenotypes. We incorporate two of them: the elastic net regularization for sparsity and the angular regularization for diversity.

\subsubsection{Promoting Sparsity: Elastic Net Regularization}
To encourage sparse latent factors, we incorporate the elastic net regularization~\cite{zou2005regularization}, which takes the convex combination of the $\ell_1$ norm and $\ell_2$ norm of the parameter vector. %
Formally, we define:
\begin{equation}\label{eq:elastic_net}
  \Omega_1 = \gamma \sum_{n=1}^{N} \sum_{r=1}^{R} \left( \alpha \lVert \mathbf{u}^{(n)}_r \rVert_2^2 + (1-\alpha) \lVert \mathbf{u}^{(n)}_r \rVert_1 \right),
\end{equation}
where $\gamma$ and $\alpha$ are hyper-parameters controlling the strength of the overall regularization and the $\ell_1$ term.

\subsubsection{Promoting Diversity: Pairwise Angular Regularization}
It is also of critical importance to derive diverse phenotypes, instead of having a set of phenotypes that are very similar to each other. Angular constraint~\cite{xie2017learning} was found promising for encouraging diversity in latent variable models, and a variant was also introduced in discovering diversified phenotypes from EHR data~\cite{henderson2017granite} by penalizing the factors which has pairwise cosine similarity above a certain threshold. We follow these works to incorporate the pairwise angular regularization defined as below:
\begin{equation}\label{eq:angular}\small
  \Omega_2 = \beta \sum_{n=1}^N \sum_{r=2}^{R} \sum_{r^\prime=1}^{r-1} \left( \max\left\{ 0, \frac{(\mathbf{u}^{(n)}_r)^\intercal \mathbf{u}^{(n)}_{r^\prime}}{\lVert \mathbf{u}^{(n)}_r \rVert_2 \lVert \mathbf{u}^{(n)}_{r^\prime} \rVert_2} - \theta_n \right\} \right)^2,
\end{equation}
where $\beta$ is a hyper-parameter controlling the regularization strength. $\theta_n$ is a hyper-parameter defining the angular penalization threshold. Factors having pairwise cosine similarity larger than this threshold will be penalized.

\subsection{Learning Algorithms}
We estimate the parameters by solving the optimization problem (\ref{eq:chitf_optimization}) via block coordinate descent (BCD) approach~\cite{xu2013block}. Specifically, we alternate between the latent factors in $\mathcal{U}$ and update each of them with all others fixed. We use projected gradient descent to update the latent factors. Such block coordinate gradient projection method has been shown to enjoy a global sublinear rate of convergence~\cite{beck2013convergence}. We summarize the optimization procedure in Appendix A, available in the supplemental materials. Our implementation is publicly available at \url{https://github.com/jakeykj/cHITF}.

\section{Experiments}\label{sec:expts}

\subsection{Datasets}
We evaluate our model using two open-source, large-scale, de-identified and ICU patients related EHR datasets, MIMIC-III~\cite{mimiciii} and eICU~\cite{pollard2018eicu}. MIMIC-III contains data related to over forty thousand patients who stayed in the intensive care units at Beth Israel Deaconess Medical Center between 2001 and 2012, and eICU is a multi-center ICU database covering ICU admissions across the United States. The two datasets are different from outpatient datasets since patients in ICU are mostly with severe and life-threatening illnesses or injuries, and are likely to have multiple complications. For example, each patient has 11 diagnoses per clinical visit on average in MIMIC-III. Moreover, they contain considerably many medications which are used not for treating specific diseases, such as pain relievers, making the diagnosis-medication correspondence more obscure.

To avoid over sparsity, we group the diagnosis codes according to the first three digits of their corresponding ICD-9 codes and only use the items in each modality that appeared in the records of at least 5\% of the patients. We also exclude the base type medications such as D5W. For MIMIC-III, we obtain a dataset containing 22,080 patients with 160 distinct diagnosis codes, 177 distinct medications, and 150 distinct lab tests (hereinafter referred to as ``full dataset''). For eICU, we use the modality of treatment instead of lab test, and extract a subset containing 10,000 admissions and use the top 100 most frequent items for each modality.

\subsection{Baselines and Hyperparameter Tuning}
We compare our model against some of the following baselines for different tasks:
\begin{itemize}[\setlength{\IEEElabelindent}{5pt} \setlength{\labelsep}{3pt}]
  \item \textbf{CP-APR}%
  \cite{chi2012tensors} is a widely used Poisson NTF model.

  \item \textbf{Marble}%
  \cite{ho2014marble} is a computational phenotyping model based on CP-APR with a bias tensor to account for the baseline characteristics of the overall population.

  \item \textbf{Rubik}%
  \cite{wang2015rubik} is a non-negative tensor factorization model based on a quadratic loss between the input tensor and the reconstruction; a pairwise constraint is introduced to encourage diversity of phenotypes.  %

  \item \textbf{Granite}%
  \cite{henderson2017granite} is a variant of Marble with a
  regularization as in Eq.~(\ref{eq:angular}) to encourage diversity of phenotypes.

  \item \textbf{SiCNMF}%
  \cite{arxiv2016cmf,gunasekar2016mining} is a collective matrix factorization model developed for computational phenotyping. This model is the only one that does not assume that co-occurring items correspond to each other equally.
  \end{itemize}

Similar to a recent work~\cite{hong2020generalized}, we choose the distribution for each hidden interaction tensor based on their data types. Specifically, we use Poisson distribution for medications, lab tests (in MIMIC-III) and treatments (in eICU) as they are recorded as counts. We use Gaussian distribution for input fluid volumes, because they are recorded as real values. An empirical validation and more discussion can be found in the supplementary materials.

We tune the hyperparameters using grid search. Only 50\% of the data is used due to the efficiency reason. To balance the representational power and interpretability, we first pick the hyperparameter combinations that yield top-three highest PR-AUC scores for the in-hospital prediction task.
The hyperparameters are then determined by manually inspecting the quality of inferred phenotypes and correspondence. The only exception is the variance in the Gaussian distribution ($\sigma^2$), which is tuned purely based on prediction performance. Finally, we set $\gamma$ to 1e-5, $\alpha$ to 0.7, $\beta$ to 1, $\theta_n$ to 0.5 for all $n$, and $\sigma^2$ to 1e-9. The sensitivity analysis can be found in Appendix B. The hyperparameters of all the baselines are also carefully tuned by grid search. We set the number of phenotypes to 50 for all models, unless otherwise specified.

\subsection{HITF Discovers Correspondence with Significantly Improved Clinical Meaningfulness}
We first evaluate the inter-modal correspondence inferred by HITF, the building block of the proposed framework. As previously stated, the correspondence among more than two modalities are difficult to be interpreted, so we separately construct a \textit{patient-diagnosis-medication} tensor and a \textit{patient-diagnosis-labtest} tensor for baselines. We exclude SiCNMF  %
as it does not model the high-order interactions. %

\subsubsection{Correspondence Extraction}\label{sec:correspondence}
The inter-modal correspondence matrix of an individual patient with index $i$ can be obtained by fixing the patient index of the \textit{reconstructed hidden interaction tensor} at $i$. As it is infeasible to examine the quality of inferred correspondence for each individual patient, we focus on the average correspondence over the population with the same diagnosis. For instance, regarding the diagnosis-medication correspondence, we extract all the patients with the index of a diagnosis of interest $j$ as the base population, and accumulate the inferred interaction tensor along the patient dimension over the base population, resulting in a diagnosis-by-medication average correspondence matrix $\mathbf{C}$. We extract the $j^\text{th}$ row and normalize it using its $\ell_1$ norm. The normalized value then can be interpreted as the probability of each medication being corresponding to the selected diagnosis. We define the entry $c_{ij}$ as the \textit{correspondence score} of the $j^\text{th}$ medication to the $i^\text{th}$ diagnosis. We follow the same procedure to extract the correspondence between diagnoses and lab tests.

\subsubsection{Evaluation Methods}

\begin{figure}
    \centering
    \includegraphics[width=0.95\columnwidth]{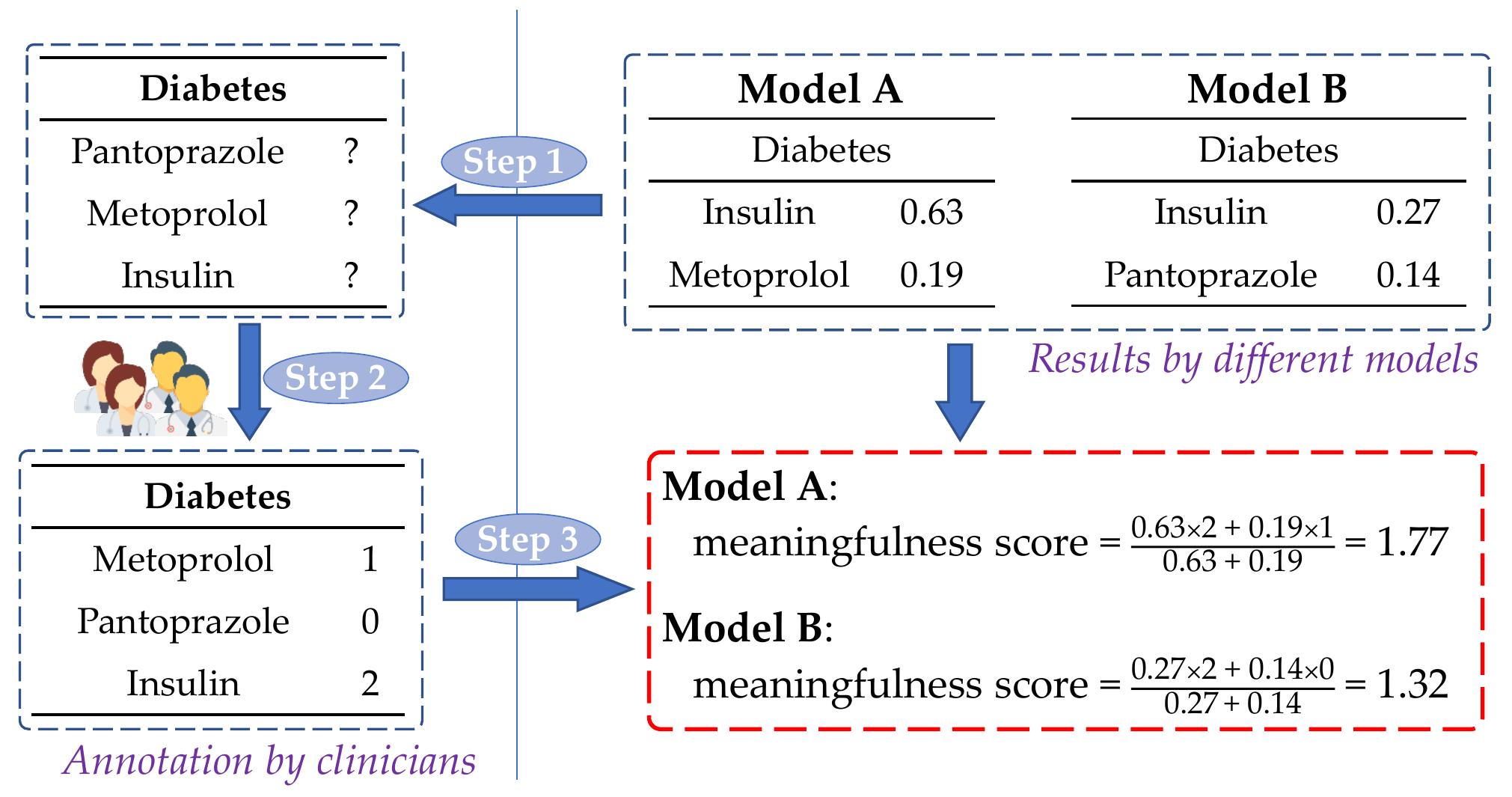}
    \caption{The process of quantitatively evaluating the inter-modal correspondence, illustrated with an example of medications for diabetes. \textbf{Step 1}: We gather all the correspondence items from all models and randomly shuffle them. \textbf{Step 2}: We present the items to clinicians for blind scoring. There are three options: 0 for ``not clinically relevant'', 1 for ``possibly clinically relevant'', and 2 for ``clinically relevant''. \textbf{Step 3}: We compute the quality score of each model by taking weighted sum of the score given by the clinician weighted by the weighting produced by each model.}
    \label{fig:quality_score}
\end{figure}

We collaborate with a clinical expert to conduct the quantitative evaluation of the inferred correspondence, and the evaluation process is illustrated in Fig.~\ref{fig:quality_score}. As it is not feasible to evaluate the quality of the inferred correspondence for all diagnoses, we select 10 diagnoses with different frequencies in the dataset. For each diagnosis, we collect the top ten medications or lab tests inferred by each model and take the union of them. After randomly shuffling, we present them to the clinician and ask the clinician to annotate the inferred correspondence as 0 (not meaningful), 1 (possibly meaningful), or 2 (meaningful). The correspondence between a medication and a diagnosis can be either the medication being used for treating the diagnosis, or the diagnosis being caused by applying the medication. The evaluation process is ``blind'' in that the clinician is unaware of the model inferring the medications or lab tests.

We define a meaningfulness score to quantitatively measure the performance of each model. For each model and diagnosis, we first re-normalize the correspondence score of the top ten items obtained in Section~\ref{sec:correspondence}. Then we compute the weighted sum of the annotation scores given by the clinician with the weights being the normalized correspondence score. Formally, we define:
\begin{equation}
    \text{Meaningfulness Score}(j) = \frac{\sum_{j\in\mathcal{J}_i} c_{ij} z_{ij}}{\sum_{j\in\mathcal{J}_i} c_{ij}},
\end{equation}
where $i$ is the index of the target diagnosis, $\mathcal{J}_i$ is the index set of the top ten items for the $i^\text{th}$ diagnosis, and $z_{ij}$ denotes the clinician annotation score of the $j^\text{th}$ item to the $i^\textrm{th}$ diagnosis. The meaningfulness score takes range between 0 and 2 (inclusive), and higher score indicates more meaningful items inferred in the top ten corresponding items.

\subsubsection{Results and Discussions}

\begin{figure*}
\centering
    \subfloat[Meaningfulness score of medications inferred by different models]{\includegraphics[width=0.9\textwidth]{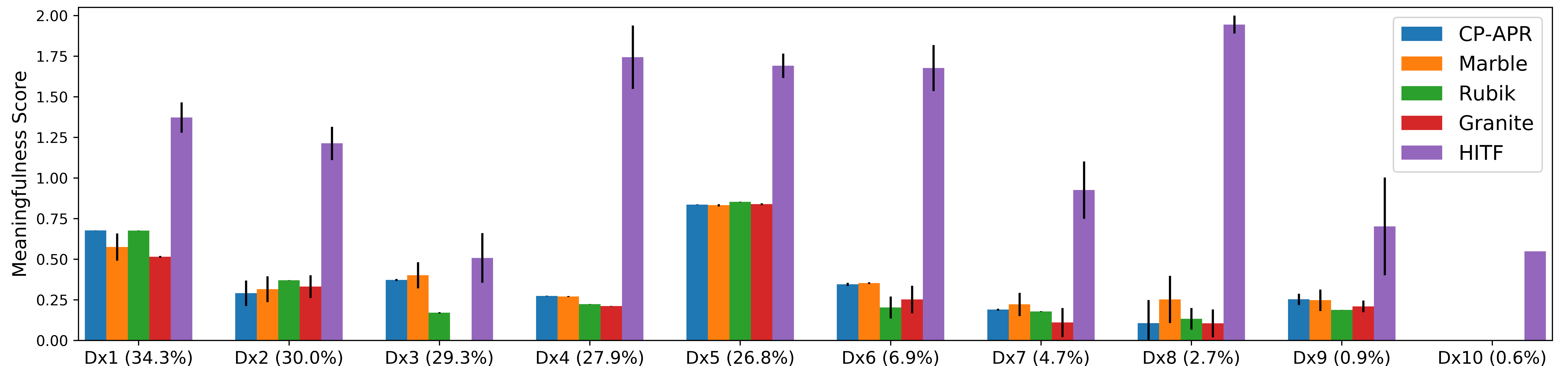}\label{fig:corrs:dxrx}}
    \vfil
    \subfloat[Meaningfulness score of lab tests inferred by different models]{\includegraphics[width=0.9\textwidth]{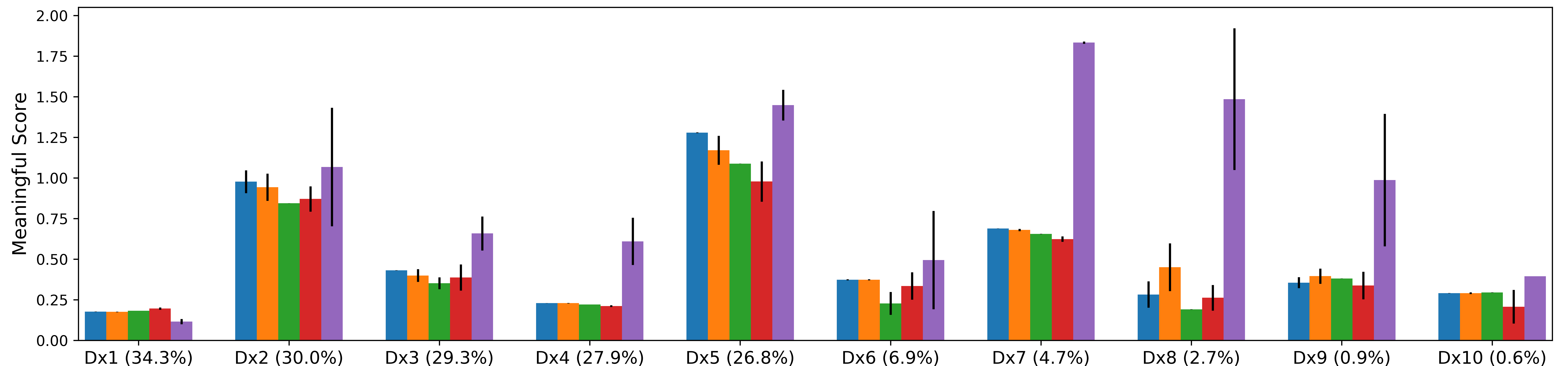}\label{fig:corrs:dxlab}}
    \caption{The meaningfulness score of medications and lab tests inferred by HITF and baselines. The inter-modal correspondence inferred by HITF are significantly better than baselines. ``Dx'' is the abbreviation of diagnosis, and the percentage inside the parentheses denotes the frequency of the corresponding diagnosis. The ten diagnoses listed in the figure are as follows. \textbf{Dx1}: Cardiac dysrhythmias, \textbf{Dx2}: Heart failure, \textbf{Dx3}: Other forms of chronic ischemic heart disease, \textbf{Dx4}: Diabetes mellitus, \textbf{Dx5}: Disorders of fluid electrolyte and acid-base balance, \textbf{Dx6}: Bacterial infection in conditions classified elsewhere and of unspecified site, \textbf{Dx7}: Iron deficiency anemias, \textbf{Dx8}: Chronic bronchitis, \textbf{Dx9}: Arterial embolism and thrombosis, \textbf{Dx10}: Symptoms involving nervous and musculoskeletal systems.}
    \label{fig:corrs}
\end{figure*}

\begin{table*}
\renewcommand{\arraystretch}{1.1}
  \centering
  \caption{Examples of Diagnosis-Medication Correspondence Inferred by HITF and Rubik}\label{tab:correspondence_examples_dxrx}
\begin{tabular}{cl}\toprule

& \multicolumn{1}{c}{\textbf{Heart Failure}}\\\hline
HITF  &  \begin{tabular}[t]{@{}l@{}} \textcolor{red}{\textbf{Furosemide (0.56)}}; ~{Potassium Chloride (0.23)}; {Magnesium Sulfate (0.03)}; ~{Prednisone (0.02)}.\end{tabular}\\
Rubik  &  \begin{tabular}[t]{@{}l@{}} {Potassium Chloride (0.02)}; ~{Acetaminophen (0.02)}; {Insulin (0.02)}; ~{Magnesium Sulfate (0.02)}; ~\textcolor{red}{\textbf{Furosemide (0.02)}}.\end{tabular}\\\toprule

& \multicolumn{1}{c}{\textbf{Diabetes Mellitus}}\\\hline
HITF  &  \begin{tabular}[t]{@{}l@{}} \textcolor{red}{\textbf{Insulin (0.88)}}; ~\textcolor{red}{\textbf{Insulin Human Regular (0.05)}}; \textcolor{red}{\textbf{Dextrose 50\% (0.01)}}; ~\textcolor{red}{\textbf{Metformin (0.01)}}.\end{tabular}\\
Rubik  &  \begin{tabular}[t]{@{}l@{}} {Acetaminophen (0.02)}; {Potassium Chloride (0.02)}; ~\textcolor{red}{\textbf{Insulin (0.02)}}; ~{Magnesium Sulfate (0.02)}; ~{Sodium Chloride 0.9\%  Flush (0.02)}.\end{tabular}\\\toprule

& \multicolumn{1}{c}{\begin{tabular}[c]{@{}l@{}} \textbf{Iron Deficiency Anemias} \end{tabular}} \\\hline
HITF  &  \begin{tabular}[t]{@{}l@{}} \textcolor{red}{\textbf{Pantoprazole Sodium (0.33)}}; ~Sodium Chloride 0.9\%  Flush (0.27); ~\textcolor{red}{\textbf{Pantoprazole (0.21)}}; ~Acetaminophen (0.09); ~Heparin (0.03)\end{tabular}\\
Rubik  &  \begin{tabular}[t]{@{}l@{}} {Acetaminophen (0.02)}; ~Potassium Chloride (0.02); ~Insulin (0.02); ~Sodium Chloride 0.9\%  Flush (0.02); ~\textcolor{red}{\textbf{Pantoprazole (0.02)}}\end{tabular}\\\toprule

&\multicolumn{1}{c}{\begin{tabular}[c]{@{}l@{}} \textbf{Arterial Embolism and Thrombosis} \end{tabular}} \\\hline
HITF  &  \begin{tabular}[c]{@{}l@{}} {Sodium Bicarb (0.36)}; ~\textcolor{red}{\textbf{Enoxaparin Sodium (0.33)}}; ~\textcolor{red}{\textbf{Isosorbide Mononitrate (0.22)}}; ~Tacrolimus (0.03); ~Mycophenolate Mofetil (0.01)\end{tabular}\\
Rubik & \begin{tabular}[c]{@{}l@{}} Potassium Chloride (0.02); ~{Acetaminophen (0.02)}; ~ ~Insulin (0.02); ~Magnesium Sulfate (0.02); ~\textcolor{red}{\textbf{Furosemide (0.02)}}\end{tabular}\\\bottomrule

\end{tabular}
\end{table*}

\begin{table*}
\renewcommand{\arraystretch}{1.1}
  \centering
  \caption{Examples of Diagnosis-Lab-Test Correspondence Inferred by HITF and Rubik}\label{tab:correspondence_examples_dxlab}
\begin{tabular}{cl}\toprule

& \multicolumn{1}{c}{\textbf{Heart Failure}}\\\hline
HITF &  \begin{tabular}[t]{@{}l@{}} {PT {[}B{]} (0.16)}; ~~\textcolor{red}{\textbf{Hematocrit {[}B{]} (0.14)}}; ~\textcolor{red}{\textbf{Hemoglobin {[}B{]} (0.13)}}; ~\textcolor{red}{\textbf{Red Blood Cells {[}B{]} (0.13)}}; ~{Urea Nitrogen {[}B{]} (0.11)}. \end{tabular}\\
Rubik &  \begin{tabular}[t]{@{}l@{}} {Red Blood Cells {[}B{]} (0.03)}; ~\textcolor{red}{\textbf{Hemoglobin {[}B{]} (0.03)}}; ~\textcolor{red}{\textbf{Hematocrit {[}B{]} (0.03)}}; ~{Glucose {[}B{]} (0.03)}; {Urea Nitrogen {[}B{]} (0.03)}; \end{tabular}\\\toprule

& \multicolumn{1}{c}{\textbf{Diabetes Mellitus}}\\\hline
HITF &  \begin{tabular}[t]{@{}l@{}}\textcolor{red}{\textbf{Glucose {[}B{]} (0.21)}}; ~~\textcolor{red}{\textbf{Urea Nitrogen {[}B{]} (0.16)}}; ~Hemoglobin {[}B{]} (0.09); ~Creatinine {[}B{]} (0.08); ~\textcolor{blue}{\textit{Red Blood Cells {[}B{]} (0.08)}}. \end{tabular}\\
Rubik &  \begin{tabular}[t]{@{}l@{}} {Red Blood Cells {[}B{]} (0.03)}; ~{Hemoglobin {[}B{]} (0.03)}; ~{Hematocrit {[}B{]} (0.03)}; ~\textcolor{red}{\textbf{Glucose {[}B{]} (0.03)}}; {White Blood Cells {[}B{]} (0.03)}. \end{tabular}\\\toprule

& \multicolumn{1}{c}{\begin{tabular}[c]{@{}l@{}} \textbf{Iron Deficiency Anemias} \end{tabular}} \\\hline
HITF &  \begin{tabular}[t]{@{}l@{}}\textcolor{red}{\textbf{Transferrin {[}B{]} (0.26)}};  ~\textcolor{red}{\textbf{Total Iron Binding Capacity {[}B{]} (0.25)}}; ~\textcolor{red}{\textbf{Iron {[}B{]} (0.19)}}; ~\textcolor{red}{\textbf{Ferritin {[}B{]} (0.17)}}; ~Vitamin B12 {[}B{]} (0.04) \end{tabular}\\
Rubik &  \begin{tabular}[t]{@{}l@{}} \textcolor{red}{\textbf{Red Blood Cells {[}B{]} (0.03)}}; ~\textcolor{red}{\textbf{Hemoglobin {[}B{]} (0.03)}}; ~\textcolor{red}{\textbf{Hematocrit {[}B{]} (0.03)}}; ~{Glucose {[}B{]} (0.03)}; {White Blood Cells {[}B{]} (0.03)}; \end{tabular}\\\toprule

&\multicolumn{1}{c}{\begin{tabular}[c]{@{}l@{}} \textbf{Arterial Embolism and Thrombosis} \end{tabular}} \\\hline
HITF &  \begin{tabular}[t]{@{}l@{}}\textcolor{red}{\textbf{PT {[}B{]} (0.77)}};  ~{Sodium, Whole Blood {[}B{]} (0.06)}; ~\textcolor{blue}{\textit{Lactate {[}B{]} (0.05)}}; ~{Chloride, Whole Blood {[}B{]} (0.03)}; ~Vancomycin {[}B{]} (0.03) \end{tabular}\\
Rubik &  \begin{tabular}[t]{@{}l@{}} {Red Blood Cells {[}B{]} (0.03)}; ~{Hematocrit {[}B{]} (0.03)}; ~{Hemoglobin {[}B{]} (0.03)}; ~{Glucose {[}B{]} (0.03)}; \textcolor{red}{\textbf{PT {[}B{]} (0.03)}}; \end{tabular}\\\bottomrule

\end{tabular}
\end{table*}

We report the meaningfulness scores of the medications and the lab tests to ten diagnoses obtained by each model in Fig.~\ref{fig:corrs}. The percentage inside the parentheses after the diagnosis index indicates the frequency of that diagnosis in the dataset. We run each model five times and report the average value in the bar graph and the standard deviation by the error bar. As shown in Fig.~\ref{fig:corrs:dxrx}, HITF outperforms all baselines consistently. %
More interestingly, HITF is especially advantageous for less frequent diseases. For example, all baselines fail to correctly infer any medications corresponding to Dx10 (\textit{Symptoms involving nervous and musculoskeletal systems}), yet HITF can discover its corresponding medications that are annotated to be clinically meaningful. Fig.~\ref{fig:corrs:dxlab} shows that HITF also outperforms the baselines for inferring the correspondence between the diagnoses and the lab tests, although by a smaller margin. It is also worth noting that the superiority of HITF for less frequent diagnoses is consistent with what we observe in Fig.~\ref{fig:corrs:dxrx}, \textit{e.g.}, HITF outperforms all baselines substantially for Dx7-9.

We list some inferred correspondence obtained by HITF and Rubik in Table~\ref{tab:correspondence_examples_dxrx} (medications) and Table~\ref{tab:correspondence_examples_dxlab} (lab tests) for further comparison.
The number following each item is the corresponding score inferred. Annotated by the clinician, items in red bold text are clinically meaningful, that in blue italic text are possibly meaningful, and the rest are not meaningful. As shown, the advantage of HITF over Rubik is twofold. First, HITF infers meaningful medications or lab tests that cannot be discovered by Rubik. For example, HITF discovered that the medication \textit{enoxaparin sodium} is corresponding to the diagnosis \textit{arterial embolism and thrombosis}, and the lab test \textit{transferrin [Blood]} is corresponding to \textit{iron deficiency anemias}. Second, HITF assigns much larger weights to the meaningful items, yet Rubik assigns almost all items even weights. For instance, HITF and Rubik both inferred the medication \textit{insulin} and the lab test \textit{glucose [Blood]} to be corresponding to \textit{diabetes mellitus}; however, HITF assigned corresponding scores of 0.88 and 0.21 to them, whereas Rubik assigned only 0.02 and 0.03 to them, respectively. The results obtained by other models also suggest similar conclusions, and more examples are included in the supplemental materials.

\subsection{cHITF Infers Clinically Relevant Phenotypes}\label{sec:phenotypes}
Inferring phenotypes that are clinically interpretable and relevant from multi-modal EHR data is our primary task. In this section, we evaluate the discovered phenotypes in two important aspects: the \textit{clinical relevance} and the \textit{diversity}.

We set the number of phenotypes to be 50, and run cHITF and the baselines to derive phenotypes. Since most of the baselines cannot simultaneously handle modalities with different distributions, we do not involve the modality of input events in this section. Most of the baselines are based on factorizing the input tensor, which has an exponential complexity with respect to the number of modes and makes them infeasible to converge in a reasonable time. For fair comparison, we shrink the size of the dataset by 50\%, and only use the top 80 most frequent items for each modality.

\subsubsection{Clinical Relevance}
The clinical relevance are evaluated by the clinical expert, who annotates each phenotype as either ``not clinically relevant'', ``possibly clinically relevant'', or ``clinically relevant''. As the annotation process is time-consuming and labor-intensive, it is not feasible to evaluate all the baselines. So we only compare against SiCNMF and Granite in terms of the clinical relevance. The former takes the same observation matrices as cHITF but ignoring the inter-modal interactions; the latter relies on the ``equal-correspondence'' assumption to construct the input tensor.

\begin{figure}
    \centering
    \includegraphics[width=0.9\columnwidth]{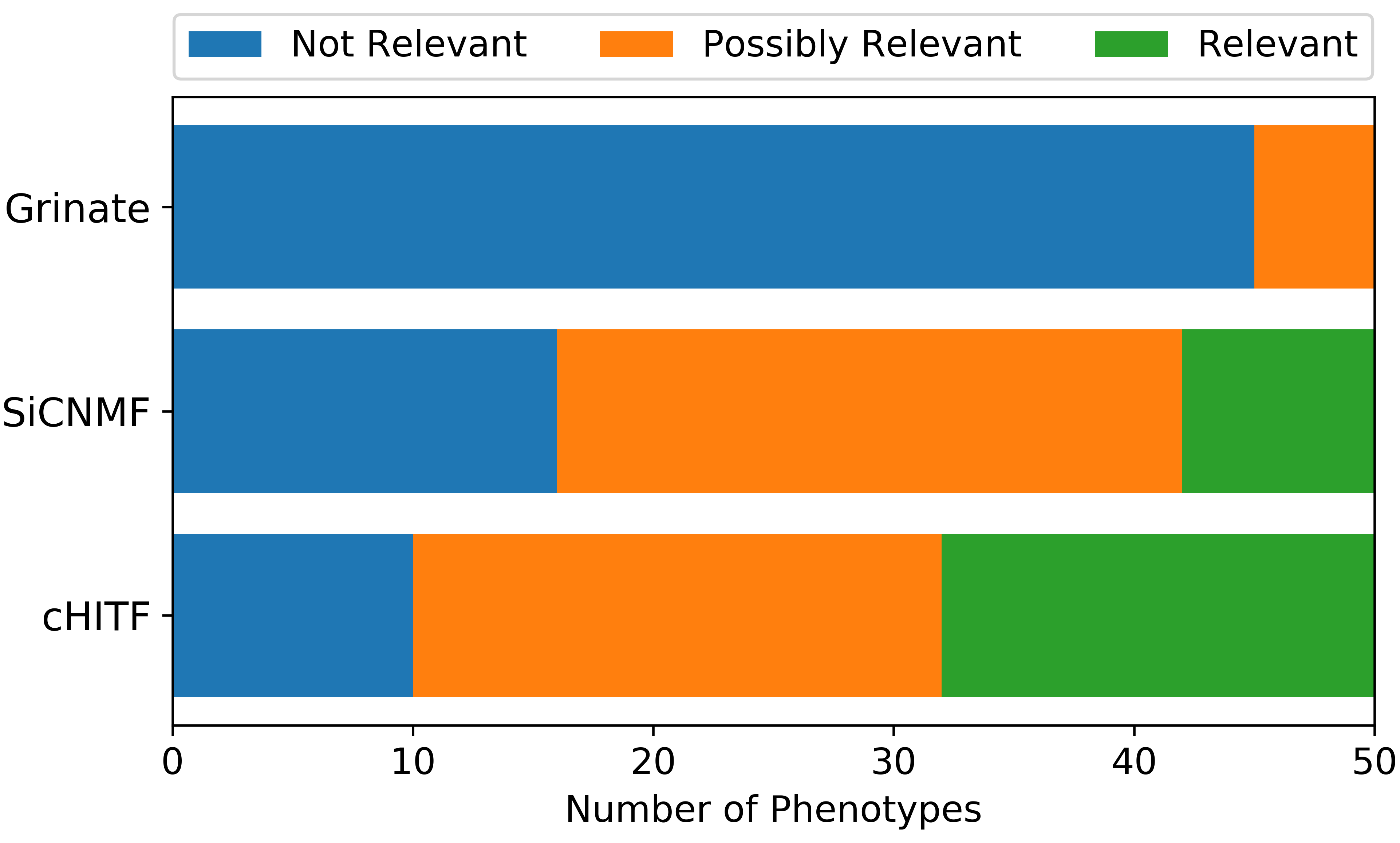}
    \caption{Quantitative evaluations of the clinical relevance of the phenotypes inferred. cHITF outperforms the baselines significantly by having 18 phenotypes annotated as relevant by the clinical expert.}
    \label{fig:phenotype_relevance}
\end{figure}
We summarize the clinical relevance of the inferred phenotypes in Fig.~\ref{fig:phenotype_relevance}. Our cHITF framework significantly outperforms Grinate and SiCNMF in that 18 and 22 out of 50 phenotypes are annotated by the clinical expert to be clinically relevant and possibly relevant, respectively. SiCNMF, by contrast, only generates 8 relevant phenotypes. SiCNMF takes the same input as cHITF, namely each matrix for one modality. However, it differs from our model significantly as it does not model the interactions between different modalities. Our framework explicitly takes the inter-modal interaction into consideration, leading to the impressive improvement of the phenotype quality. Another interesting observation is that Grinate fails to generate any phenotypes that are clinically relevant, and the majority of the generated phenotypes are annotated as irrelevant. The reason behind its failure is twofold. First, as described previously, Grinate relies on the pre-established inter-modal interactions that are constructed by the ``equal-correspondence'' assumption which leads to inevitable and massive noise in the input to the model. The second is related to the nature of the data we used: unlike the longitudinal records or outpatient data\footnote{Granite was originally developed and tested with a longitudinal EHR dataset~\cite{henderson2017granite}.}, MIMIC-III is collected in ICU, where patients usually have a large amount of diagnoses and medications. As a result, only a small portion of the constructed interactions are truly clinically relevant. In general, the comparison shown in Fig.~\ref{fig:phenotype_relevance} suggests that modeling the inter-modal interactions is of crucial importance for discovering phenotypes that are clinically relevant, but the interactions should be established in a reasonable way, or be inferred jointly from the data, like our cHITF framework.

\begin{table}
\renewcommand{\arraystretch}{1.1}
  \centering
  \caption{Three examples of clinically relevant phenotypes inferred by cHITF}\label{tab:phenotype-examples}
\begin{tabular}{m{3mm}l}\toprule

\multicolumn{2}{l}{\textbf{Phenotype 1 (Trauma)}}\\\hline
\textcolor{Plum}{\textit{Dx}}& \begin{tabular}[c]{@{}l@{}}\textcolor{Plum}{Fracture of vertebral column without}\\~~~ \textcolor{Plum}{mention of spinal cord injury (0.139);}\\
\textcolor{Plum}{Fracture of rib(s) sternum larynx and trachea (0.128);}\\
\textcolor{Plum}{Other open wound of head (0.105);~~...} \end{tabular}\\\hline
\color{red}{\textit{Rx}}& \begin{tabular}[c]{@{}l@{}}\textcolor{red}{Morphine Sulfate (0.287); ~~Famotidine (0.141);}\\
\textcolor{red}{Acetaminophen (0.104);~~...} \end{tabular}\\\hline
\textcolor{blue}{\textit{Lab}}& \begin{tabular}[c]{@{}l@{}}\textcolor{blue}{MCHC [Blood] (0.998); ~~Specific Gravity [Urine] (0.001);}\\
\textcolor{blue}{Calculated Bicarbonate, Whole Blood [Blood] (0.001).}\\\end{tabular}\\\toprule

\multicolumn{2}{l}{\textbf{Phenotype 2 (Fluid, electrolyte and acid-base disorders)}}\\\hline
\textcolor{Plum}{\textit{Dx}}& \begin{tabular}[c]{@{}l@{}}\textcolor{Plum}{Disorders of fluid electrolyte and acid-base balance (1.000).} \end{tabular}\\\hline
\color{red}{\textit{Rx}}& \begin{tabular}[c]{@{}l@{}}\textcolor{red}{Potassium Chloride (0.284); ~~Magnesium Sulfate (0.078);}\\
\textcolor{red}{Calcium Gluconate (0.039);~~...} \end{tabular}\\\hline
\textcolor{blue}{\textit{Lab}}&\begin{tabular}[c]{@{}l@{}} \textcolor{blue}{Bicarbonate [Blood] (0.143); ~~Calcium, Total [Blood] (0.136);}\\
\textcolor{blue}{Phosphate [Blood] (0.108).}\\\end{tabular}\\\toprule

\multicolumn{2}{l}{\textbf{Phenotype 3 (Acute myocardial infarction)}}\\\hline
\textcolor{Plum}{\textit{Dx}}& \begin{tabular}[c]{@{}l@{}}\textcolor{Plum}{Acute myocardial infarction (1.000).} \end{tabular}\\\hline
\color{red}{\textit{Rx}}& \begin{tabular}[c]{@{}l@{}}\textcolor{red}{Captopril (0.088); ~~Clopidogrel Bisulfate (0.071);}\\
\textcolor{red}{Metoprolol Tartrate (0.067);~~...} \end{tabular}\\\hline
\textcolor{blue}{\textit{Lab}}&\begin{tabular}[c]{@{}l@{}} \textcolor{blue}{Creatine Kinase (CK) [Blood] (0.108);} \\
\textcolor{blue}{Creatine Kinase, MB Isoenzyme [Blood] (0.103);}\\
\textcolor{blue}{Troponin T [Blood] (0.103);~~...}\\\end{tabular}\\\toprule

\end{tabular}
\end{table}

To qualitatively illustrate the superiority of cHITF in terms of the clinical relevance of the phenotypes, we present three examples of the phenotypes inferred by cHITF in Table~\ref{tab:phenotype-examples}. The first phenotype in Table~\ref{tab:phenotype-examples} is annotated by the clinician to be highly related to trauma. The diagnoses are different subtypes of trauma; the medications are very relevant in that morphine and acetaminophen are typical pain killers that are important in trauma management, and famotidine is a protectant to prevent stress ulcers of the stomach. Moreover, the lab tests are also typical for trauma diagnoses as they quantify blood loss (anaemia), acidosis (inadequate organ perfusion due to blood loss) and dehydration (low blood pressure due to blood loss). The remaining two phenotypes are evidently related to fluid, electrolyte and acid-base disorders and acute myocardial infarction, respectively. More examples can be found in the supplemental materials.

\subsubsection{Sparsity and Diversity}
The sparsity and diversity are another two desired properties to indicate that the set of inferred phenotypes are distinct enough for characterizing different disease states. We measure the sparsity by the ratio of non-zero entries in the factor matrix, and use two metrics to quantify the diversity: the average cosine similarity and the Jaccard similarity at $K$. The average cosine similarity is defined as below~\cite{kim2017discriminative}:
\begin{equation}\label{eq:similarity}\small
     \text{Cosine Similarity} = \frac{\sum_{r_1=1}^{R}\sum_{r_2>r_1}^{R}\left\{ \sum_{n=1}^N \cos(\mathbf{U}^{(n)}_{:r_1}, \mathbf{U}^{(n)}_{:r_2}) \right\}}{N\times R\times(R-1)},
\end{equation}
where $R$ is the number of phenotypes, $N$ is the number of modalities, and $\mathbf{U}^{(n)}_{:r}$ is the $r^{th}$ column of the factor matrix corresponding to the $n^{th}$ modality. We define the Jaccard similarity at $K$ as:
\begin{equation}\small
    \text{Jaccard}@K=\frac{1}{R(R-1)} \sum_{r_1=1}^{R}\sum_{r_2>r_1}^{R}\frac{\lvert Q_{r_1}(K) \cap Q_{r_2}(K) \rvert}{\lvert Q_{r_1}(K) \cup Q_{r_2}(K) \rvert},
\end{equation}
where $Q_{r}(K)$ is the union of the top $K$ items of each modality of the $r^{th}$ phenotype, and $\lvert Q \rvert$ is the size of the set $Q$. The cosine similarity measures the overall distinctness of the phenotypes, whereas the Jaccard$@K$ measures the distinctness of the top $K$ items in each phenotype as they are often of utmost interest to clinicians. In this paper, we set $K$ to be ten. Both smaller cosine similarity and Jaccard$@K$ indicate more diverse phenotypes.

\begin{table}[]
\renewcommand{\arraystretch}{1.2}
\centering
\caption{Sparsity and diversity of phenotypes inferred from MIMIC-III dataset
}
\label{tab:diversity}
\begin{tabular}{ccccc}
\toprule
 & Reg. & Sparsity & \begin{tabular}[c]{@{}c@{}}Cosine\\ Similarity\end{tabular} & Jaccard@10  \\ \hline
CP-APR  & --  & 0.67 (0.001)   & 0.64 (0.001)   &   0.26 (0.008) \\
Marble & --    & 0.51 (0.001)   & 0.50 (0.001)   &   0.18 (0.007) \\
Rubik  & --    & 0.95 (0.003)   & 0.87 (0.005)   &   0.26 (0.006) \\
Granite & --   & 0.90 (0.003)   & 0.78 (0.018)   &   0.12 (0.062) \\\hline
SiCNMF & --    & \textbf{0.08 (0.026)}   & 0.33 (0.069)   &   0.12 (0.017) \\
cHITF & Both & 0.17 (0.002) & \textbf{0.11 (0.005)}   &   \textbf{0.04 (0.002)}  \\\hline
cHITF & Neither & 0.24 (0.002) & 0.19 (0.008) & 0.06 (0.002)\\
cHITF & Elastic net & 0.19 (0.001) & 0.17 (0.004) & 0.06 (0.002)\\
cHITF & Angular & 0.21 (0.002) &  0.11 (0.002) & 0.04 (0.001)\\
\bottomrule
\end{tabular}
\begin{flushleft}\textit{\textbf{Reg.}: abbreviation of ``Regularization'' indicating the active regularization(s).}\end{flushleft}
\end{table}

We summarize the results obtained using the three metrices in Table~\ref{tab:diversity}. cHITF can generate most diverse phenotypes, with cosine similarity of 0.11 and Jaccard@10 of as small as 0.04. CP-APR and Rubik generate the least diverse phenotypes as they rely on the ``equal-correspondence'' assumption that does not reflect the true inter-modal interactions; thus, the input tensor is dominated by the false and redundant interactions, leading to non-distinct phenotypes. Marble has a more effective sparsity constraint that enforces the majority of the entries to be zero, which indirectly contributes to the improved diversity. Granite, although with its input dominated by false interactions, achieved impressive diversity performance with Jaccard@10 of 0.12. This attributes to the angular regularization that Granite adopts. However, we observe a larger cosine similarity for Granite, and we speculate that this is because the sparsity constraint adopted by Granite is less effective. SiCNMF achieves the second-best diversity as its input does not consider the inter-modal interactions, and thus not dominated by the false interactions. cHITF considers the inter-modal interactions and learn them from the data to alleviate the false-interaction issue in its latent interaction tensor. In addition, cHITF also adopts the angular constraint that further improves the diversity.

Regarding the impact of the two regularization terms incorporated, we conduct an ablation study and summarize the results in the last three rows in Table~\ref{tab:diversity}. Compared to the case without any regularization, the elastic net regularization improves the sparsity and the cosine similarity. And the angular regularization improves the diversity with 33.3\% relative improvement of the Jaccard@10 score. With both regularization terms incorporated, cHITF finally achieves the best performance in terms of diversity.

\begin{table}[]
\renewcommand{\arraystretch}{1.2}
\centering
\caption{The AUPRC score for predicting in-hospital mortality of MIMIC-III}
\label{tab:predictions}
\begin{tabular}{cccc}
\toprule
   & {Dx \& Rx} & {Dx \& Lab} & {Dx \& Rx \& Lab}\\ \midrule
CP-APR  & 0.34 (0.030)   &   0.38 (0.024)   &   0.36 (0.020)   \\
Marble  & 0.33 (0.024)   &   0.36 (0.020)   &   0.34 (0.021)   \\
Rubik   & 0.32 (0.030)   &   0.30 (0.021)   &   0.34 (0.024)   \\
Granite & 0.30 (0.016)   &   0.33 (0.017)   &   0.30 (0.019)   \\
SiCNMF  & 0.31 (0.036)   &   0.24 (0.021)   &   0.38 (0.076)   \\\midrule
HITF    & \textbf{0.46 (0.013)}   &   \textbf{0.41 (0.011)}   &   0.39 (0.032)   \\
cHITF   & --               &   --               &   \textbf{0.47 (0.012)}   \\\bottomrule
\end{tabular}
\end{table}

\begin{table}[]
\renewcommand{\arraystretch}{1.2}
\centering
\caption{The AUPRC score for predicting in-hospital mortality of eICU}
\label{tab:predictions_eicu}
\begin{tabular}{cccc}
\toprule
   & {Dx \& Rx} & {Dx \& Treatment} & {Dx \& Rx \& Treatment}\\ \midrule
CP-APR & 0.31 (0.027) & 0.29 (0.036) & 0.32 (0.028) \\
Marble & 0.31 (0.021) & 0.33 (0.062) & 0.34 (0.019) \\
Rubik & 0.43 (0.020) & 0.35 (0.040) & 0.35 (0.136) \\
Granite & 0.32 (0.043) & 0.26 (0.045) & 0.23 (0.028) \\
SiCNMF & 0.38 (0.037) & 0.28 (0.045) & 0.37 (0.045) \\\midrule
HITF    & \textbf{0.48 (0.016)}   &   \textbf{0.39 (0.051)}   &   0.38 (0.025)  \\
cHITF   & --               &   --               &   \textbf{0.51 (0.022)}   \\\bottomrule
\end{tabular}
\end{table}

\subsection{cHITF Infers Phenotypes with Improved Predictive Power}

We evaluate the performance of using the phenotypes inferred by cHITF as features for the subsequent in-hospital mortality prediction task. We first divide the dataset for training and testing with a proportion of 8:2 and adopt five-fold cross validation. Then we apply cHITF and all of the baselines with different modalities to learn the phenotypes and the patient representations of the training subset, after which we project the test subset onto the learned phenotypes to obtain the patient representations of the test subset. We use a lasso-regularized logistic regression to perform the binary classification with the patient representations as the features.
We use AUPRC (Area Under the Precision-Recall Curve) as the evaluation metric calculated.

Tables~\ref{tab:predictions} and \ref{tab:predictions_eicu} summarize the prediction performance of cHITF and the baselines with different modalities on MIMIC-III and eICU, respectively. Each column corresponds to a particular combination of modalities. ``Dx'' denotes diagnosis, ``Rx'' denotes medication, and ``Lab'' denotes lab test.
All baseline models give similar prediction performance, whereas HITF outperforms all the baselines by a large margin. One phenomenon we observe is that most of the baselines based on ``equal-correspondence'' assumption obtain better prediction performance with lab tests than medications; however, the prediction performance of SiCNMF and HITF using lab tests are worse than that using medications. The reason is that unlike medications, many abnormal lab tests do not correspond to specific diagnoses. So the issue of dominating false interaction is less severe than the case of medications, and inferring the relationship between diagnoses and lab tests is much harder. Nevertheless, HITF still outperforms all the baselines. Given only two modalities, one hidden interaction tensor is sufficient, and in this case the cHITF reduces to the HITF model. The third column shows the results obtained based on all three modalities. All baselines except SiCNMF have similar performance to their counterparts obtained based on only two modalities. The AUPRC of SiCNMF with all three modalities was boosted to 0.38, achieving almost 23\% relative improvement compared to that with only Dx and Rx. Yet its prediction performance is not obviously better than other baselines, even though the phenotypes inferred by SiCNMF are impressively sparse as shown in Table~\ref{tab:diversity}. This is mainly because it is based on collective matrix factorization and does not consider the inter-modal interactions at all.  We tried HITF with one forth-order hidden interaction tensor to account for all the modalities, leading to a surprising drop of around 15\%. This suggests that inferring the diagnosis-medication-lab-test interactions is much more challenging and can easily compromise the representational power of the learned factors. With our cHITF framework, where the diagnosis-medication interactions and the diagnosis-lab-test interactions are separately modeled with two latent interaction tensors, the prediction performance further improved by nearly 24\% compared with the best-performing baseline.

\subsection{cHITF Models Modalities with Different Distributions}\label{sec:rq4}

Finally, we incorporate the modality of input fluids to cHITF in MIMIC-III. The input fluids are the total amount of fluids input to the patients, and thus we model it using the Gaussian distribution. As the baselines are not capable of simultaneously considering modalities with different distributions, we focus on two aspects in this section: first, whether incorporating the input fluids helps improve representational power; second, whether cHITF infer interpretable and clinically relevant diagnosis-fluid correspondence.

Table~\ref{tab:chitf-all-modalities} shows the AUPRC for the mortality prediction and the two diversity measures based on different combinations of modalities. Incorporating the input fluids further improves the prediction performance by 4-9\% for different combinations of modalities, demonstrating that the use of fluids enhances the representational power of the learned factors and that the cHITF framework is capable of modelling modalities with different distributions. On the other hand, integrating the input fluids compromises the diversity of the inferred phenotypes. This is related to the dataset we used. In fact, many fluids being input into patients are rather generic. For instance, lactated ringers is commonly used in ICU for fluid resuscitation after a blood loss, which can be caused by many different diagnoses, \textit{e.g.}, trauma, burn injury and surgery.
Nonetheless, cHITF is still capable of identifying relevant items, for example, insulin for diabetes mellitus and midazolam for sleep disorders.

\begin{table}
\renewcommand{\arraystretch}{1.2}
  \centering
  \caption{Performance with different combinations of modalities in MIMIC-III}\label{tab:chitf-all-modalities}
\begin{tabular}{cccc}\toprule
Modalities  & AUPRC & \begin{tabular}[c]{@{}c@{}}Cosine\\ Similarity\end{tabular} & Jaccard@10 \\\midrule
Dx \& Fluid & 0.44 (0.029) & 0.31 (0.010) & 0.04 (0.004) \\\hline
Dx \& Rx & 0.46 (0.013) & \textbf{0.05 (0.002)} & \textbf{0.02 (0.001)}\\
Dx \& Rx \& Fluid & \textbf{0.48 (0.021)} & 0.35 (0.057) & 0.05 (0.011) \\\hline
Dx \& Lab & 0.41 (0.011) & \textbf{0.08 (0.009)} & \textbf{0.03 (0.002)}\\
Dx \& Lab \& Fluid & \textbf{0.47 (0.019)} & 0.32 (0.008) & 0.05 (0.001) \\\hline
Dx \& Rx \& Lab  & 0.47 (0.012) & \textbf{0.11 (0.005)} & \textbf{0.04 (0.002)}  \\
Dx \& Rx \& Lab \& Fluid  & \textbf{0.51 (0.059)} & 0.21 (0.064) & 0.06 (0.017)  \\\bottomrule
\end{tabular}
\end{table}

\begin{table}
\renewcommand{\arraystretch}{1.1}
  \centering
  \caption{Examples of diagnosis-fluid correspondence inferred by cHITF}\label{tab:dx-input-corrs}
\begin{tabular}{l}\toprule

\textbf{Heart Failure}\\\hline
PO Intake (0.19); ~~Insulin (0.17); ~~Propofol (0.10); \\
Lactated Ringers (0.09); ~~Levophed-k (0.07); ~~...\\\toprule

\textbf{Diabetes Mellitus}\\\hline
Insulin (0.31); ~~Gastric Meds (0.22); ~~D5W (0.07); \\
0.9\% Normal Saline (0.04); ~~GT Flush (0.03); ~~...\\\toprule

\textbf{Organic sleep disorders}\\\hline
Midazolam (0.25); ~~Fentanyl (0.15); ~~Amiodarone (0.10); \\
Lasix (0.08); ~~D5W (0.06); ~~...\\\toprule

\textbf{Peritonitis and retroperitoneal infections}\\\hline
Lactated Ringers (0.32); ~~Albumin 25\% (0.18); ~~D5W (0.06); \\
Gastric Meds (0.05); ~~Po Intake (0.05); ~~...\\\toprule

\end{tabular}
\end{table}

\subsection{cHITF is Efficient and Scalable}
\begin{figure}
    \centering
    \includegraphics[width=0.88\columnwidth]{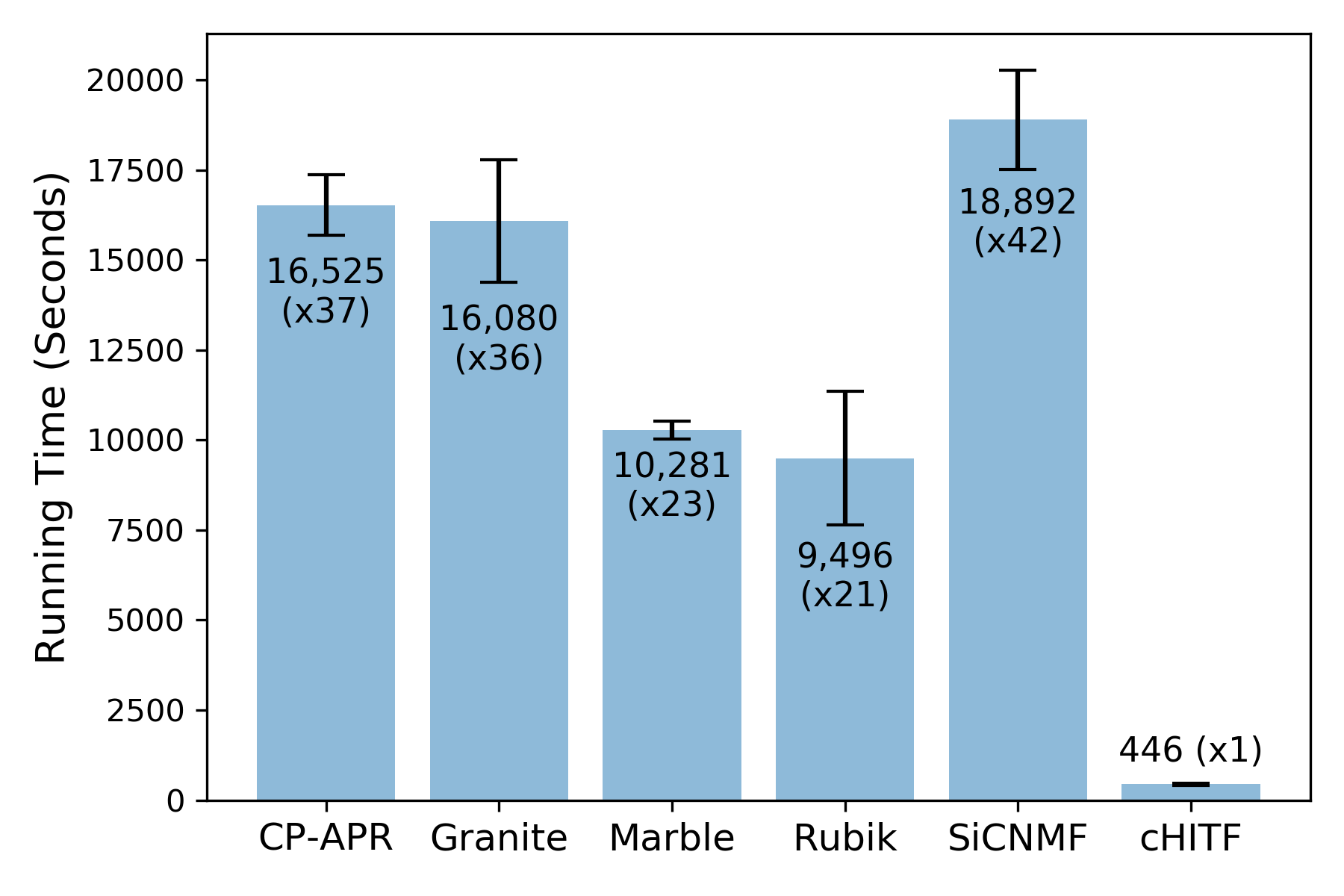}
    \caption{Running time of cHITF and baselines on the MIMIC-III subset with modalities of diagnosis, medication and lab test.}
    \label{fig:running_time}
\end{figure}

We also compare the running time against the baselines. All baselines are run on high-end servers equipped with dual 64-core AMD EPYC 7742 CPUs and 1TB memory. Our model is run on GPU servers equipped with a 10-core Intel E5-2630v4 CPU, 256 GB memory and dual NVIDIA Tesla P40 GPUs. It is worth noting that none of the baseline models can run on GPUs as they rely on heavy tensor operations that are not available in GPU computing environments. Fig.~\ref{fig:running_time} shows the running time of cHITF and all baselines on the MIMIC-III subset. We run all models until they reach their convergence criteria. cHITF only takes 446 seconds, while the fastest baseline, Rubik, takes 9,496 secons, which is 21 times slower than cHITF.

\subsection{Summary}
Through extensive and comprehensive experiments with real-world EHR dataset, we demonstrated the superiority of HITF in terms of inferring the inter-modal interactions and cHITF in terms of modelling multiple modalities simultaneously. The reason behind the significant performance improvement of both clinical relevance is that HITF does not rely on the ``equal-correspondence'' assumption. Instead, it infers the inter-modal correspondence from the data so that false interactions can be effectively avoided. The phenotypes inferred by cHITF are more clinically relevant, more diverse and of better prediction performance. To summarize, considering the inter-modal interactions is critically important to discover phenotypes of better quality and cHITF provides a better way than the existing models to handle the unobserved inter-modal correspondence.

\section{Conclusion}\label{sec:conclusion}
In this paper, we introduce a novel tensor factorization method, called HITF, to infer the hidden interactions between modalities and learn the phenotypes jointly, given only the marginalized observations. We present its formulation with both Poisson distribution for integer data and Gaussian distribution for real-valued data. Furthermore, using HITF as building blocks, we propose the cHITF framework that allows multiple hidden interaction tensors to account for the interactions between different modalities. We evaluate the proposed model and framework using a real-world EHR dataset. The results demonstrate that the inter-modal correspondence inferred by HITF are significantly more meaningful and relevant than the ``equal-corrrespondence'' assumption adopted by the existing computational phenotyping models based on tensor factorization. Consequently, the phenotypes learned by cHITF are more clinically relevant and diverse. Moreover, the remarkable improvement of the predictive performance also validates the effectiveness of representing patients using the phenotypes learned by cHITF.

\vspace{-3mm}\section*{Acknowledgments}
This research is partially supported by General Research Fund RGC/HKBU12201219 and RGC/HKBU12202117 from the Research Grants Council of Hong Kong.

\bibliographystyle{IEEEtran}
{\footnotesize
\bibliography{IEEEabrv,main_refs}

\begin{thebibliography}{10}
\providecommand{\url}[1]{#1}
\csname url@samestyle\endcsname
\providecommand{\newblock}{\relax}
\providecommand{\bibinfo}[2]{#2}
\providecommand{\BIBentrySTDinterwordspacing}{\spaceskip=0pt\relax}
\providecommand{\BIBentryALTinterwordstretchfactor}{4}
\providecommand{\BIBentryALTinterwordspacing}{\spaceskip=\fontdimen2\font plus
\BIBentryALTinterwordstretchfactor\fontdimen3\font minus
  \fontdimen4\font\relax}
\providecommand{\BIBforeignlanguage}[2]{{%
\expandafter\ifx\csname l@#1\endcsname\relax
\typeout{** WARNING: IEEEtran.bst: No hyphenation pattern has been}%
\typeout{** loaded for the language `#1'. Using the pattern for}%
\typeout{** the default language instead.}%
\else
\language=\csname l@#1\endcsname
\fi
#2}}
\providecommand{\BIBdecl}{\relax}
\BIBdecl

\bibitem{jensen2012mining}
P.~B. Jensen, L.~J. Jensen, and S.~Brunak, ``Mining electronic health records:
  Towards better research applications and clinical care,'' \emph{Nature
  Reviews Genetics}, vol.~13, no.~6, pp. 395--405, 2012.

\bibitem{wang2016diagnosis}
S.~Wang, X.~Chang, X.~Li, G.~Long, L.~Yao, and Q.~Z. Sheng, ``Diagnosis code
  assignment using sparsity-based disease correlation embedding,'' \emph{IEEE
  Transactions on Knowledge and Data Engineering}, vol.~28, no.~12, pp.
  3191--3202, 2016.

\bibitem{luo2016tensor}
Y.~Luo, F.~Wang, and P.~Szolovits, ``Tensor factorization toward precision
  medicine,'' \emph{Briefings in Bioinformatics}, vol.~18, no.~3, pp. 511--514,
  2016.

\bibitem{xu2017patient}
H.~Xu, W.~Wu, S.~Nemati, and H.~Zha, ``Patient flow prediction via
  discriminative learning of mutually-correcting processes,'' \emph{IEEE
  transactions on Knowledge and Data Engineering}, vol.~29, no.~1, pp.
  157--171, 2017.

\bibitem{liu2019complication}
B.~Liu, Y.~Li, S.~Ghosh, Z.~Sun, K.~Ng, and J.~Hu, ``Complication risk
  profiling in diabetes care: A bayesian multi-task and feature relationship
  learning approach,'' \emph{IEEE Transactions on Knowledge and Data
  Engineering}, 2019.

\bibitem{hripcsak2013next}
G.~Hripcsak and D.~J. Albers, ``Next-generation phenotyping of electronic
  health records,'' \emph{Journal of the American Medical Informatics
  Association}, vol.~20, no.~1, pp. 117--121, 2013.

\bibitem{ho2014limestone}
J.~C. Ho, J.~Ghosh, S.~R. Steinhubl, W.~F. Stewart, J.~C. Denny, B.~A. Malin,
  and J.~Sun, ``Limestone: High-throughput candidate phenotype generation via
  tensor factorization,'' \emph{Journal of Biomedical Informatics}, vol.~52,
  pp. 199--211, 2014.

\bibitem{lasko2013computational}
T.~A. Lasko, J.~C. Denny, and M.~A. Levy, ``Computational phenotype discovery
  using unsupervised feature learning over noisy, sparse, and irregular
  clinical data,'' \emph{PLOS One}, vol.~8, no.~6, p. e66341, 2013.

\bibitem{dl:inHealthInfomatics:review}
D.~Rav{\`\i}, C.~Wong, F.~Deligianni, M.~Berthelot, J.~Andreu-Perez, B.~Lo, and
  G.-Z. Yang, ``Deep learning for health informatics,'' \emph{IEEE Journal of
  Biomedical and Health Informatics}, vol.~21, no.~1, pp. 4--21, 2017.

\bibitem{wang2015rubik}
Y.~Wang, R.~Chen, J.~Ghosh, J.~C. Denny, A.~Kho, Y.~Chen, B.~A. Malin, and
  J.~Sun, ``Rubik: Knowledge guided tensor factorization and completion for
  health data analytics,'' in \emph{Proceedings of the 21st ACM SIGKDD
  International Conference on Knowledge Discovery and Data Mining}.\hskip 1em
  plus 0.5em minus 0.4em\relax ACM, 2015, pp. 1265--1274.

\bibitem{kim2017discriminative}
Y.~Kim, R.~El-Kareh, J.~Sun, H.~Yu, and X.~Jiang, ``Discriminative and distinct
  phenotyping by constrained tensor factorization.'' \emph{Scientific Reports},
  vol.~7, no.~1, p. 1114, 2017.

\bibitem{lee1999learning}
D.~D. Lee and H.~S. Seung, ``Learning the parts of objects by non-negative
  matrix factorization,'' \emph{Nature}, vol. 401, no. 6755, pp. 788--791,
  1999.

\bibitem{hu2015scalable}
C.~Hu, P.~Rai, C.~Chen, M.~Harding, and L.~Carin, ``Scalable bayesian
  non-negative tensor factorization for massive count data,'' in \emph{Joint
  European Conference on Machine Learning and Knowledge Discovery in
  Databases}.\hskip 1em plus 0.5em minus 0.4em\relax Springer, 2015, pp.
  53--70.

\bibitem{johnson2016machine}
A.~E. Johnson, M.~M. Ghassemi, S.~Nemati, K.~E. Niehaus, D.~A. Clifton, and
  G.~D. Clifford, ``Machine learning and decision support in critical care,''
  \emph{Proceedings of the IEEE}, vol. 104, no.~2, p. 444, 2016.

\bibitem{ho2014marble}
J.~C. Ho, J.~Ghosh, and J.~Sun, ``Marble: High-throughput phenotyping from
  electronic health records via sparse nonnegative tensor factorization,'' in
  \emph{Proceedings of the 20th ACM SIGKDD International Conference on
  Knowledge Discovery and Data Mining}.\hskip 1em plus 0.5em minus 0.4em\relax
  ACM, 2014, pp. 115--124.

\bibitem{yang2017tagited}
K.~Yang, X.~Li, H.~Liu, J.~Mei, G.~T. Xie, J.~Zhao, B.~Xie, and F.~Wang,
  ``{TaGiTeD: Predictive task guided tensor decomposition for representation
  learning from electronic health records},'' in \emph{Thirty-First AAAI
  Conference on Artificial Intelligence}.\hskip 1em plus 0.5em minus
  0.4em\relax AAAI, 2017.

\bibitem{yin2019learning}
K.~Yin, D.~Qian, W.~K. Cheung, B.~C.~M. Fung, and J.~Poon, ``Learning
  phenotypes and dynamic patient representations via {RNN} regularized
  collective non-negative tensor factorization,'' in \emph{Proceedings of the
  Thirty-Third AAAI Conference on Artificial Intelligence}.\hskip 1em plus
  0.5em minus 0.4em\relax AAAI Press, 2019, pp. 1246--1253.

\bibitem{perros2017spartan}
I.~Perros, E.~E. Papalexakis, F.~Wang, R.~Vuduc, E.~Searles, M.~Thompson, and
  J.~Sun, ``{SPARTan}: Scalable {PARAFAC2} for large \& sparse data,'' in
  \emph{Proceedings of the 23rd ACM SIGKDD International Conference on
  Knowledge Discovery and Data Mining}.\hskip 1em plus 0.5em minus 0.4em\relax
  ACM, 2017, pp. 375--384.

\bibitem{arxiv2016cmf}
S.~Gunasekar, J.~C. Ho, J.~Ghosh, S.~Kreml, A.~N. Kho, J.~C. Denny, B.~A.
  Malin, and J.~Sun, ``{Phenotyping using structured collective matrix
  factorization of multi--source EHR data},'' \emph{ArXiv e-prints}, Sep. 2016.

\bibitem{gunasekar2016mining}
S.~Gunasekar, ``Mining structured matrices in high dimensions,'' Ph.D.
  dissertation, The University of Texas at Austin, 2016.

\bibitem{zhu2017expectile}
R.~Zhu, D.~Niu, L.~Kong, and Z.~Li, ``Expectile matrix factorization for skewed
  data analysis,'' in \emph{Proceedings of the Thirty-First AAAI Conference on
  Artificial Intelligence}, 2017, pp. 259--265.

\bibitem{tu2019m}
W.~Tu, P.~Liu, J.~Zhao, Y.~Liu, L.~Kong, G.~Li, B.~Jiang, G.~Tian, and H.~Yao,
  ``M-estimation in low-rank matrix factorization: a general framework,'' in
  \emph{2019 IEEE International Conference on Data Mining (ICDM)}.\hskip 1em
  plus 0.5em minus 0.4em\relax IEEE, 2019, pp. 568--577.

\bibitem{choi2017gram}
E.~Choi, M.~T. Bahadori, L.~Song, W.~F. Stewart, and J.~Sun, ``Gram:
  Graph-based attention model for healthcare representation learning,'' in
  \emph{Proceedings of the 23rd ACM SIGKDD International Conference on
  Knowledge Discovery and Data Mining}, 2017, pp. 787--795.

\bibitem{song2019medical}
L.~Song, C.~W. Cheong, K.~Yin, W.~K. Cheung, B.~C. Fung, and J.~Poon, ``Medical
  concept embedding with multiple ontological representations.'' in
  \emph{Proceedings of the Twenty-Eighth International Joint Conference on
  Artificial Intelligence ({IJCAI-19})}, 2019, pp. 4613--4619.

\bibitem{papalexakis2017tensors}
E.~E. Papalexakis, C.~Faloutsos, and N.~D. Sidiropoulos, ``Tensors for data
  mining and data fusion: Models, applications, and scalable algorithms,''
  \emph{ACM Transactions on Intelligent Systems and Technology (TIST)}, vol.~8,
  no.~2, p.~16, 2017.

\bibitem{hitchcock1927expression}
F.~L. Hitchcock, ``The expression of a tensor or a polyadic as a sum of
  products,'' \emph{Journal of Mathematics and Physics}, vol.~6, no. 1-4, pp.
  164--189, 1927.

\bibitem{chi2012tensors}
E.~C. Chi and T.~G. Kolda, ``On tensors, sparsity, and nonnegative
  factorizations,'' \emph{SIAM Journal on Matrix Analysis and Applications},
  vol.~33, no.~4, pp. 1272--1299, 2012.

\bibitem{kolda2009tensor}
T.~G. Kolda and B.~W. Bader, ``Tensor decompositions and applications,''
  \emph{SIAM Review}, vol.~51, no.~3, pp. 455--500, 2009.

\bibitem{shashua2005non}
A.~Shashua and T.~Hazan, ``Non-negative tensor factorization with applications
  to statistics and computer vision,'' in \emph{Proceedings of the 22nd
  International Conference on Machine Learning}.\hskip 1em plus 0.5em minus
  0.4em\relax ACM, 2005, pp. 792--799.

\bibitem{yin2018joint}
K.~Yin, W.~K. Cheung, Y.~Liu, B.~C.~M. Fung, and J.~Poon, ``Joint learning of
  phenotypes and diagnosis-medication correspondence via hidden interaction
  tensor factorization,'' in \emph{Proceedings of the Twenty-Seventh
  International Joint Conference on Artificial Intelligence
  ({IJCAI-18})}.\hskip 1em plus 0.5em minus 0.4em\relax AAAI Press, 2018, pp.
  3627--3633.

\bibitem{hsieh2015pu}
C.-J. Hsieh, N.~Natarajan, and I.~S. Dhillon, ``{PU} learning for matrix
  completion,'' in \emph{Proceedings of the 32nd International Conference on
  International Conference on Machine Learning}.\hskip 1em plus 0.5em minus
  0.4em\relax JMLR. org, 2015, pp. 2445--2453.

\bibitem{xiong2010temporal}
L.~Xiong, X.~Chen, T.-K. Huang, J.~Schneider, and J.~G. Carbonell, ``Temporal
  collaborative filtering with bayesian probabilistic tensor factorization,''
  in \emph{Proceedings of the 2010 SIAM International Conference on Data
  Mining}.\hskip 1em plus 0.5em minus 0.4em\relax SIAM, 2010, pp. 211--222.

\bibitem{zou2005regularization}
H.~Zou and T.~Hastie, ``Regularization and variable selection via the elastic
  net,'' \emph{Journal of the Royal Statistical Society: Series B (Statistical
  Methodology)}, vol.~67, no.~2, pp. 301--320, 2005.

\bibitem{xie2017learning}
P.~Xie, Y.~Deng, Y.~Zhou, A.~Kumar, Y.~Yu, J.~Zou, and E.~P. Xing, ``Learning
  latent space models with angular constraints,'' in \emph{Proceedings of the
  34th International Conference on Machine Learning}, vol.~70.\hskip 1em plus
  0.5em minus 0.4em\relax JMLR. org, 2017, pp. 3799--3810.

\bibitem{henderson2017granite}
J.~Henderson, J.~C. Ho, A.~N. Kho, J.~C. Denny, B.~A. Malin, J.~Sun, and
  J.~Ghosh, ``Granite: Diversified, sparse tensor factorization for electronic
  health record-based phenotyping,'' in \emph{2017 IEEE International
  Conference on Healthcare Informatics (ICHI)}.\hskip 1em plus 0.5em minus
  0.4em\relax IEEE, 2017, pp. 214--223.

\bibitem{xu2013block}
Y.~Xu and W.~Yin, ``A block coordinate descent method for regularized
  multiconvex optimization with applications to nonnegative tensor
  factorization and completion,'' \emph{SIAM Journal on Imaging Sciences},
  vol.~6, no.~3, pp. 1758--1789, 2013.

\bibitem{beck2013convergence}
A.~Beck and L.~Tetruashvili, ``On the convergence of block coordinate descent
  type methods,'' \emph{SIAM journal on Optimization}, vol.~23, no.~4, pp.
  2037--2060, 2013.

\bibitem{mimiciii}
A.~E. Johnson, T.~J. Pollard, L.~Shen, H.~L. Li-wei, M.~Feng, M.~Ghassemi,
  B.~Moody, P.~Szolovits, L.~A. Celi, and R.~G. Mark, ``{MIMIC-III, a freely
  accessible critical care database},'' \emph{Scientific Data}, vol.~3, 2016.

\bibitem{pollard2018eicu}
T.~J. Pollard, A.~E. Johnson, J.~D. Raffa, L.~A. Celi, R.~G. Mark, and
  O.~Badawi, ``The {eICU} collaborative research database, a freely available
  multi-center database for critical care research,'' \emph{Scientific Data},
  vol.~5, 2018.

\bibitem{hong2020generalized}
D.~Hong, T.~G. Kolda, and J.~A. Duersch, ``Generalized canonical polyadic
  tensor decomposition,'' \emph{SIAM Review}, vol.~62, no.~1, pp. 133--163,
  2020.

\end{thebibliography}
}

\vspace{-2mm}
\begin{IEEEbiography}[{\includegraphics[width=1in,height=1.25in,clip,keepaspectratio]{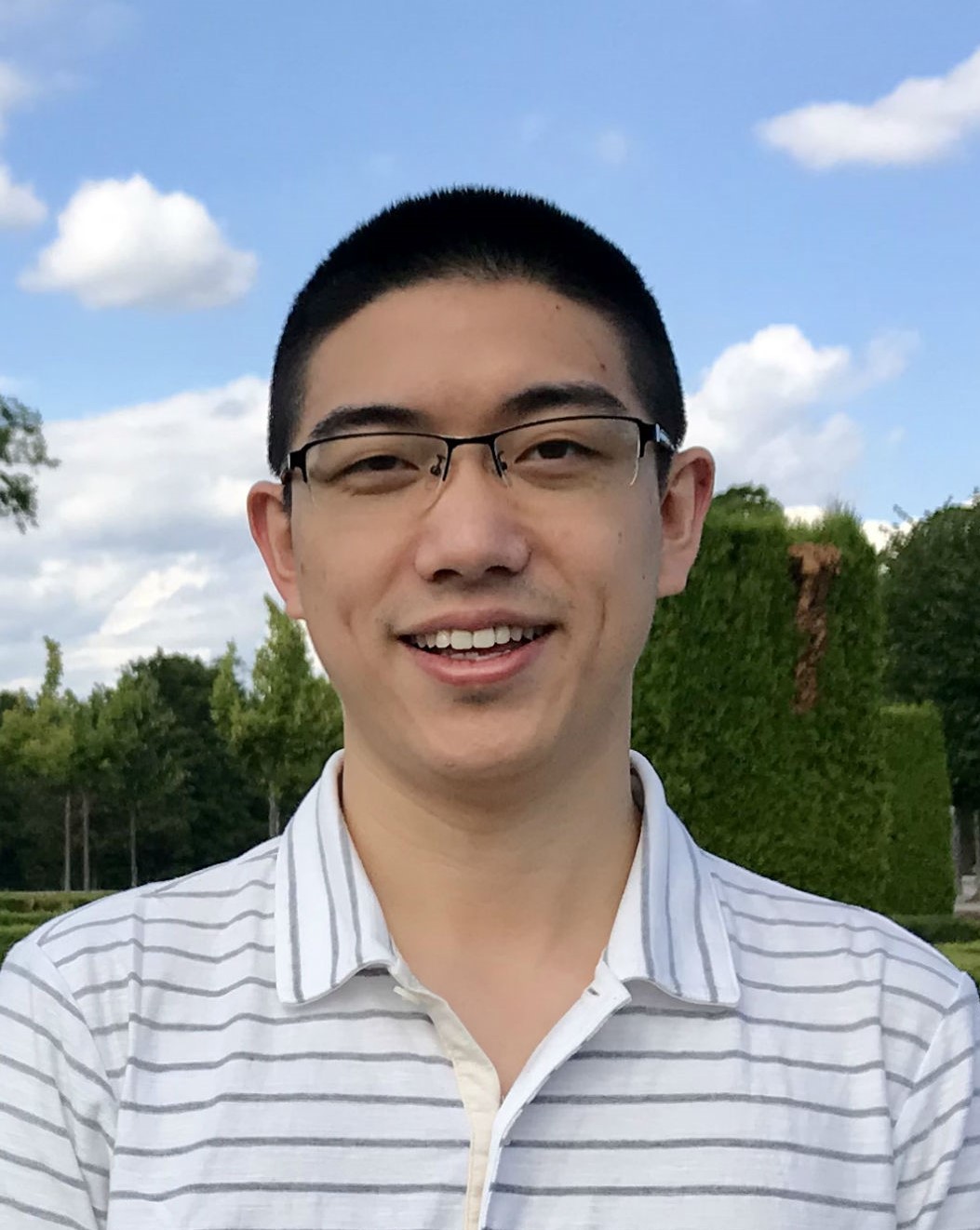}}]{Kejing Yin} received the B.Eng. degree from South China University of Technology, Guangzhou, China, in 2015. He is currently pursuing the Ph.D. degree with the Department of Computer Science, Hong Kong Baptist University. His research interests focus on computational phenotyping, tensor factorization, machine learning, data mining, and their applications to healthcare.
\end{IEEEbiography}

\begin{IEEEbiography}[{\includegraphics[width=1in,height=1.25in,clip,keepaspectratio]{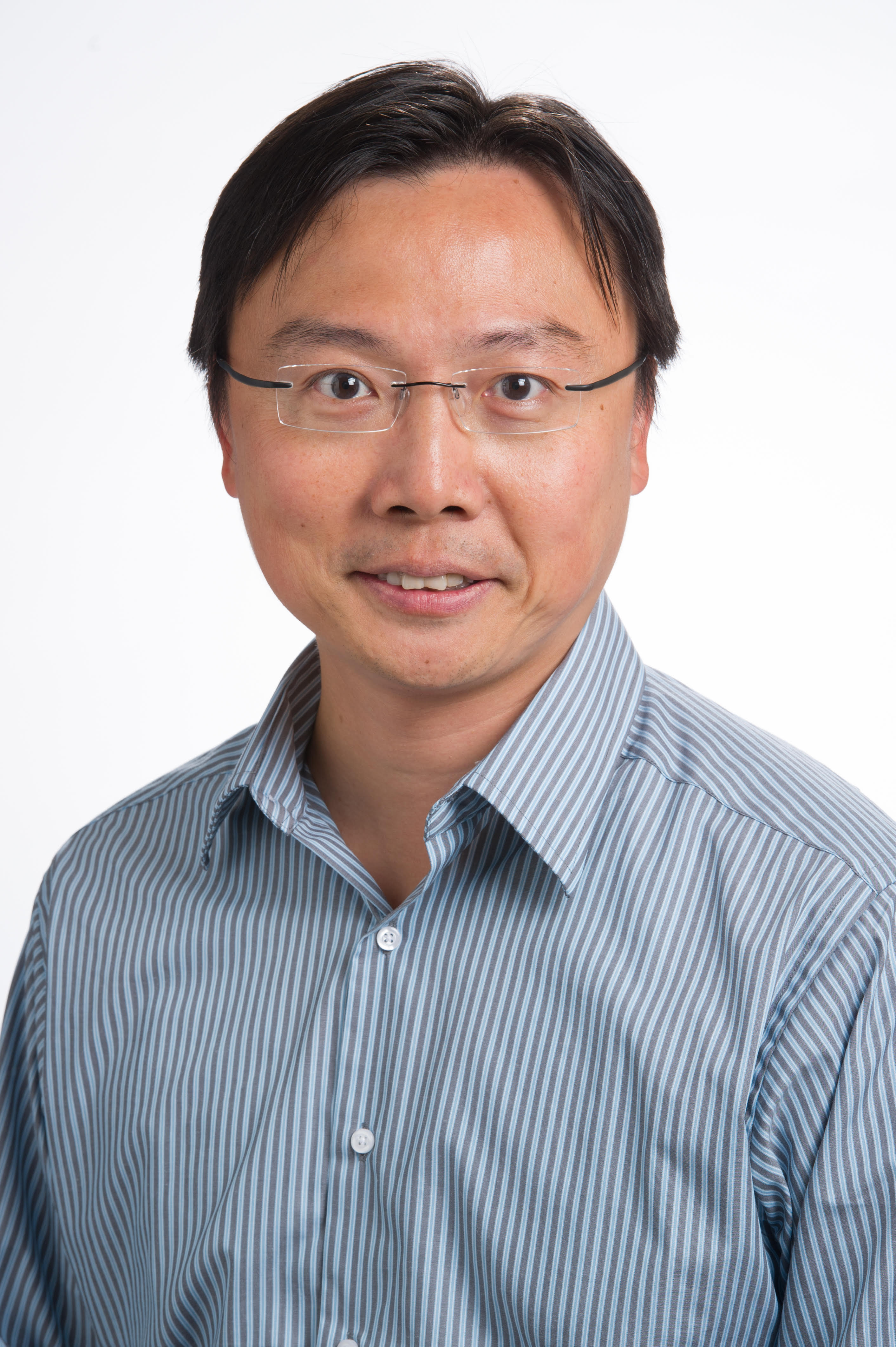}}]{William K. Cheung} received the Ph.D. degree in computer science from the Hong Kong University of Science and Technology in Hong Kong in 1999. He is currently the Head and Associate Professor of the Department of Computer Science, Hong Kong Baptist University, Hong Kong. His current research interests include artificial intelligence, data mining, collaborative information filtering, social network analysis, and healthcare informatics. He has served as the Co-Chairs and Program Committee Members for a number of international conferences and workshops, as well as Guest Editors of journals on areas including artificial intelligence, Web intelligence, data mining, Web services, e-commerce technologies, and health informatics. From 2002-2018, he was on the Editorial Board of the IEEE Intelligent Informatics Bulletin.
\end{IEEEbiography}

\begin{IEEEbiography}[{\includegraphics[width=1in,height=1.25in,clip,keepaspectratio]{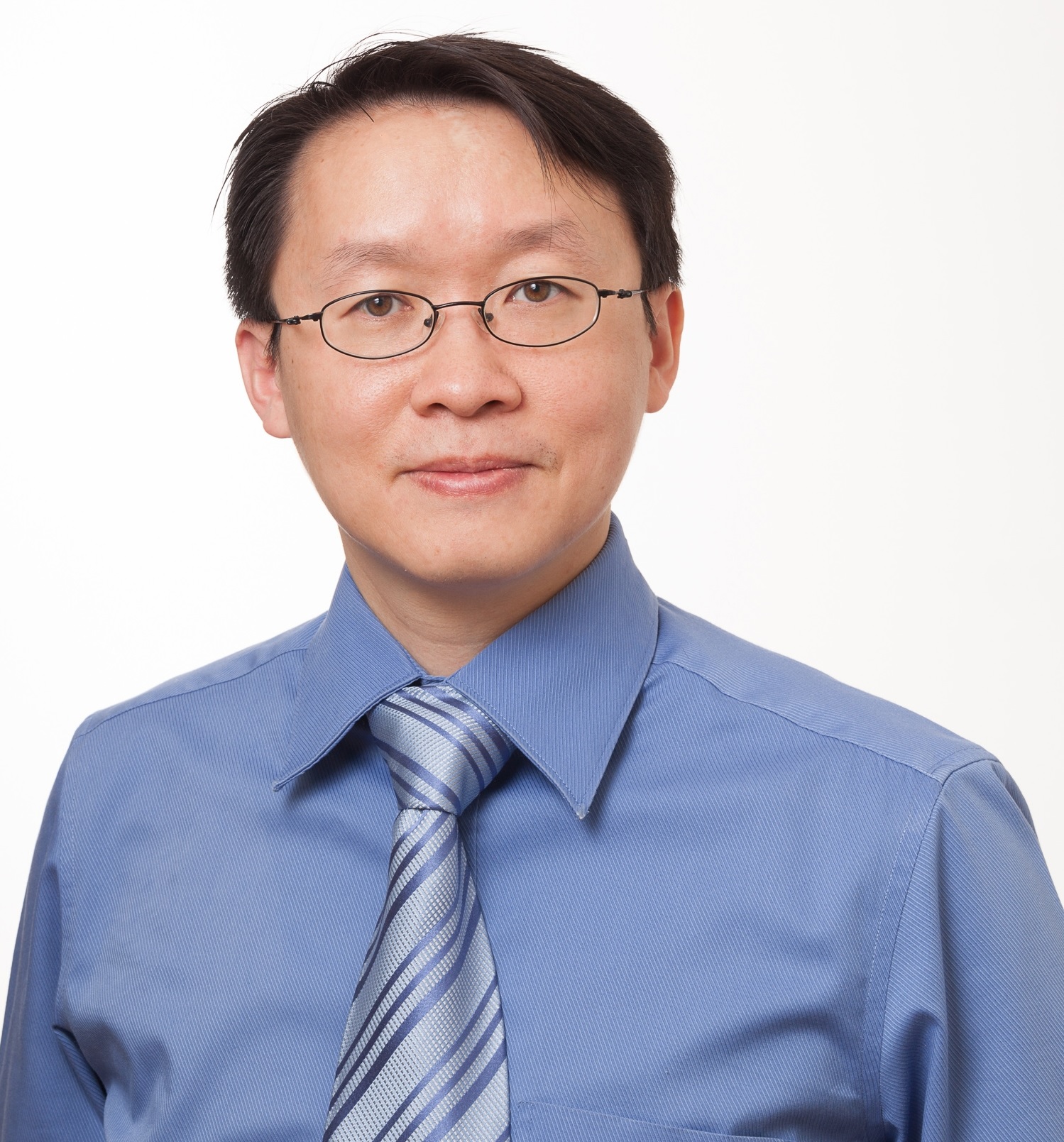}}]{Benjamin C. M. Fung} (M'09-SM'13) received the Ph.D. degree in Computing Science from Simon Fraser University in Canada in 2007. He is a Canada Research Chair in Data Mining for Cybersecurity, a Professor with the School of Information Studies at McGill University in Canada, and a Co-Curator of cybersecurity in the World Economic Forum. He also serves as Associate Editors for IEEE Transactions of Knowledge and Data Engineering (TKDE) and Elsevier Sustainable Cities and Society (SCS). He has over 120 refereed publications that span the research forums of data mining, privacy protection, cybersecurity, services computing, and building engineering. Dr. Fung is also a licensed Professional Engineer of software engineering in Ontario, Canada.
\end{IEEEbiography}

\begin{IEEEbiography}[{\includegraphics[width=1in,height=1.25in,clip,keepaspectratio]{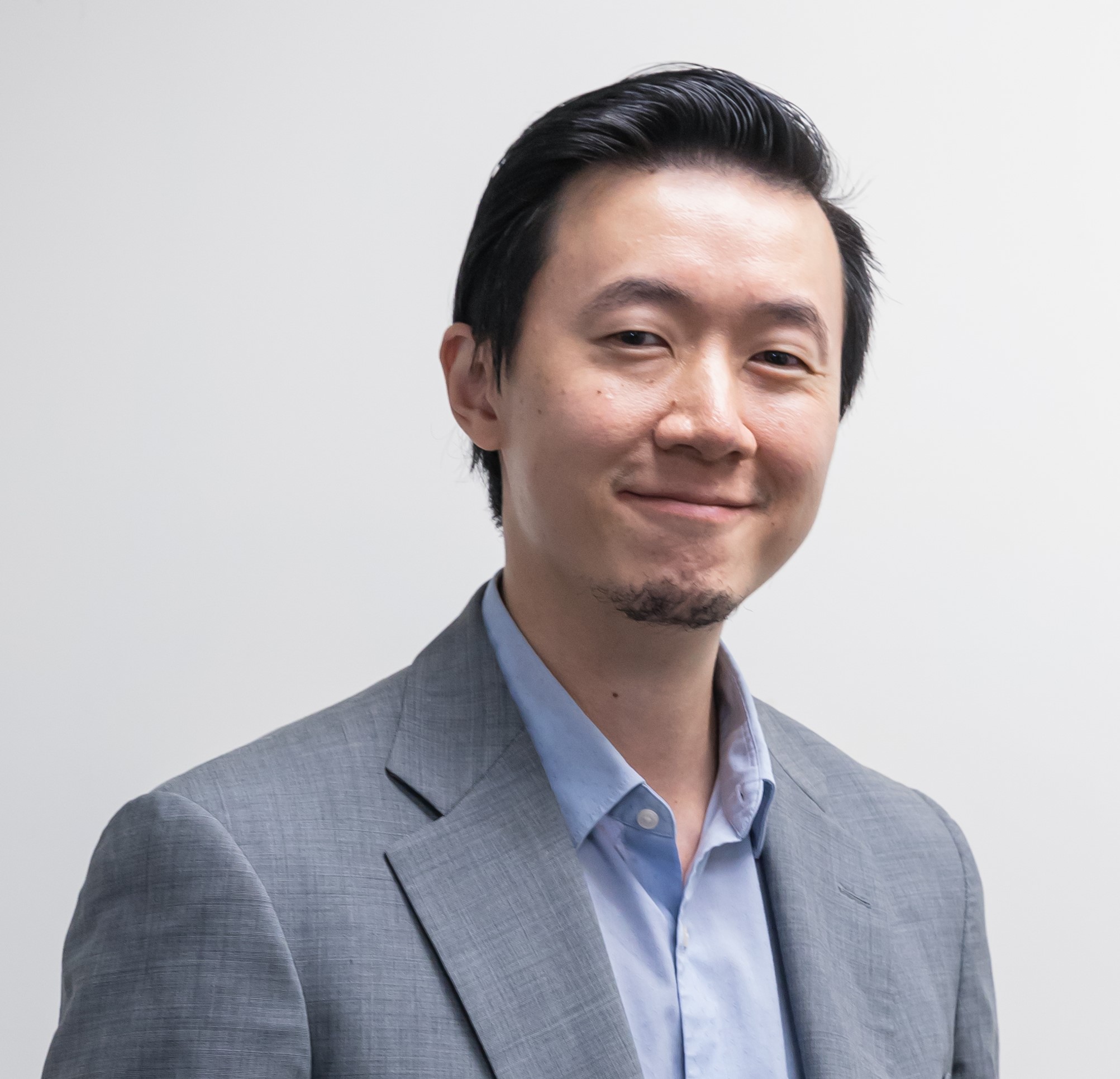}}]{Jonathan Poon} is a registered medical doctor and full time clinical informatics practitioner and advocate, currently working on the technical team responsible for Hong Kong national electronic-Health Record (eHR). He has managed several Digital Radiology and Laboratory Data sharing projects and eHealth projects enabling them to successfully achieve their objectives. He is heavily involved in engaging different levels of governance at healthcare organizations and fostering relationships for their sharing of clinical data, which is a key part of any national scale eHR initiative. Dr Poon has a Masters of Science in Genomics \& Bio-informatics from the Chinese University of Hong Kong, and has lectured in the field of bioinformatics / computational biology at the Polytechnic University of Hong Kong. He can be contacted at the Health Informatics Section, Division of IT \& HI via email: jonathan@ha.org.hk.
\end{IEEEbiography}

\clearpage

\appendices

\setcounter{table}{0}
\renewcommand{\thetable}{A\arabic{table}}

\setcounter{equation}{0}
\renewcommand{\theequation}{A\arabic{equation}}

\setcounter{figure}{0}
\renewcommand{\thefigure}{A\arabic{figure}}

\section{Optimization Procedures}
We summarize the optimization procedures in Algorithm~\ref{app:alg:chitf}. We implement the learning algorithm in PyTorch, and the gradients of the variables are computed by backpropagation on the fly.

Note that Eq. (16) involves computing the error function ($\operatorname{erf}$) of a standard Gaussian distribution, which is defiend in Eq. (17). Although computing $\operatorname{erf}$ requires numerical approximation, its derivative can be derived analytically as follows:
\begin{equation}
    \frac{d}{d x} \operatorname{erf}(x)=\frac{2}{\sqrt{\pi}} e^{-x^{2}},
\end{equation}
which follows immediately from the definition of $\operatorname{erf}(\cdot)$.

\begin{algorithm}[h]
\small
\caption{collective Hidden Interaction Tensor Factorization}\label{app:alg:chitf}
\SetAlgoLined
\SetKwInOut{Input}{Input}
\SetKwInOut{Output}{Output}
\Input{observation matrices: $\mathcal{V} = \{ \mathbf{V}^{(n)} \}_{n=1}^N$\\
       }
\Output{patient representation: $\mathbf{U}^{(s)}$\\
        phenotype definitions: $\{\mathbf{U}^{(n)}\}_{n=1}^N$}
 initialize $\mathbf{U}^{(s)}$ and $\{\mathbf{U}^{(n)}\}_{n=1}^N$ randomly\;
 \Repeat{converge or reach maximum number of iterations}{
      compute gradient w.r.t. $\mathbf{U}^{(s)}$:
      $$\nabla_{\mathbf{U}^{(s)}} \leftarrow \frac{\partial f}{\partial \mathbf{U}^{(s)}}$$\\
      update $\mathbf{U}^{(s)}$ by:~~$\mathbf{U}^{(s)} \leftarrow \mathbf{U}^{(s)} - \eta \nabla_{\mathbf{U}^{(s)}}$\;
      non-negative projection:$$\mathbf{U}^{(s)} \leftarrow \max\left\{\mathbf{0}, \mathbf{U}^{(s)}\right\}\;$$\\
  \For{$n=1:N$}{
    compute gradient w.r.t. $\mathbf{U}^{(n)}$: $$\nabla_{\mathbf{U}^{(n)}} \leftarrow \frac{\partial (f(\mathbf{U}^{(n)})+\Omega(\mathbf{U}^{(n)}))}{\partial \mathbf{U}^{(n)}}$$\\
    update $\mathbf{U}^{(n)}$ by:~~$\mathbf{U}^{(n)} \leftarrow \mathbf{U}^{(n)} - \eta \nabla_{\mathbf{U}^{(n)}}$\;
      non-negative projection:$$\mathbf{U}^{(n)} \leftarrow \max\left\{\mathbf{0}, \mathbf{U}^{(n)}\right\}\;$$\\
  }
   }
\end{algorithm}

\section{Distribution Selection, Hyperparameter Setting and Sensitivity Analysis}
\subsection{Distribution Selection}

Selecting appropriate distribution is crucial to accurately model the input data. A useful rule of thumb is to dedicate the choice of distributions based on the data type of the particular modality. For example, medications are recorded in form of non-negative counts; therefore, Poisson distribution is preferred over Gaussian distribution for the hidden interaction tensor of diagnoses and medications. On the other hand, Gaussian distribution is more suitable than Poisson distribution for the diagnosis-fluid hidden interaction tensor.

We also conduct experiments to validate our choice of distributions. We fit different modalities in MIMIC-III using Poisson and Gaussian distributions, respectively, and measure the AUPRC score for predicting in-hospital mortality. We run each experiment for five times and report the mean and standard deviation in Table~\ref{app:tab:distribution_selection}. From the results, it is obvious that Poisson distribution outperforms Gaussian distribution for counting data while Gaussian is better for real-valued input. It also shows that wrong distribution specification not only worsens the predictive performance, but also increases the uncertainty.

\begin{table}[h]
\centering
\caption{The AUPRC score for predicting in-hospital mortality of MIMIC-III using different modalities fitted with Poisson and Gaussian distributions}\label{app:tab:distribution_selection}
\begin{tabular}{cccc}
\toprule
 & Data Type & Poisson & Gaussian  \\\midrule
Dx \& Rx  & Counting  &  \textbf{0.46 (0.013)}  & 0.42 (0.031)	\\
Dx \& Lab  & Counting  &  \textbf{0.41 (0.011)}  & 0.39 (0.017)\\
Dx \& Fluid  & Real value  &  0.42 (0.058)  & \textbf{0.44 (0.029)} \\\bottomrule
\end{tabular}
\end{table}

\subsection{Hyperparameter Setting and Sensitivity Analysis}
\begin{table*}
    \centering
    \caption{Search spaces and optimal values of hyperparameters}\label{app:tab:hyperparameters}
    \begin{tabular}{clcc}\toprule
        Parameter & Description & Search Space & Optimal Value \\\midrule
        $\gamma$ & Strength of the elastic net regularization & \{1e-7, 5e-6, 1e-6, 5e-5, 1e-5, 5e-4, 1e-4, 5e-3, 1e-3\} & 1e-5 \\
        $\alpha$ & Strength of the $\ell_1$ term in elastic net & \{0, 0.1, 0.2, 0.3, 0.4, 0.5, 0.6, 0.7, 0.8, 0.9, 1\} & 0.7 \\
        $\beta$ & Strength of the angular regularization & \{ 1e-3, 1e-2, 5e-2, 0.1, 0.5, 1, 5, 10, 50, 100 \}& 1 \\
        $\theta_n$ & Angular penalization threshold & \{0, 0.1, 0.2, 0.3, 0.4, 0.5, 0.6, 0.7, 0.8, 0.9, 1\} & 0.5 \\
        $\sigma^2$ & Variance in Gaussian distribution & \{1e-11, 1e-10, 1e-9, 1e-8, 1e-7, 1e-6\} & 1e-9 \\
    \bottomrule\end{tabular}
\end{table*}

\begin{figure*}
    \centering
    \begin{tabular}{ccc}
    \includegraphics[width=.3\textwidth]{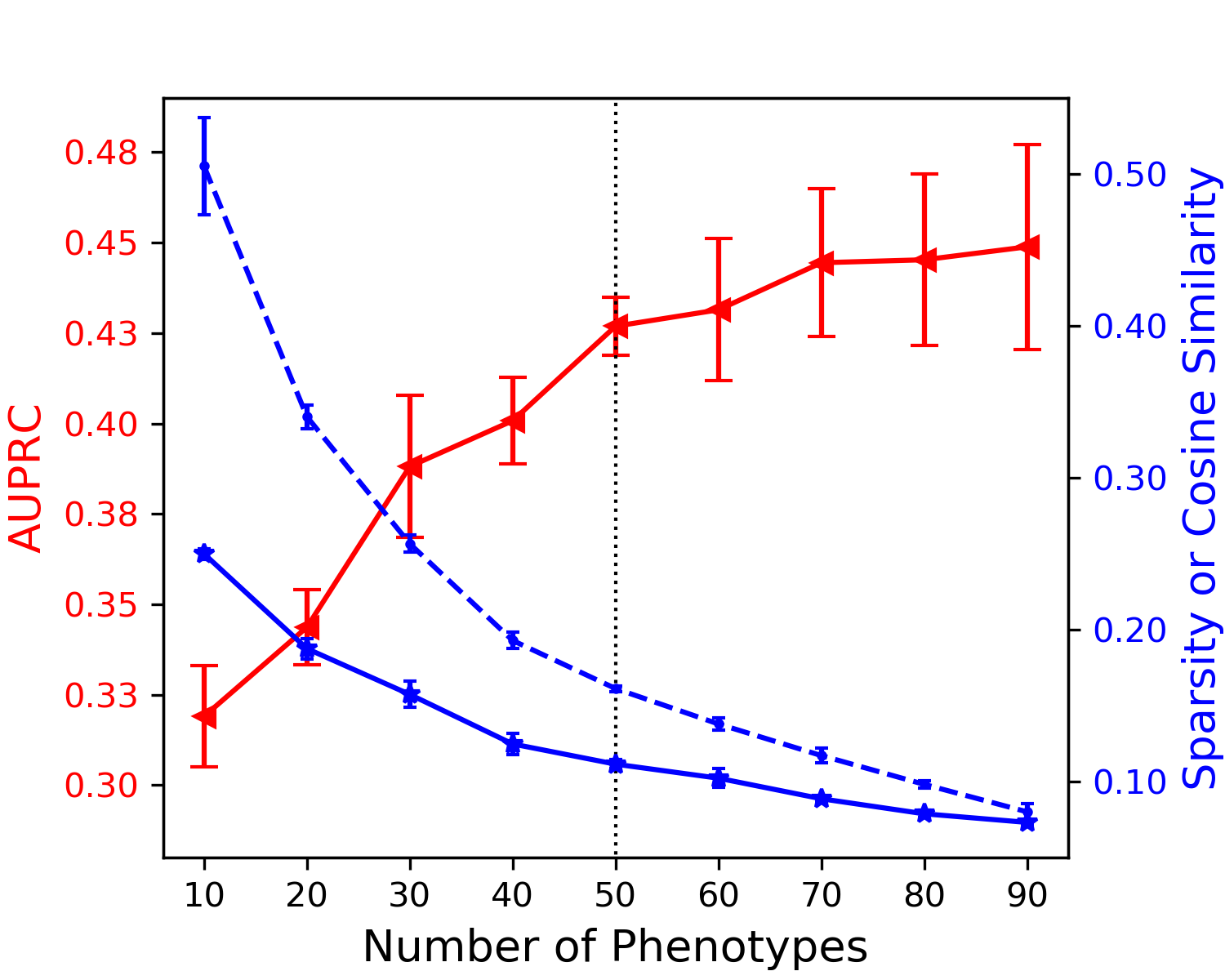} &
    \includegraphics[width=.3\textwidth]{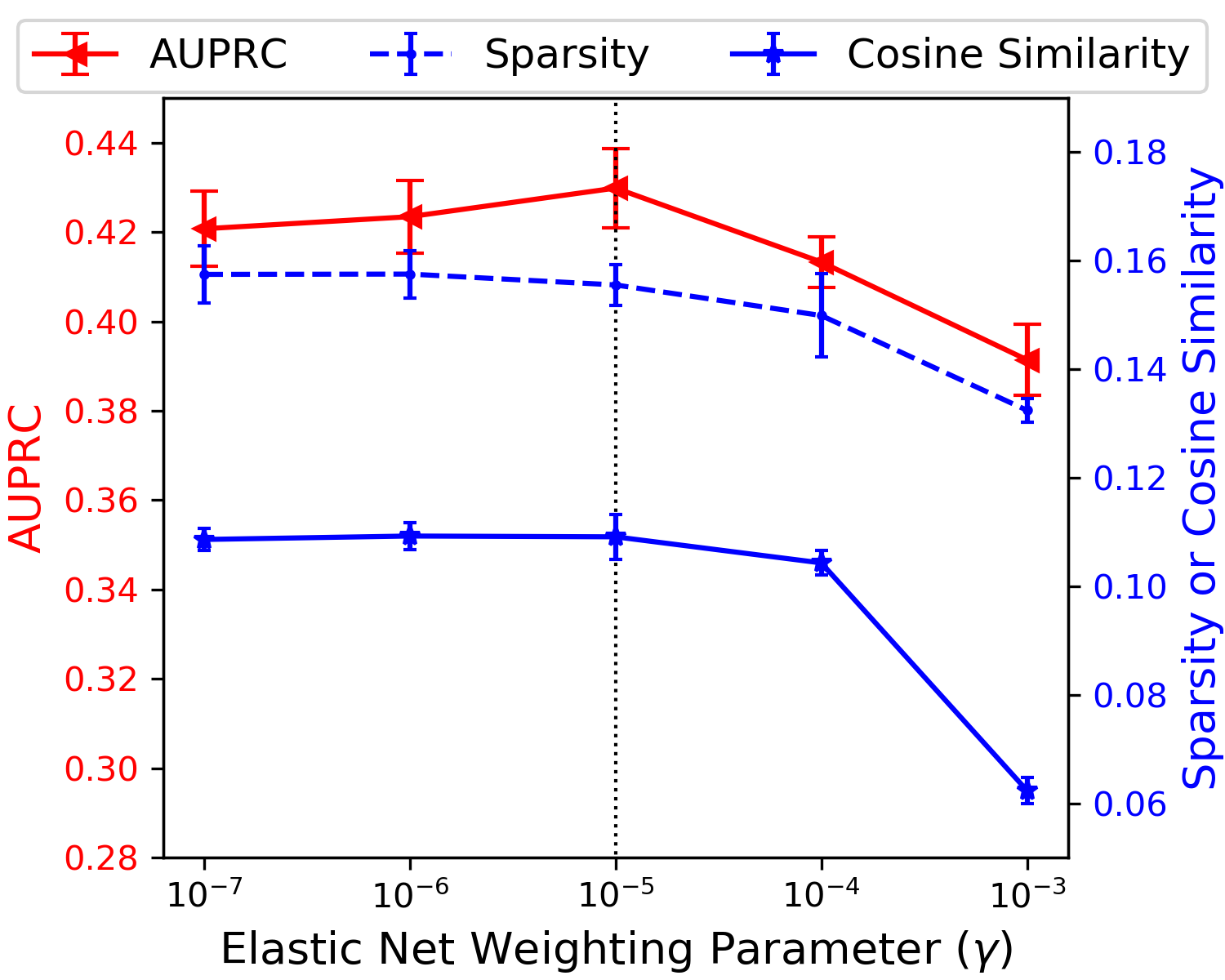} &
    \includegraphics[width=.3\textwidth]{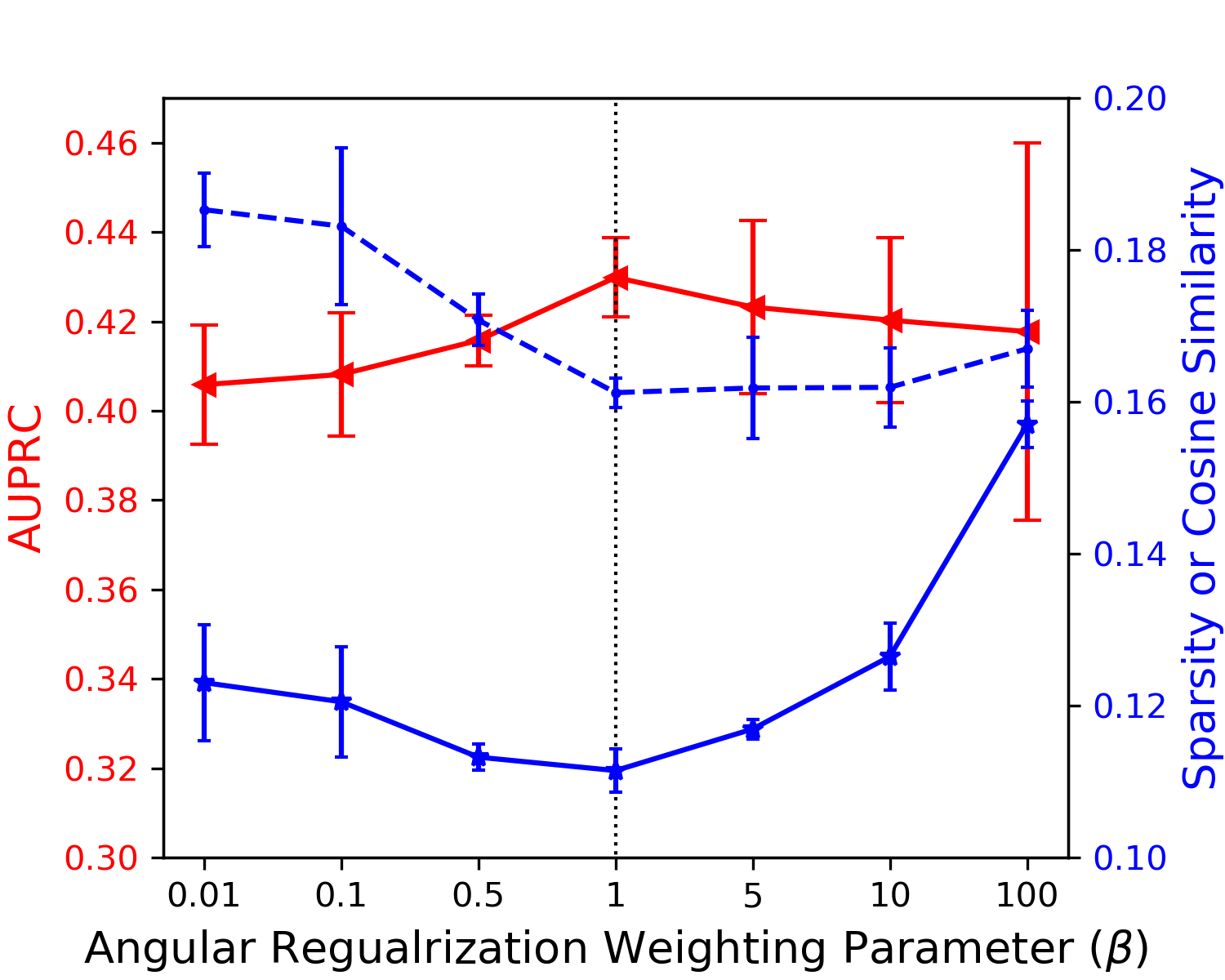} \\
    (a)  & (b) & (c)  \\[6pt]
    \end{tabular}
    \begin{tabular}{ccc}
    \includegraphics[width=.3\textwidth]{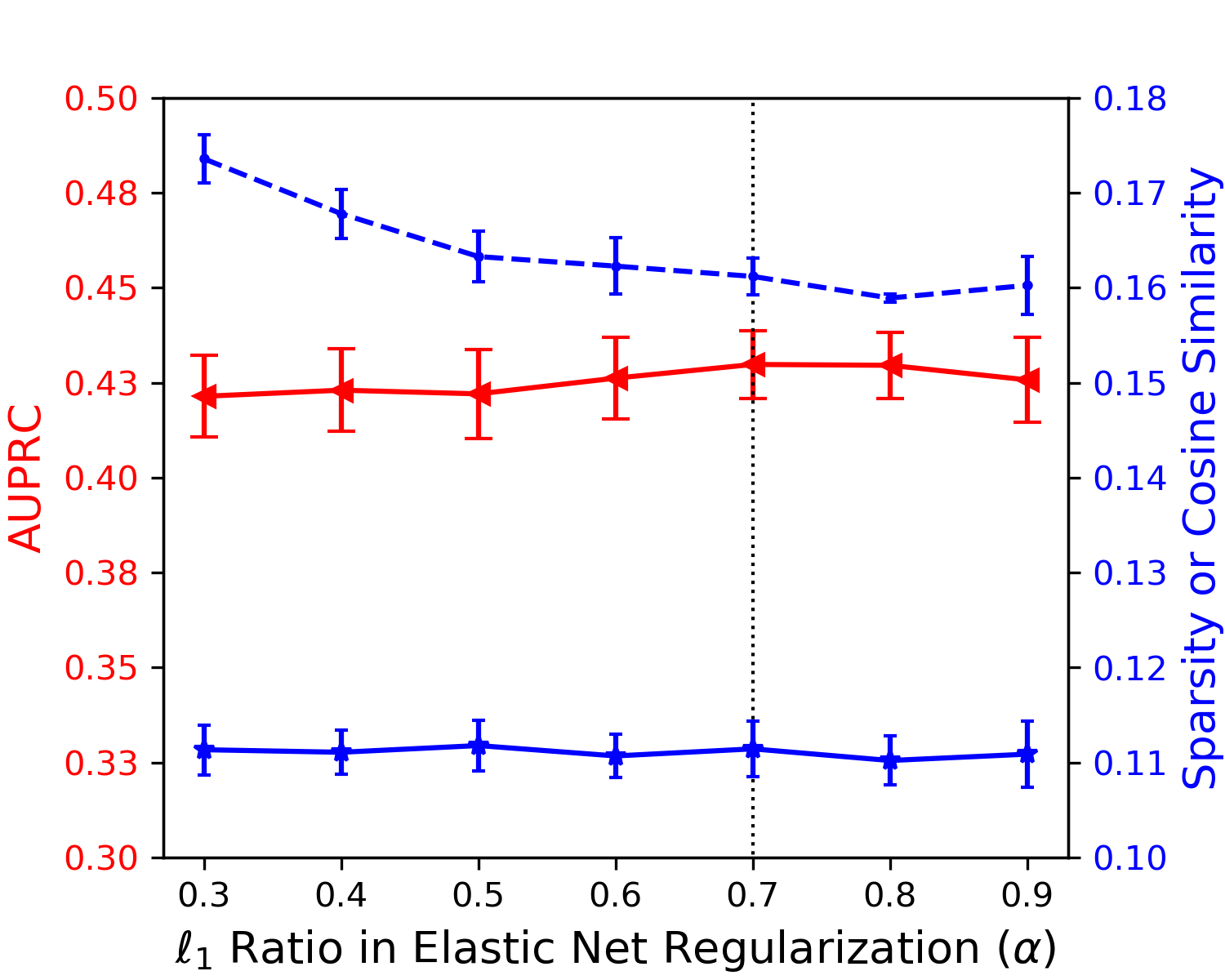} &
    \includegraphics[width=.3\textwidth]{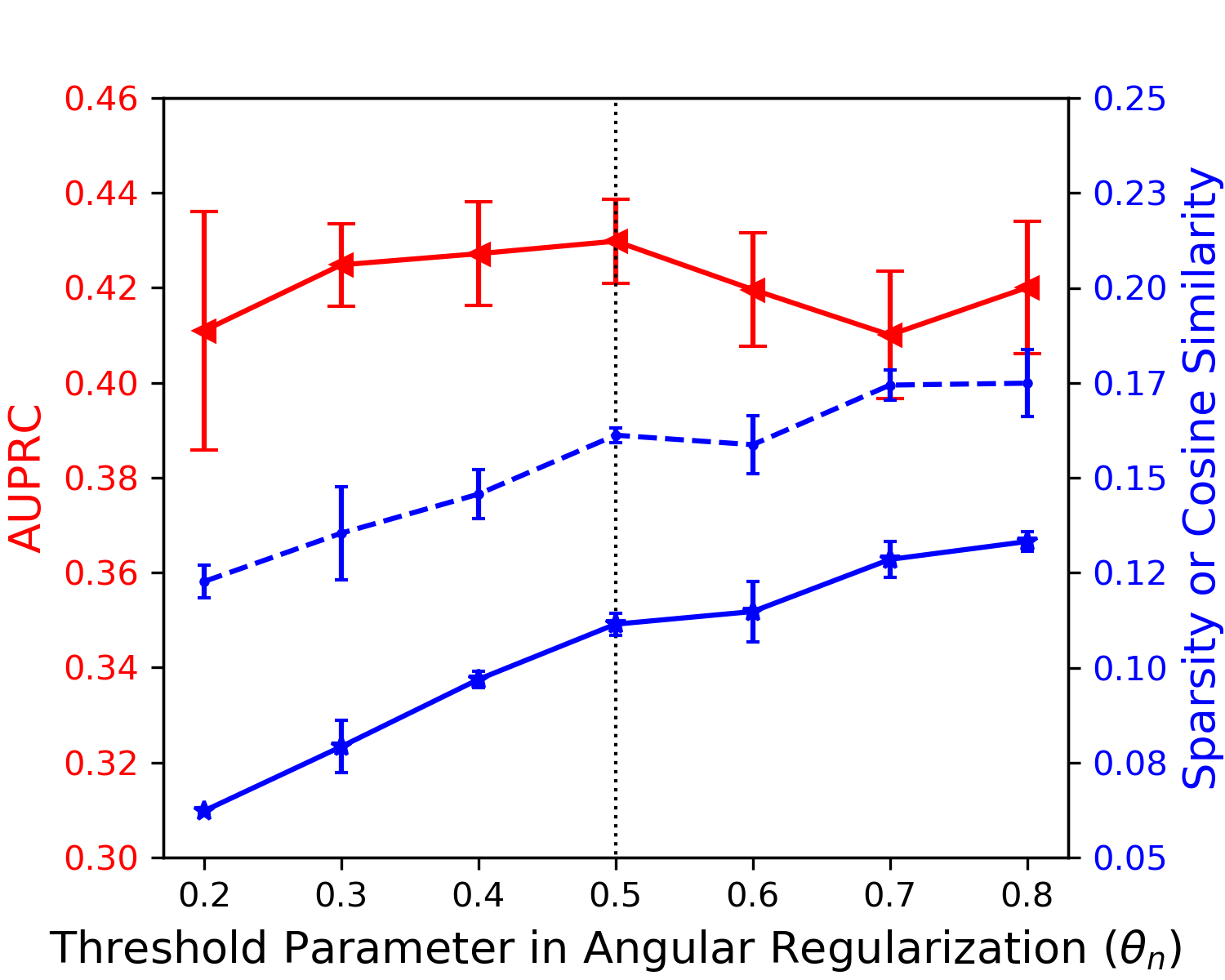} & \includegraphics[width=.3\textwidth]{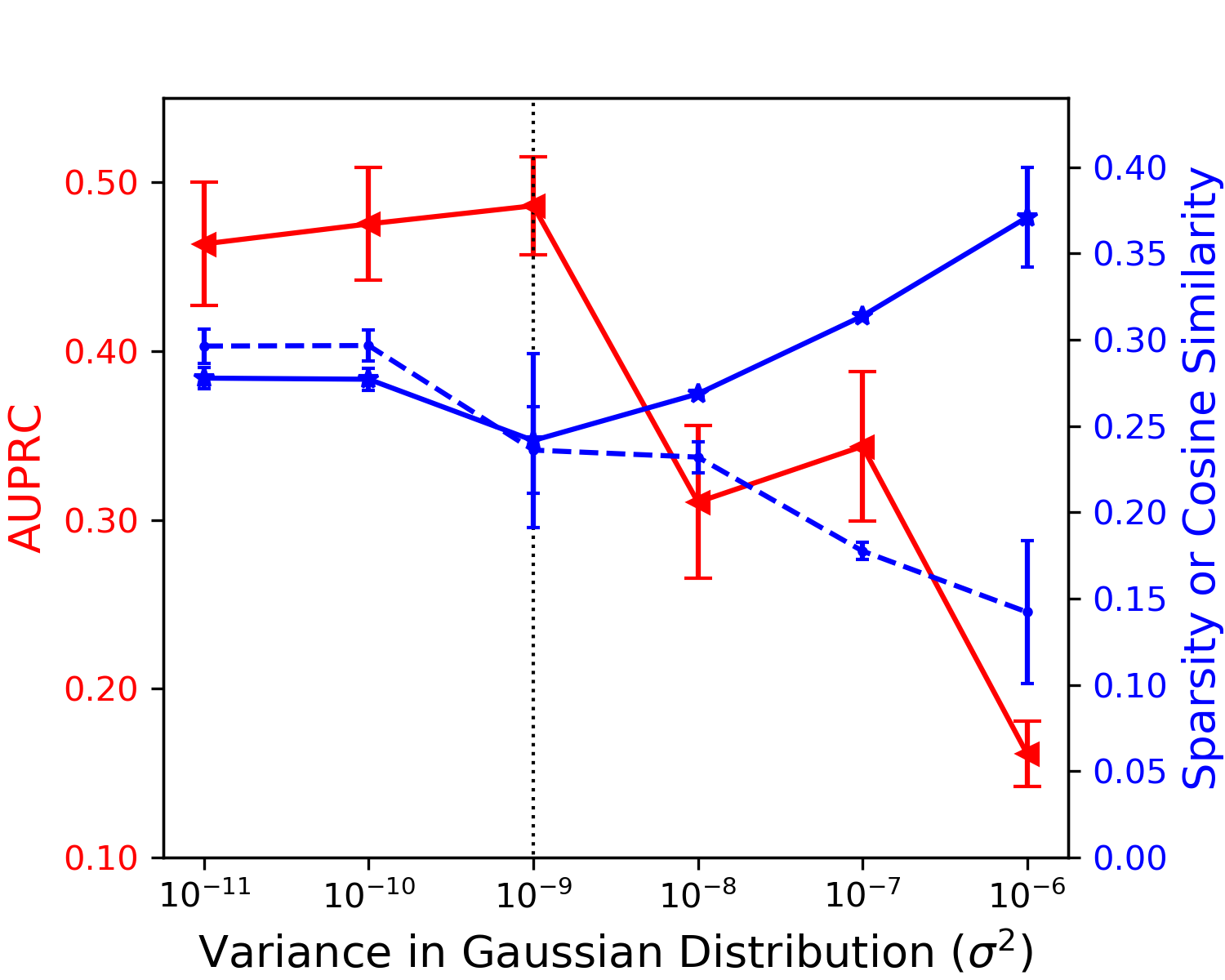}\\
    (d)  & (e)  & (f) \\[6pt]
    \end{tabular}
    \caption{Sensitivity of the prediction performance, sparsity and similarity to different hyperparameters. \textbf{(a)}: Number of phenotypes;  \textbf{(b)}: Weighting parameter of the elastic net regularization ($\gamma$); \textbf{(c)}: Weighting parameter of the angular regularization ($\beta$); \textbf{(d)}: The ratio of $\ell_1$ term in the elastic net regularization ($\alpha$); \textbf{(e)}: The threshold parameter in the angular regularization ($\theta_n$); \textbf{(e)}: The variance in the Gaussian distribution for input fluid modality ($\sigma^2$). The red line represent the in-hospital mortality prediction performance, the dotted blue line represent the sparsity, and the solid blue lines represent the cosine similarity. The vertical dotted black lines represent the final optimal value we used in the experiments.}
    \label{app:fig:sensitivity}
\end{figure*}

As described in Section 5.3, we tune the hyperparameters using grid search. We summarize the search spaces and the final optimal values of the hyperameters in Table~\ref{app:tab:hyperparameters}. We use the same set of hyperparameters for both datasets. We further analyze the sensitivity of prediction performance, and sparsity and similarity of the phenotypes inferred to the hyperparameter settings. To this end, we conduct experiments using the MIMIC-III subset containing modalities of diagnosis, medication and lab test, as described in Section 5.7. For each hyperparameter, we vary its value and measure the changes of the quantitative measures with all other hyperparameters fixed to their optimal values. The results are summarized in Fig.~\ref{app:fig:sensitivity}(a)-(e), where the red solid lines represent the prediction performance measured by AUPRC, the blue solid and dotted lines represent the cosine similarity and sparsity of the inferred phenotypes, respectively. The vertical black dotted lines represent the final optimal values for the corresponding hyperparameter.

\textbf{Number of Phenotypes}:  As shown in Fig~\ref{app:fig:sensitivity}(a), the prediction performance increases significantly with the increasing number of phenotypes, which is expected due to the increased number of parameters to represent the underlying patterns. Meanwhile, the sparsity and similarity also decrease. This indicates that a larger number of phenotypes would allow more distinct phenotypes to be discovered. However, with the number of phenotypes larger than 50, it becomes very time-consuming for clinicians to manually examine the quality of the phenotypes inferred.

\textbf{Elastic Net Weighting parameter ($\gamma$)}: Fig.~\ref{app:fig:sensitivity}(b) shows that the algorithm is not very sensitive to $\gamma$ when it takes value between $10^{-7}$ and $10^{-5}$ as the changes of AUPRC, sparsity and similarity are marginal. When $\gamma$ takes value larger or equal to $10^{-4}$, although the sparsity and similarity decreases significantly as the strength of regularization increases, the prediction performance also drops rapidly. This shows that a too-large $\gamma$ hurts the representational power of the model.

\textbf{Angular Regularization Weighing Parameter ($\beta$)}: Fig.~\ref{app:fig:sensitivity}(c) shows that $\beta$ significantly affects both prediction performance and interpretability of the model. When $\beta$ is set to 1, highest AUPRC and lowest sparsity and similarity are obtained simultaneously. When $\beta$ is larger than 1, the similarity surprisingly increases by a wide margin.

\textbf{$\ell_1$ Ratio in Elastic Net Regularization ($\alpha$)}: Fig.~\ref{app:fig:sensitivity}(d) shows that the prediction performance and the similarity are not sensitive to $\alpha$ as their changes over different values of $\alpha$ are quite small. However, it does have an impact on the sparsity, which reduces significantly with $\alpha$ increasing from 0.3 to 0.8.

\textbf{Threshold Parameter in Angular Regularization ($\theta_n$)}: Fig.~\ref{app:fig:sensitivity}(e) shows that $\theta_n$ impacts both prediction and interpretability. Note that smaller threshold implies stronger regularization because factors having pairwise cosine similarity larger than the threshold is penalized by the angular regularization. The highest AUPRC is obtained when $\theta_n$ takes value of 0.5. If $\theta_n$ is further decreased, sparsity and similarity improves due to stronger regularization. However, the prediction performance decrease as well.

\textbf{Variance in Gaussian Distribution ($\sigma^2$)}: Fig.~\ref{app:fig:sensitivity}(f) is obtained using another subset of MIMIC-III containing the modality of input fluid as well. In general, smaller variance of the Gaussian distribution tends to force the Gaussian distribution to concentrate its density very close to its mean value. Our experiments reveal that best prediction performance is obtained when $\sigma^2$ is set to $10^{-9}$. Meanwhile, the most distinct phenotypes (smallest similarity) are also obtained.

\section{Convergence of the Learning Algorithm}
\begin{figure}
    \centering
    \includegraphics[width=0.95\columnwidth]{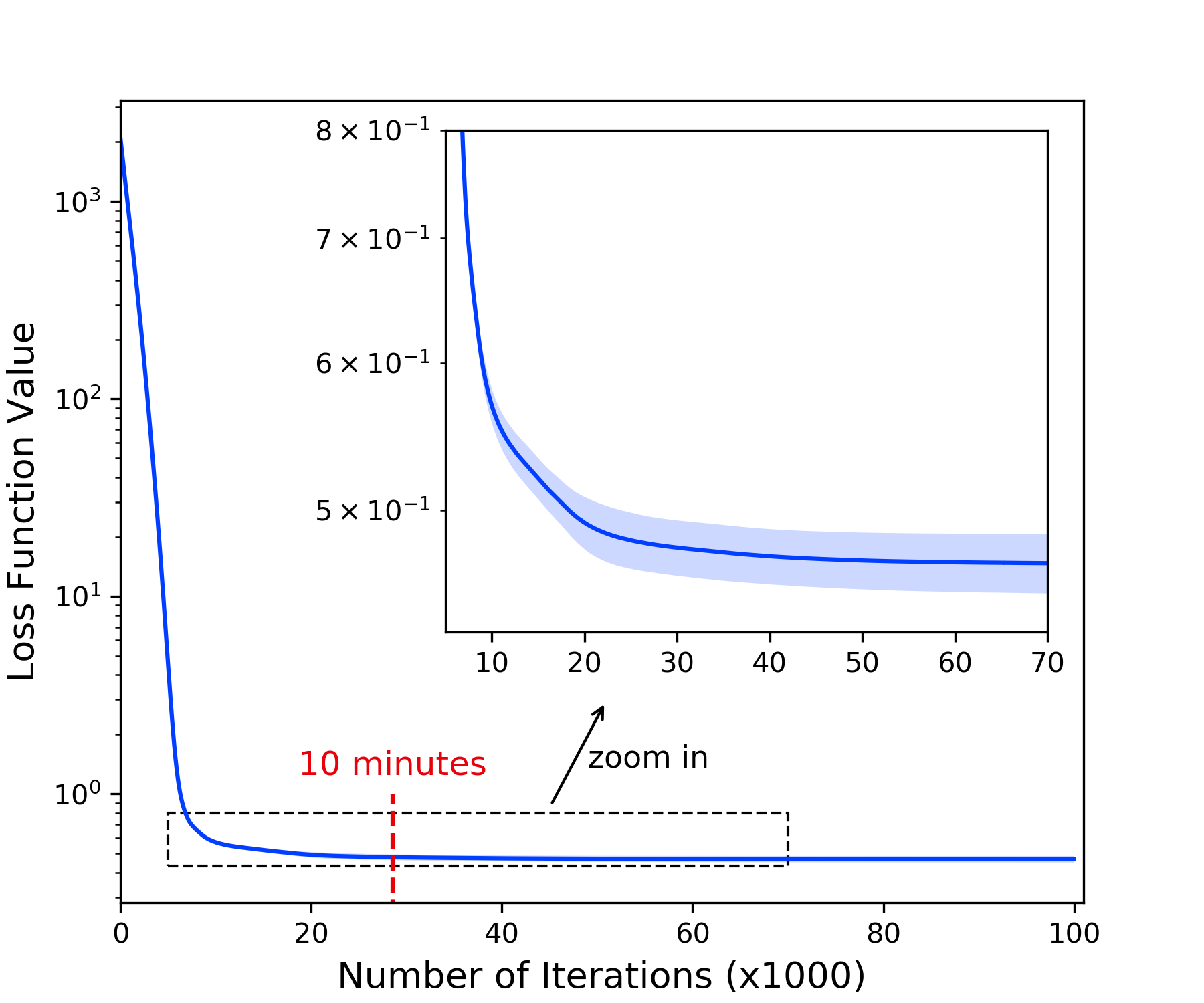}
    \caption{Convergence curve of cHITF obtained by running the algorithm with ten different train/test data split and random initialization. The loss function value does not include regularization terms.}
    \label{app:fig:convergence}
\end{figure}

As described in Section 4.4, the BCD optimization framework we used to learn the parameters has previously been shown to converge with a sublinear rate. Here we provide empirical evidence on its convergence. Fig.~\ref{app:fig:convergence} shows the convergence curve of  our algorithm on the MIMIC-III subset with the modalities of diagnosis, medication and lab test. We run the algorithm for 100k iterations with ten different train/test data split and random initialization, and the dark blue line in Fig.~\ref{app:fig:convergence} shows the average of the loss function values excluding the regularization terms over the ten runs. The area filled in light blue indicates the standard deviation. Fig.~\ref{app:fig:convergence} clearly shows that the learning algorithm converges fast on the MIMIC-III subset.

\section{Extended Results}

\subsection{Sparsity and Diversity of Phenotypes from eICU Dataset}
\begin{table}[]
\renewcommand{\arraystretch}{1.2}
\centering
\caption{Sparsity and diversity of phenotypes inferred from eICU dataset
}
\label{app:tab:diversity_eicu}
\begin{tabular}{ccccc}
\toprule
 & Reg. & Sparsity & \begin{tabular}[c]{@{}c@{}}Cosine\\ Similarity\end{tabular} & Jaccard@10  \\ \midrule
CP-APR  & --  & 0.62 (0.004)  & 0.59 (0.003)   &   0.12 (0.002) \\
Marble & --    & 0.67 (0.007)   & 0.75 (0.001)   &   0.10 (0.003) \\
Rubik  & --    & 0.84 (0.007)   & 0.72 (0.010)   &   0.15 (0.004) \\
Granite & --   & 0.98 (0.002)   & 0.87 (0.001)  &   0.14 (0.001) \\\hline
SiCNMF & --    & \textbf{0.05 (0.001)}   & 0.27 (0.002)   &   0.09 (0.001) \\
cHITF & Both & 0.07 (0.003) & \textbf{0.04 (0.001)}   &   \textbf{0.03 (0.002)}  \\\hline
cHITF & Neither & 0.17 (0.003) & 0.16 (0.006) & 0.06 (0.002)\\
cHITF & Elastic net & 0.11 (0.004) & 0.15 (0.005) & 0.06 (0.001)\\
cHITF & Angular & 0.14 (0.002) &  0.11 (0.001) & 0.04 (0.001)\\
\bottomrule
\end{tabular}
\vspace{2pt} \begin{flushleft}\textit{\textbf{Reg.}: abbreviation of ``Regularization'' indicating the active regularization term(s).}\end{flushleft}
\end{table}

We summarize the sparsity and diversity of the phenotypes inferred by cHITF and baselines in Table~\ref{app:tab:diversity_eicu}. The results follow similar patterns to that obtained from MIMIC-III, as shown in Table 5. Specifically, SiCNMF obtains the best sparsity, followed by cHITF with a small margin. cHITF obtains the best cosine similarity and Jaccard@10, suggesting that cHITF infers the most distinct and interpretable phenotypes. Besides, the elastic net and angular regularization terms both help improve the interpretability.

\subsection{Inter-Modal Correspondence}\label{app:correspondence}
Due to space limit, we only exhibited some representative examples of the inferred diagnosis-medication and diagnosis-lab-test correspondence in Table~2 and Table~3. Here we provide the full list of the top five corresponding medications and lab tests inferred by all models for all of the ten diagnoses in Table~\ref{app:tab:correspondence_dxrx} and Table~\ref{app:tab:correspondence_dxlab}, respectively.

\begin{table*}
\renewcommand{\arraystretch}{1.1}
  \centering
  \caption{The Diagnosis-Medication Correspondence Inferred by HITF and the Baselines.}\label{app:tab:correspondence_dxrx}
\begin{tabular}{cl}\toprule

& \multicolumn{1}{c}{\textbf{Dx1: Cardiac dysrhythmias}}\\\hline
HITF  &  \textcolor{red}{\textbf{Metoprolol (0.17)}}; ~\textcolor{red}{\textbf{Metoprolol Tartrate (0.12)}}; ~\textcolor{red}{\textbf{Warfarin (0.11)}}; ~\textcolor{red}{\textbf{Amiodarone (0.11)}}; ~Heparin Sodium (0.10).\\
CP-APR  &  \textcolor{blue}{\textit{Potassium Chloride (0.03)}}; ~Acetaminophen (0.03); ~\textcolor{red}{\textbf{Magnesium Sulfate (0.02)}}; ~Insulin (0.02); ~\textcolor{red}{\textbf{Furosemide (0.02)}}.\\
Marble  &  Acetaminophen (0.03); ~\textcolor{blue}{\textit{Potassium Chloride (0.03)}}; ~\textcolor{red}{\textbf{Magnesium Sulfate (0.02)}}; ~Insulin (0.02); ~\textcolor{red}{\textbf{Furosemide (0.02)}}.\\
Rubik  &  \textcolor{blue}{\textit{Potassium Chloride (0.02)}}; ~{Acetaminophen (0.02)}; {Insulin (0.02)}; ~\textcolor{red}{\textbf{Magnesium Sulfate (0.02)}}; ~\textcolor{red}{\textbf{Furosemide (0.02)}}.\\
Granite  & Acetaminophen (0.02); ~\textcolor{blue}{\textit{Potassium Chloride (0.01)}}; ~\textcolor{red}{\textbf{Magnesium Sulfate (0.01)}}; ~Insulin (0.01); ~Sodium Chloride 0.9\%  Flush (0.01).\\\toprule

& \multicolumn{1}{c}{\textbf{Dx2: Heart failure}}\\\hline
HITF  &  \textcolor{red}{\textbf{Furosemide (0.56)}}; ~{Potassium Chloride (0.23)}; {Magnesium Sulfate (0.03)}; ~{Prednisone (0.02)}.\\
CP-APR  &  Acetaminophen (0.03); ~\textcolor{red}{\textbf{Furosemide (0.03)}}; ~Potassium Chloride (0.03); ~Insulin (0.02); ~Magnesium Sulfate (0.02).\\
Marble  &  Acetaminophen (0.03); ~Potassium Chloride (0.03); ~Insulin (0.02); ~\textcolor{red}{\textbf{Furosemide (0.02)}}; ~Magnesium Sulfate (0.02).\\
Rubik  &  {Potassium Chloride (0.02)}; ~{Acetaminophen (0.02)}; {Insulin (0.02)}; ~{Magnesium Sulfate (0.02)}; ~\textcolor{red}{\textbf{Furosemide (0.02)}}.\\
Granite  & Acetaminophen (0.02); ~Potassium Chloride (0.01); ~Magnesium Sulfate (0.01); ~Insulin (0.01); ~Sodium Chloride 0.9\%  Flush (0.01).\\\toprule

& \multicolumn{1}{c}{\textbf{Dx3: Other forms of chronic ischemic heart disease}}\\\hline
HITF  &  Furosemide (0.07); ~Acetaminophen (0.06); ~\textcolor{red}{\textbf{Aspirin EC (0.06)}}; ~\textcolor{red}{\textbf{Metoprolol Tartrate (0.05)}}; ~Potassium Chloride (0.05).\\
CP-APR  &  Acetaminophen (0.03); ~Potassium Chloride (0.03); ~Magnesium Sulfate (0.02); ~Docusate Sodium (0.02); ~Furosemide (0.02).\\
Marble  &  Acetaminophen (0.03); ~Potassium Chloride (0.03); ~Docusate Sodium (0.02); ~Magnesium Sulfate (0.02); ~Furosemide (0.02).\\
Rubik  &  Potassium Chloride (0.02); ~Acetaminophen (0.02); ~Magnesium Sulfate (0.02); ~Insulin (0.02); ~Docusate Sodium (0.02).\\
Granite  & Acetaminophen (0.02); ~Potassium Chloride (0.02); ~Magnesium Sulfate (0.01); ~Insulin (0.01); ~Sodium Chloride 0.9\%  Flush (0.01).\\\toprule

& \multicolumn{1}{c}{\textbf{Dx4: Diabetes
mellitus}}\\\hline
HITF  &  \textcolor{red}{\textbf{Insulin (0.88)}}; ~\textcolor{red}{\textbf{Insulin Human Regular (0.05)}}; \textcolor{red}{\textbf{Dextrose 50\% (0.01)}}; ~\textcolor{red}{\textbf{Metformin (0.01)}}.\\
CP-APR  &  \textcolor{red}{\textbf{Insulin (0.03)}}; ~Acetaminophen (0.03); ~Potassium Chloride (0.03); ~Sodium Chloride 0.9\%  Flush (0.02); ~Magnesium Sulfate (0.02).\\
Marble  &  \textcolor{red}{\textbf{Insulin (0.03)}}; ~Acetaminophen (0.03); ~Potassium Chloride (0.03); ~Sodium Chloride 0.9\%  Flush (0.03); ~Heparin (0.02).\\
Rubik  &  Acetaminophen (0.02); ~Potassium Chloride (0.02); ~\textcolor{red}{\textbf{Insulin (0.02)}}; ~Magnesium Sulfate (0.02); ~Sodium Chloride 0.9\%  Flush (0.02).\\
Granite  & Acetaminophen (0.02); ~Potassium Chloride (0.01); ~\textcolor{red}{\textbf{Insulin (0.01)}}; ~Sodium Chloride 0.9\%  Flush (0.01); ~Magnesium Sulfate (0.01).\\\toprule

& \multicolumn{1}{c}{\textbf{Dx5: Disorders of fluid electrolyte and acid-base balance}}\\\hline
HITF  &  \textcolor{red}{\textbf{Potassium Chloride (0.49)}}; ~\textcolor{red}{\textbf{Magnesium Sulfate (0.17)}}; ~Metoprolol (0.12); ~\textcolor{red}{\textbf{Calcium Gluconate (0.05)}}; ~\textcolor{red}{\textbf{Potassium Phosphate (0.05)}}.\\
CP-APR  &  \textcolor{red}{\textbf{Potassium Chloride (0.03)}}; ~Acetaminophen (0.03); ~Heparin (0.03); ~\textcolor{red}{\textbf{Insulin (0.03)}}; ~\textcolor{red}{\textbf{Magnesium Sulfate (0.02)}}.\\
Marble  &  Acetaminophen (0.03); ~\textcolor{red}{\textbf{Potassium Chloride (0.03)}}; ~Heparin (0.03); ~\textcolor{red}{\textbf{Insulin (0.03)}}; ~Sodium Chloride 0.9\%  Flush (0.02).\\
Rubik  &  \textcolor{red}{\textbf{Potassium Chloride (0.02)}}; ~Acetaminophen (0.02); ~\textcolor{red}{\textbf{Insulin (0.02)}}; ~\textcolor{red}{\textbf{Magnesium Sulfate (0.02)}}; ~Sodium Chloride 0.9\%  Flush (0.02).\\
Granite  & Acetaminophen (0.01); ~\textcolor{red}{\textbf{Potassium Chloride (0.01)}}; ~\textcolor{red}{\textbf{Magnesium Sulfate (0.01)}}; ~Sodium Chloride 0.9\%  Flush (0.01); ~\textcolor{red}{\textbf{Insulin (0.01)}}.\\\toprule

& \multicolumn{1}{c}{\textbf{Dx6: Bacterial infection in conditions classified elsewhere and of unspecified site}}\\\hline
HITF  &  \textcolor{red}{\textbf{Vancomycin (0.67)}}; ~\textcolor{red}{\textbf{Piperacillin-Tazobactam Na (0.15)}}; ~\textcolor{red}{\textbf{Gentamicin (0.08)}}; ~Heparin Flush CVL (0.08); ~\textcolor{blue}{\textit{Miconazole Powder 2\% (0.02)}}.\\
CP-APR  &  Acetaminophen (0.03); ~Potassium Chloride (0.03); ~Insulin (0.02); ~Magnesium Sulfate (0.02); ~Heparin (0.02).\\
Marble  &  Potassium Chloride (0.03); ~Acetaminophen (0.03); ~Magnesium Sulfate (0.02); ~Heparin (0.02); ~Insulin (0.02).\\
Rubik  &  Potassium Chloride (0.03); ~Acetaminophen (0.02); ~Insulin (0.02); ~Magnesium Sulfate (0.02); ~Sodium Chloride 0.9\%  Flush (0.02).\\
Granite  & Acetaminophen (0.02); ~Potassium Chloride (0.01); ~Magnesium Sulfate (0.01); ~Sodium Chloride 0.9\%  Flush (0.01); ~Insulin (0.01).\\\toprule

& \multicolumn{1}{c}{\textbf{Dx7: Iron deficiency anemias}}\\\hline
HITF  &  \textcolor{red}{\textbf{Pantoprazole Sodium (0.33)}}; ~Sodium Chloride 0.9\%  Flush (0.27); ~\textcolor{red}{\textbf{Pantoprazole (0.21)}}; ~Acetaminophen (0.09); ~Heparin (0.03).\\
CP-APR  &  Acetaminophen (0.03); ~Potassium Chloride (0.03); ~Insulin (0.02); ~Magnesium Sulfate (0.02); ~Heparin (0.02).\\
Marble  &  Acetaminophen (0.03); ~Potassium Chloride (0.03); ~Heparin (0.02); ~Sodium Chloride 0.9\%  Flush (0.02); ~Insulin (0.02).\\
Rubik  & Acetaminophen (0.02); ~Potassium Chloride (0.02); ~Insulin (0.02); ~Sodium Chloride 0.9\%  Flush (0.02); ~Magnesium Sulfate (0.02).\\
Granite  & Acetaminophen (0.02); ~Potassium Chloride (0.01); ~Insulin (0.01); ~Magnesium Sulfate (0.01); ~Sodium Chloride 0.9\%  Flush (0.01).\\\toprule

& \multicolumn{1}{c}{\textbf{Dx8: Chronic bronchitis}}\\\hline
HITF  &  \textcolor{red}{\textbf{PredniSONE (0.76)}}; ~\textcolor{red}{\textbf{MethylPREDNISolone Sodium Succ (0.22)}}; ~Pantoprazole (0.01).\\
CP-APR  &  Heparin (0.03); ~Potassium Chloride (0.03); ~Insulin (0.03); ~Acetaminophen (0.03); ~Sodium Chloride 0.9\%  Flush (0.03).\\
Marble  &  Heparin (0.03); ~Acetaminophen (0.03); ~Insulin (0.03); ~Sodium Chloride 0.9\%  Flush (0.03); ~Potassium Chloride (0.03).\\
Rubik  & Acetaminophen (0.02); ~Potassium Chloride (0.02); ~Insulin (0.02); ~Heparin (0.02); ~Sodium Chloride 0.9\%  Flush (0.02).\\
Granite  & Acetaminophen (0.02); ~Potassium Chloride (0.01); ~Magnesium Sulfate (0.01); ~Insulin (0.01); ~Furosemide (0.01).\\\toprule

& \multicolumn{1}{c}{\textbf{Dx9: Arterial embolism and thrombosis}}\\\hline
HITF  &  {Sodium Bicarb (0.36)}; ~\textcolor{red}{\textbf{Enoxaparin Sodium (0.33)}}; ~\textcolor{red}{\textbf{Isosorbide Mononitrate (0.22)}}; ~Tacrolimus (0.03); ~Mycophenolate Mofetil (0.01).\\
CP-APR  &  Potassium Chloride (0.03); ~Magnesium Sulfate (0.03); ~Acetaminophen (0.02); ~Insulin (0.02); ~\textcolor{red}{\textbf{Heparin (0.02)}}.\\
Marble  &  Potassium Chloride (0.03); ~Magnesium Sulfate (0.03); ~Insulin (0.02); ~Acetaminophen (0.02); ~Pantoprazole Sodium (0.02).\\
Rubik  & Potassium Chloride (0.02); ~Acetaminophen (0.02); ~Insulin (0.02); ~Magnesium Sulfate (0.02); ~Furosemide (0.02).\\
Granite  & Potassium Chloride (0.01); ~Acetaminophen (0.01); ~Magnesium Sulfate (0.01); ~Insulin (0.01); ~Furosemide (0.01).\\\toprule

&\multicolumn{1}{c}{\begin{tabular}[c]{@{}l@{}} \textbf{Dx10: Symptoms involving nervous and musculoskeletal systems} \end{tabular}} \\\hline
HITF  &  \textcolor{red}{\textbf{Dexamethasone (0.27)}}; ~Bisacodyl (0.18); ~Senna (0.13); ~Docusate Sodium (0.12); ~Sodium Chloride 0.9\%  Flush (0.11).\\
CP-APR  &  Acetaminophen (0.03); ~Insulin (0.03); ~Potassium Chloride (0.03); ~Sodium Chloride 0.9\%  Flush (0.03); ~Magnesium Sulfate (0.03).\\
Marble  &  Potassium Chloride (0.03); ~Insulin (0.03); ~Acetaminophen (0.03); ~Magnesium Sulfate (0.03); ~Heparin (0.03).\\
Rubik  & Potassium Chloride (0.03); ~Acetaminophen (0.03); ~Insulin (0.02); ~Magnesium Sulfate (0.02); ~Heparin (0.02).\\
Granite  & Acetaminophen (0.01); ~Insulin (0.01); ~Potassium Chloride (0.01); ~Sodium Chloride 0.9\%  Flush (0.01); ~Magnesium Sulfate (0.01).\\\bottomrule

\end{tabular}
\end{table*}

\begin{table*}
\renewcommand{\arraystretch}{1.1}
  \centering
  \caption{The Diagnosis-Lab-Test Correspondence Inferred by HITF and the Baselines.}\label{app:tab:correspondence_dxlab}
\begin{tabular}{cl}\toprule

& \multicolumn{1}{c}{\textbf{Dx1: Cardiac dysrhythmias}}\\\hline
HITF  & PT [B] (0.19); ~Hematocrit [B] (0.16); ~Red Blood Cells [B] (0.12); ~Hemoglobin [B] (0.12); ~Urea Nitrogen [B] (0.07). \\
CP-APR  & Red Blood Cells [B] (0.03); ~Glucose [B] (0.03); ~Hemoglobin [B] (0.03); ~Hematocrit [B] (0.03); ~PT [B] (0.03). \\
Marble  & Red Blood Cells [B] (0.03); ~Glucose [B] (0.03); ~Hemoglobin [B] (0.03); ~Hematocrit [B] (0.03); ~PT [B] (0.03). \\
Rubik  & Red Blood Cells [B] (0.03); ~Hemoglobin [B] (0.03); ~Hematocrit [B] (0.03); ~Glucose [B] (0.03); ~PT [B] (0.03). \\
Granite  & Hemoglobin [B] (0.01); ~Glucose [B] (0.01); ~Calcium, Total [B] (0.01); ~Red Blood Cells [B] (0.01); ~Hematocrit [B] (0.01).\\\toprule

& \multicolumn{1}{c}{\textbf{Dx2: Heart failure}}\\\hline
HITF  & {PT [B] (0.16)}; ~~\textcolor{red}{\textbf{Hematocrit [B] (0.14)}}; ~\textcolor{red}{\textbf{Hemoglobin [B] (0.13)}}; ~\textcolor{red}{\textbf{Red Blood Cells [B] (0.13)}}; ~{Urea Nitrogen [B] (0.11)}. \\
CP-APR  & Glucose [B] (0.03); ~~\textcolor{red}{\textbf{Hemoglobin [B] (0.03)}}; ~~\textcolor{red}{\textbf{Red Blood Cells [B] (0.03)}}; ~~\textcolor{red}{\textbf{Hematocrit [B] (0.03)}}; ~Urea Nitrogen [B] (0.03). \\
Marble  & Glucose [B] (0.03); ~~\textcolor{red}{\textbf{Hemoglobin [B] (0.03)}}; ~~\textcolor{red}{\textbf{Red Blood Cells [B] (0.03)}}; ~~\textcolor{red}{\textbf{Hematocrit [B] (0.03)}}; ~Urea Nitrogen [B] (0.03). \\
Rubik  & {Red Blood Cells [B] (0.03)}; ~\textcolor{red}{\textbf{Hemoglobin [B] (0.03)}}; ~\textcolor{red}{\textbf{Hematocrit [B] (0.03)}}; ~{Glucose [B] (0.03)}; {Urea Nitrogen [B] (0.03)}. \\
Granite  & ~\textcolor{red}{\textbf{Hemoglobin [B] (0.01)}}; ~Glucose [B] (0.01); ~Chloride [B] (0.01); ~~\textcolor{red}{\textbf{Calcium, Total [B] (0.01)}}; ~~\textcolor{red}{\textbf{Red Blood Cells [B] (0.01)}}.\\\toprule

& \multicolumn{1}{c}{\textbf{Dx3: Other forms of chronic ischemic heart disease}}\\\hline
HITF  & \textcolor{blue}{\textit{pO2 [B] (0.24)}}; ~\textcolor{blue}{\textit{Glucose [B] (0.15)}}; ~\textcolor{blue}{\textit{Hematocrit [B] (0.10)}}; ~pH [B] (0.09); ~\textcolor{blue}{\textit{Hemoglobin [B] (0.09)}}. \\
CP-APR  & \textcolor{blue}{\textit{Hemoglobin [B] (0.04)}}; ~\textcolor{blue}{\textit{Hematocrit [B] (0.04)}}; ~Red Blood Cells [B] (0.04); ~\textcolor{blue}{\textit{Glucose [B] (0.04)}}; ~PT [B] (0.03). \\
Marble  & Red Blood Cells [B] (0.04); ~\textcolor{blue}{\textit{Hematocrit [B] (0.04)}}; ~\textcolor{blue}{\textit{Hemoglobin [B] (0.04)}}; ~\textcolor{blue}{\textit{Glucose [B] (0.04)}}; ~PT [B] (0.03). \\
Rubik  & Red Blood Cells [B] (0.03); ~\textcolor{blue}{\textit{Hemoglobin [B] (0.03)}}; ~\textcolor{blue}{\textit{Hematocrit [B] (0.03)}}; ~\textcolor{blue}{\textit{Glucose [B] (0.03)}}; ~PT [B] (0.03). \\
Granite  & \textcolor{blue}{\textit{Hemoglobin [B] (0.01)}}; ~\textcolor{blue}{\textit{Hematocrit [B] (0.01)}}; ~\textcolor{blue}{\textit{Glucose [B] (0.01)}}; ~White Blood Cells [B] (0.01); ~Urea Nitrogen [B] (0.01).\\\toprule

& \multicolumn{1}{c}{\textbf{Dx4: Diabetes
mellitus}}\\\hline
HITF  & \textcolor{red}{\textbf{Glucose [B] (0.21)}}; ~~\textcolor{red}{\textbf{Urea Nitrogen [B] (0.16)}}; ~Hemoglobin [B] (0.09); ~Creatinine [B] (0.08); ~\textcolor{blue}{\textit{Red Blood Cells [B] (0.08)}}. \\
CP-APR  & \textcolor{red}{\textbf{Glucose [B] (0.03)}}; ~Hematocrit [B] (0.03); ~Hemoglobin [B] (0.03); ~Red Blood Cells [B] (0.03); ~Urea Nitrogen [B] (0.03). \\
Marble  & \textcolor{red}{\textbf{Glucose [B] (0.03)}}; ~Red Blood Cells [B] (0.03); ~Hematocrit [B] (0.03); ~Hemoglobin [B] (0.03); ~Urea Nitrogen [B] (0.03). \\
Rubik  & {Red Blood Cells [B] (0.03)}; ~{Hemoglobin [B] (0.03)}; ~{Hematocrit [B] (0.03)}; ~\textcolor{red}{\textbf{Glucose [B] (0.03)}}; {White Blood Cells [B] (0.03)}. \\
Granite  & \textcolor{red}{\textbf{Glucose [B] (0.01)}}; ~pO2 [B] (0.01); ~Hemoglobin [B] (0.01); ~Calcium, Total [B] (0.01); ~Chloride [B] (0.01).\\\toprule

& \multicolumn{1}{c}{\textbf{Dx5: Disorders of fluid electrolyte and acid-base balance}}\\\hline
HITF  & \textcolor{red}{\textbf{Sodium [B] (0.14)}}; ~\textcolor{red}{\textbf{Calcium, Total [B] (0.11)}}; ~White Blood Cells [B] (0.10); ~\textcolor{red}{\textbf{Chloride [B] (0.09)}}; ~\textcolor{blue}{\textit{Phosphate [B] (0.08)}}. \\
CP-APR  & \textcolor{blue}{\textit{Glucose [B] (0.03)}}; ~\textcolor{blue}{\textit{Red Blood Cells [B] (0.03)}}; ~\textcolor{blue}{\textit{Hematocrit [B] (0.03)}}; ~\textcolor{blue}{\textit{Hemoglobin [B] (0.03)}}; ~\textcolor{blue}{\textit{Phosphate [B] (0.03)}}. \\
Marble  & \textcolor{blue}{\textit{Red Blood Cells [B] (0.03)}}; ~\textcolor{blue}{\textit{Glucose [B] (0.03)}}; ~\textcolor{blue}{\textit{Hematocrit [B] (0.03)}}; ~\textcolor{blue}{\textit{Hemoglobin [B] (0.03)}}; ~\textcolor{blue}{\textit{Phosphate [B] (0.03)}}. \\
Rubik  & \textcolor{blue}{\textit{Red Blood Cells [B] (0.03)}}; ~\textcolor{blue}{\textit{Hemoglobin [B] (0.03)}}; ~\textcolor{blue}{\textit{Hematocrit [B] (0.03)}}; ~\textcolor{blue}{\textit{Glucose [B] (0.03)}}; ~\textcolor{red}{\textbf{Urea Nitrogen [B] (0.02)}}. \\
Granite  & \textcolor{blue}{\textit{Glucose [B] (0.01)}}; ~\textcolor{red}{\textbf{Chloride [B] (0.01)}}; ~pO2 [B] (0.01); ~\textcolor{blue}{\textit{Hemoglobin [B] (0.01)}}; ~\textcolor{red}{\textbf{Calcium, Total [B] (0.01)}}.\\\toprule

& \multicolumn{1}{c}{\textbf{Dx6: Bacterial infection in conditions classified elsewhere and of unspecified site}}\\\hline
HITF  & \textcolor{blue}{\textit{Vancomycin [B] (0.37)}}; ~Hemoglobin [B] (0.13); ~Red Blood Cells [B] (0.13); ~Hematocrit [B] (0.12); ~Urea Nitrogen [B] (0.06). \\
CP-APR  & Red Blood Cells [B] (0.03); ~Glucose [B] (0.03); ~Hematocrit [B] (0.03); ~Hemoglobin [B] (0.03); ~Phosphate [B] (0.03). \\
Marble  & Hematocrit [B] (0.03); ~Red Blood Cells [B] (0.03); ~Glucose [B] (0.03); ~Hemoglobin [B] (0.03); ~Phosphate [B] (0.03). \\
Rubik  & Red Blood Cells [B] (0.03); ~Glucose [B] (0.03); ~Hematocrit [B] (0.03); ~Hemoglobin [B] (0.03); ~\textcolor{red}{\textbf{White Blood Cells [B] (0.03)}}. \\
Granite  & PT [B] (0.01); ~Glucose [B] (0.01); ~Hemoglobin [B] (0.01); ~Phosphate [B] (0.01); ~Hematocrit [B] (0.01).\\\toprule

& \multicolumn{1}{c}{\textbf{Dx7: Iron deficiency anemias}}\\\hline
HITF  & \textcolor{red}{\textbf{Transferrin [B] (0.26)}};  ~\textcolor{red}{\textbf{Total Iron Binding Capacity [B] (0.25)}}; ~\textcolor{red}{\textbf{Iron [B] (0.19)}}; ~\textcolor{red}{\textbf{Ferritin [B] (0.17)}}; ~Vitamin B12 [B] (0.04) \\
CP-APR  & \textcolor{red}{\textbf{Hematocrit [B] (0.03)}}; ~\textcolor{red}{\textbf{Hemoglobin [B] (0.03)}}; ~Glucose [B] (0.03); ~\textcolor{red}{\textbf{Red Blood Cells [B] (0.03)}}; ~Urea Nitrogen [B] (0.03). \\
Marble  & \textcolor{red}{\textbf{Hemoglobin [B] (0.03)}}; ~\textcolor{red}{\textbf{Hematocrit [B] (0.03)}}; ~\textcolor{red}{\textbf{Red Blood Cells [B] (0.03)}}; ~Glucose [B] (0.03); ~Urea Nitrogen [B] (0.03). \\
Rubik  & \textcolor{red}{\textbf{Red Blood Cells [B] (0.03)}}; ~\textcolor{red}{\textbf{Hemoglobin [B] (0.03)}}; ~\textcolor{red}{\textbf{Hematocrit [B] (0.03)}}; ~{Glucose [B] (0.03)}; {White Blood Cells [B] (0.03)}. \\
Granite  & Calcium, Total [B] (0.01); ~Chloride [B] (0.01); ~Glucose [B] (0.01); ~\textcolor{red}{\textbf{Hemoglobin [B] (0.01)}}; ~\textcolor{red}{\textbf{Hematocrit [B] (0.01)}}.\\\toprule

& \multicolumn{1}{c}{\textbf{Dx8: Chronic bronchitis}}\\\hline
HITF  & \textcolor{red}{\textbf{Neutrophils [B] (0.32)}}; ~\textcolor{red}{\textbf{Lymphocytes [B] (0.30)}}; ~\textcolor{blue}{\textit{Calculated Total CO2 [B] (0.14)}}; ~\textcolor{blue}{\textit{pCO2 [B] (0.07)}}; ~\textcolor{red}{\textbf{Eosinophils [B] (0.07)}}. \\
CP-APR  & Glucose [B] (0.03); ~Red Blood Cells [B] (0.03); ~Hemoglobin [B] (0.03); ~Hematocrit [B] (0.03); ~\textcolor{red}{\textbf{White Blood Cells [B] (0.03)}}. \\
Marble  & Glucose [B] (0.03); ~Red Blood Cells [B] (0.03); ~Hemoglobin [B] (0.03); ~Hematocrit [B] (0.03); ~\textcolor{red}{\textbf{White Blood Cells [B] (0.03)}}. \\
Rubik  & Glucose [B] (0.03); ~Red Blood Cells [B] (0.03); ~Hemoglobin [B] (0.03); ~Hematocrit [B] (0.03); ~Urea Nitrogen [B] (0.03). \\
Granite  & Glucose [B] (0.01); ~Chloride [B] (0.01); ~\textcolor{blue}{\textit{pO2 [B] (0.01)}}; ~Hemoglobin [B] (0.01); ~Urea Nitrogen [B] (0.01).\\\toprule

& \multicolumn{1}{c}{\textbf{Dx9: Arterial embolism and thrombosis}}\\\hline
HITF  & \textcolor{red}{\textbf{PT [B] (0.77)}};  ~{Sodium, Whole Blood [B] (0.06)}; ~\textcolor{blue}{\textit{Lactate [B] (0.05)}}; ~{Chloride, Whole Blood [B] (0.03)}; ~Vancomycin [B] (0.03) \\
CP-APR  & Hemoglobin [B] (0.03); ~Hematocrit [B] (0.03); ~Red Blood Cells [B] (0.03); ~Glucose [B] (0.03); ~\textcolor{blue}{\textit{Urea Nitrogen [B] (0.03)}}. \\
Marble  & Hematocrit [B] (0.02); ~Red Blood Cells [B] (0.02); ~Hemoglobin [B] (0.02); ~Glucose [B] (0.02); ~\textcolor{red}{\textbf{PT [B] (0.02)}}. \\
Rubik  & {Red Blood Cells [B] (0.03)}; ~{Hematocrit [B] (0.03)}; ~{Hemoglobin [B] (0.03)}; ~{Glucose [B] (0.03)}; \textcolor{red}{\textbf{PT [B] (0.03)}}. \\
Granite  & Chloride [B] (0.01); ~Glucose [B] (0.01); ~Hemoglobin [B] (0.01); ~\textcolor{blue}{\textit{pO2 [B] (0.01)}}; ~\textcolor{blue}{\textit{Calcium, Total [B] (0.01)}}.\\\toprule

&\multicolumn{1}{c}{\begin{tabular}[c]{@{}l@{}} \textbf{Dx10: Symptoms involving nervous and musculoskeletal systems} \end{tabular}} \\\hline
HITF  & MCH [B] (0.39); ~Neutrophils [B] (0.25); ~Lymphocytes [B] (0.24); ~Eosinophils [B] (0.05); ~Monocytes [B] (0.04). \\
CP-APR  & \textcolor{blue}{\textit{Glucose [B] (0.03)}}; ~Red Blood Cells [B] (0.03); ~Hematocrit [B] (0.03); ~Hemoglobin [B] (0.03); ~White Blood Cells [B] (0.03). \\
Marble  & \textcolor{blue}{\textit{Glucose [B] (0.03)}}; ~Red Blood Cells [B] (0.03); ~Hematocrit [B] (0.03); ~Hemoglobin [B] (0.03); ~PT [B] (0.03). \\
Rubik  & Red Blood Cells [B] (0.03); ~\textcolor{blue}{\textit{Glucose [B] (0.03)}}; ~Hematocrit [B] (0.03); ~Hemoglobin [B] (0.03); ~Phosphate [B] (0.02). \\
Granite  & \textcolor{blue}{\textit{Glucose [B] (0.01)}}; ~Chloride [B] (0.01); ~Hemoglobin [B] (0.01); ~Hematocrit [B] (0.01); ~pO2 [B] (0.01).\\\bottomrule

\end{tabular}
\end{table*}

\subsection{Clinically Relevant Phenotypes}\label{app:phenotypes}
In Section 5.5.1, we run cHITF to generate 50 phenotypes, out of which 18 were annotated by the clinician as ``clinically relevant''. We showed three examples in Table 4 due to space limit. Here we exhibit in Table~\ref{app:tab:phenotypes_chitf} the remaining 15 phenotypes (indexed 4 to 18) that were annotated as clinically relevant.

\begin{table*}
\renewcommand{\arraystretch}{1.1}
  \centering
  \caption{Examples of the phenotypes inferred by cHITF that are clinically relevant.}\label{app:tab:phenotypes_chitf}
\begin{tabular}{cl}\toprule

\multicolumn{2}{l}{\textbf{Phenotype 4}}\\\hline
\textcolor{Plum}{\textit{Dx}} & \textcolor{Plum}{Other diseases of endocardium (0.775); ~~Acute pulmonary heart disease (0.049); ~~Other venous embolism and thrombosis (0.049);~~...}\\
\textcolor{red}{\textit{Rx}}  & \textcolor{red}{Heparin Sodium (0.561); ~~Warfarin (0.428); ~~Enoxaparin Sodium (0.011).}\\
\textcolor{blue}{\textit{Lab}}& \textcolor{blue}{PTT [Blood] (0.695); ~~PT [Blood] (0.305); ~~Thrombin [Blood] (0.001).}\\\toprule

\multicolumn{2}{l}{\textbf{Phenotype 5}}\\\hline
\textcolor{Plum}{\textit{Dx}} & \textcolor{Plum}{Diseases of pancreas (0.963); ~~Cholelithiasis (0.030); ~~Other disorders of biliary tract (0.005);~~...}\\
\textcolor{red}{\textit{Rx}}  & \textcolor{red}{Pantoprazole Sodium (0.140); ~~Metoprolol (0.132); ~~Piperacillin-Tazobactam Na (0.086).}\\
\textcolor{blue}{\textit{Lab}}& \textcolor{blue}{Lipase [Blood] (0.291); ~~Amylase [Blood] (0.252); ~~Asparate Aminotransferase (AST) [Blood] (0.095); ...}\\\toprule

\multicolumn{2}{l}{\textbf{Phenotype 6}}\\\hline
\textcolor{Plum}{\textit{Dx}} & \textcolor{Plum}{Other disorders of urethra and urinary tract (1.000).}\\
\textcolor{red}{\textit{Rx}}  & \textcolor{red}{Lansoprazole Oral Suspension (1.000).}\\
\textcolor{blue}{\textit{Lab}}& \textcolor{blue}{Bicarbonate [Blood] (0.263); ~~Urea Nitrogen [Blood] (0.210); ~~Chloride [Blood] (0.168); ...}\\\toprule

\multicolumn{2}{l}{\textbf{Phenotype 7}}\\\hline
\textcolor{Plum}{\textit{Dx}} & \textcolor{Plum}{Iron deficiency anemias (0.300); ~~Gastrointestinal hemorrhage (0.284); ~~Intestinal infections due to other organisms (0.211); ...}\\
\textcolor{red}{\textit{Rx}}  & \textcolor{red}{MetRONIDAZOLE (FLagyl) (0.208); ~~Ferrous Sulfate (0.192); ~~Vancomycin Oral Liquid (0.110); ...}\\
\textcolor{blue}{\textit{Lab}}& \textcolor{blue}{Transferrin [Blood] (0.260); ~~Iron Binding Capacity, Total [Blood] (0.255); ~~Iron [Blood] (0.193); ...}\\\toprule

\multicolumn{2}{l}{\textbf{Phenotype 8}}\\\hline
\textcolor{Plum}{\textit{Dx}} & \textcolor{Plum}{Asthma (0.254); ~~Overweight, obesity and other hyperalimentation (0.115); ~~Chronic bronchitis (0.112); ...}\\
\textcolor{red}{\textit{Rx}}  & \textcolor{red}{PredniSONE (0.337); ~~Albuterol 0.083\% Neb Soln (0.151); ~~Ipratropium Bromide Neb (0.115); ...}\\
\textcolor{blue}{\textit{Lab}}& \textcolor{blue}{Bicarbonate [Blood] (0.197); ~~Neutrophils [Blood] (0.163); ~~Lymphocytes [Blood] (0.137); ...}\\\toprule

\multicolumn{2}{l}{\textbf{Phenotype 9}}\\\hline
\textcolor{Plum}{\textit{Dx}} & \textcolor{Plum}{Hypotension (0.505); ~~Old myocardial infarction (0.190); ~~Conduction disorders (0.088); ...}\\
\textcolor{red}{\textit{Rx}}  & \textcolor{red}{Aspirin (0.238); ~~Heparin (0.160); ~~Sodium Chloride 0.9\%  Flush (0.130); ...}\\
\textcolor{blue}{\textit{Lab}}& \textcolor{blue}{Creatine Kinase (CK) [Blood] (0.613); ~~Troponin T [Blood] (0.286); ~~Creatine Kinase, MB Isoenzyme [Blood] (0.102).}\\\toprule

\multicolumn{2}{l}{\textbf{Phenotype 10}}\\\hline
\textcolor{Plum}{\textit{Dx}} & \textcolor{Plum}{Acute kidney failure (1.000).}\\
\textcolor{red}{\textit{Rx}}  & \textcolor{red}{Sodium Bicarbonate (0.246); ~~Furosemide (0.164); ~~Insulin (0.082); ...}\\
\textcolor{blue}{\textit{Lab}}& \textcolor{blue}{Creatinine [Blood] (0.409); ~~Urea Nitrogen [Blood] (0.234); ~~Glucose [Blood] (0.056); ...}\\\toprule

\multicolumn{2}{l}{\textbf{Phenotype 11}}\\\hline
\textcolor{Plum}{\textit{Dx}} & \textcolor{Plum}{Coagulation defects (0.431); ~~Acute and subacute necrosis of liver (0.431); ~~Other disorders of biliary tract (0.074); ...}\\
\textcolor{red}{\textit{Rx}}  & \textcolor{red}{Phytonadione (0.203); ~~Piperacillin-Tazobactam Na (0.143); ~~Potassium Chloride (0.142); ...}\\
\textcolor{blue}{\textit{Lab}}& \textcolor{blue}{Alanine Aminotransferase (ALT) [Blood] (0.246); ~~Asparate Aminotransferase (AST) [Blood] (0.214);}\\ & \textcolor{blue}{Alkaline Phosphatase [Blood] (0.175); ...}\\\toprule

\multicolumn{2}{l}{\textbf{Phenotype 12}}\\\hline
\textcolor{Plum}{\textit{Dx}} & \textcolor{Plum}{Diabetes mellitus (0.982); ~~Inflammatory and toxic neuropathy (0.008); ~~Other retinal disorders (0.005); ...}\\
\textcolor{red}{\textit{Rx}}  & \textcolor{red}{Insulin (0.626); ~~Insulin Human Regular (0.035); ~~Potassium Chloride (0.026); ...}\\
\textcolor{blue}{\textit{Lab}}& \textcolor{blue}{Glucose [Blood] (0.292); ~~Urea Nitrogen [Blood] (0.105); ~~Creatinine [Blood] (0.078); ...}\\\toprule

\multicolumn{2}{l}{\textbf{Phenotype 13}}\\\hline
\textcolor{Plum}{\textit{Dx}} & \textcolor{Plum}{Purpura and other hemorrhagic conditions (1.000).}\\
\textcolor{red}{\textit{Rx}}  & \textcolor{red}{Acetaminophen (0.094); ~~Phytonadione (0.082); ~~Furosemide (0.077); ...}\\
\textcolor{blue}{\textit{Lab}}& \textcolor{blue}{Platelet Count [Blood] (0.295); ~~Hematocrit [Blood] (0.108); ~~Red Blood Cells [Blood] (0.095); ...}\\\toprule

\multicolumn{2}{l}{\textbf{Phenotype 14}}\\\hline
\textcolor{Plum}{\textit{Dx}} & \textcolor{Plum}{Persistent mental disorders due to conditions classified elsewhere (0.208);}\\& \textcolor{Plum}{Transient mental disorders due to conditions classified elsewhere (0.142); ~~Episodic mood disorders (0.121); ...}\\
\textcolor{red}{\textit{Rx}}  & \textcolor{red}{Haloperidol (0.649); ~~Quetiapine Fumarate (0.113); ~~Olanzapine (Disintegrating Tablet) (0.111); ...}\\
\textcolor{blue}{\textit{Lab}}& \textcolor{blue}{Hemoglobin [Blood] (0.219); ~~Phosphate [Blood] (0.217); ~~RBC (Urine) (0.142); ...}\\\toprule

\multicolumn{2}{l}{\textbf{Phenotype 15}}\\\hline
\textcolor{Plum}{\textit{Dx}} & \textcolor{Plum}{Hypertensive chronic kidney disease (0.499); ~~Chronic kidney disease (ckd) (0.498);}\\ & \textcolor{Plum}{Nephritis and nephropathy not specified as acute or chronic (0.003); ...}\\
\textcolor{red}{\textit{Rx}}  & \textcolor{red}{Heparin (0.070); ~~Metoprolol Tartrate (0.064); ~~Sodium Chloride 0.9\%  Flush (0.054); ...}\\
\textcolor{blue}{\textit{Lab}}& \textcolor{blue}{Creatinine [Blood] (0.247); ~~RDW [Blood] (0.167); ~~Urea Nitrogen [Blood] (0.108); ...}\\\toprule

\multicolumn{2}{l}{\textbf{Phenotype 16}}\\\hline
\textcolor{Plum}{\textit{Dx}} & \textcolor{Plum}{Chronic liver disease and cirrhosis (0.409); ~~Liver abscess and sequelae of chronic liver disease (0.382);} \\ & \textcolor{Plum}{Other symptoms involving abdomen and pelvis (0.085); ...}\\
\textcolor{red}{\textit{Rx}}  & \textcolor{red}{Albumin 25\% (12.5g / 50mL) (0.122); ~~Lactulose (0.118); ~~Furosemide (0.105); ...}\\
\textcolor{blue}{\textit{Lab}}& \textcolor{blue}{RDW [Blood] (0.133); ~~PTT [Blood] (0.102); ~~PT [Blood] (0.100); ...}\\\toprule

\multicolumn{2}{l}{\textbf{Phenotype 17}}\\\hline
\textcolor{Plum}{\textit{Dx}} & \textcolor{Plum}{Disorders of fluid electrolyte and acid-base balance (0.995); ~~Disorders of the pituitary gland and its hypothalamic control (0.005).}\\
\textcolor{red}{\textit{Rx}}  & \textcolor{red}{\textit{None}}\\
\textcolor{blue}{\textit{Lab}}& \textcolor{blue}{Sodium [Blood] (0.645); ~~Chloride [Blood] (0.327); ~~Sodium, Whole Blood [Blood] (0.015); ...}\\\toprule

\multicolumn{2}{l}{\textbf{Phenotype 18}}\\\hline
\textcolor{Plum}{\textit{Dx}} & \textcolor{Plum}{Essential hypertension (0.202); ~~Cardiac dysrhythmias (0.202); ~~Other forms of chronic ischemic heart disease (0.202); ...}\\
\textcolor{red}{\textit{Rx}}  & \textcolor{red}{\textit{None}}\\
\textcolor{blue}{\textit{Lab}}& \textcolor{blue}{Platelet Count [Blood] (1.000).}\\\toprule

\end{tabular}
\end{table*}

\end{document}